\documentclass[num-refs,serif]{wiley-article}

\usepackage{booktabs}
\usepackage{siunitx}
\usepackage{amsmath} 
\usepackage{adjustbox}
\usepackage{caption}
\usepackage{subcaption}
\usepackage{multicol}
\usepackage{tkz-graph}
\usepackage{tikz}
\usetikzlibrary{positioning}
\usepackage{pgfplots}
\usepackage{lipsum}

\usepgfplotslibrary{statistics}
\usetikzlibrary{pgfplots.statistics, pgfplots.colorbrewer} 
\usepgfplotslibrary{colorbrewer}
\pgfplotsset{compat=1.8, cycle list/Set1-4}
\usetikzlibrary{decorations.pathmorphing}
\usetikzlibrary{external}
\usepackage[procnumbered,ruled,linesnumbered,vlined]{algorithm2e} 
\newcounter{algoline}
\newcommand{\nlast}{\refstepcounter{algoline}\nlset{\rlap{\textsuperscript{*}}}}
\newcommand{\nlastt}{\refstepcounter{algoline}\nlset{\rlap{\textsuperscript{+}}}}
\SetNlSty{bfseries}{\color{black}}{}
\SetKwInput{KwInput}{Inputs}
\SetKwInput{KwOutput}{Output}
\SetKwComment{Comment}{$\triangleright$\ }{}
\makeatletter
\newcommand{\algorithmfootnote}[2][\footnotesize]{%
  \let\old@algocf@finish\@algocf@finish
  \def\@algocf@finish{\old@algocf@finish
    \leavevmode\rlap{\begin{minipage}{\linewidth}
    #1#2
    \end{minipage}}%
  }%
}

\usepackage{environ}
\usepackage{comment}

\NewEnviron{myhideenv}{%
\setbox0\vbox{%
\let\xxwrite\write
\protected\def\write{\immediate\xxwrite}%
{\tiny XX\BODY XX}}
}
\renewcommand{\qed}{\hfill$\square$}
\let\oldnl\nl
\newcommand{\nonl}{\renewcommand{\nl}{\let\nl\oldnl}}
\papertype{RESEARCH ARTICLE}

\title{Enhanced Methods for the Weight Constrained Shortest Path Problem}

\author[1]{Saman Ahmadi}
\author[2]{Guido Tack}
\author[2]{Daniel Harabor}
\author[3]{Philip Kilby}
\author[1]{Mahdi Jalili}

\affil[1]{School of Engineering, RMIT University, Australia}
\affil[2]{Department of Data Science and Artificial Intelligence, Monash University, Australia}
\affil[3]{CSIRO Data61, Canberra, Australia}

\corraddress{School of Engineering, RMIT University, Australia}
\corremail{saman.ahmadi@rmit.edu.au}

\fundinginfo{Research at Monash University is supported by the Australian Research Council (ARC) under grant numbers DP190100013 and DP200100025 as well as a gift from Amazon.}

\runningauthor{S. Ahmadi, G. Tack, D. Harabor, P. Kilby, and M. Jalili}

\begin{document}

\begin{frontmatter}
\maketitle

\begin{abstract}
\textbf{Abstract}\\
The classic problem of \textit{constrained pathfinding} is a well-studied, yet challenging, topic in AI with a broad range of applications in various areas such as communication and transportation. 
The Weight Constrained Shortest Path Problem (WCSPP), the base form of constrained pathfinding with only one side constraint, aims to plan a cost-optimum path with limited weight/resource usage.
Given the bi-criteria nature of the problem (i.e., dealing with the cost and weight of paths), methods addressing the WCSPP have some common properties with bi-objective search.
This paper leverages the recent state-of-the-art techniques in both constrained pathfinding and bi-objective search and presents two new solution approaches to the WCSPP on the basis of A* search, both capable of solving hard WCSPP instances on very large graphs.
We empirically evaluate the performance of our algorithms on a set of large and realistic problem instances and show their advantages over the state-of-the-art algorithms in both time and space metrics.
This paper also investigates the importance of priority queues in constrained search with A*.
We show with extensive experiments on both realistic and randomised graphs how bucket-based queues without tie-breaking can effectively improve the algorithmic performance of exhaustive A*-based bi-criteria searches.
%
\end{abstract}
\end{frontmatter}
\section{Introduction}
The Weight Constrained Shortest Path Problem (WCSPP) is well known as a technically challenging variant of the classical shortest path problem.
The objective of the point-to-point WCSPP is to find a minimum-cost (shortest) path between two points in a graph such that the total weight (or resource consumption) of the path is limited. 
Formally, given a directed graph $G=(S,E)$ with a finite set of states $S$ and a set of edges $E$, each edge labelled with a pair of attributes ${(\mathit{cost}, \mathit{weight})}$, the task is to find a path~$\pi$ from $\mathit{start} \in S$ to $\mathit{goal} \in S$ such that $\mathit{cost}(\pi)$ is minimum among all paths between $\mathit{start}$ and $\mathit{goal}$ subject to $\mathit{weight}(\pi) \leq W$ where $W$ is the given weight limit.
The problem has been shown to be NP-complete \cite{handler1980dual}.

The WCSPP, as a core problem or a subroutine in larger problems, can be seen in various real-world applications in diverse areas such as telecommunication networks, transportation, planning and scheduling, robotics and game development.
A typical example of the WCSPP as a core problem is finding the least-cost connection between two nodes in a network with relays such that the weight of any path between two consecutive relays does not exceed a given upper bound \cite{LaporteP11}. 
As a subroutine, examples can be solving the WCSPP as a sub-problem in the context of column generation (a solving technique in linear programming) \cite{zhu2012three}, or obtaining valid (constrained) paths needed for trip planning such as in vehicle routing problems.
A valid path in this context can be defined as the quickest path between two points such that its energy requirement is less than the (limited) available energy/petrol \cite{Storandt12} or an energy-efficient path that is at most a constant factor longer than the shortest time/distance path for every transport request \cite{AhmadiTHK21_cp}.\par
Looking at the performance of the state-of-the-art search techniques for constrained pathfinding over the last decade, we can see fast algorithms that are able to optimally solve small-size instances in less than a second.
However, given the significance and difficulty of the problem, especially in large-size graphs, the WCSPP still remains a challenging problem that needs more efficient solutions in terms of time and space (memory usage).
The main motivations behind this research are therefore as follows.
\begin{enumerate}[nosep]
    \item Hardware/resource requirement: we are mainly concerned by the space requirement of recent solutions to the WCSPP, as this metric may drastically limit their application in large graphs. 
    Although space usage is highly correlated with runtime and (faster) methods are likely to consume less memory, there are still some procedures in the state of the art that are space-demanding.
    As an example, the \textit{partial path matching} procedure in the recent bidirectional approaches needs to store partial paths of each direction to be able to build a complete $\mathit{start}$-$\mathit{goal}$ path later during the search.
    In terms of memory, this procedure is quite demanding, since the number of partial paths grows exponentially over the course of the search.
    \item Applicability of algorithms: Although recent algorithms have shown that the problem can be solved faster bidirectionally, there are several problems where searching backwards is not possible or at least is not straightforward.
    As an example, in planning feasible paths for electric vehicles, the forward search can propagate estimated battery levels based on the current battery level at the origin location, but a backward search would require the final battery level at the destination to be known without actually completing the search.
\end{enumerate}
Among other variants of the shortest path problem, the Bi-Objective Shortest Path Problem (BOSPP) is considered conceptually closest to the WCSPP in terms of the number of criteria involved.
Compared to the WCSPP where we look for a single optimal solution, methods to the BOSPP find a set of Pareto-optimal solution paths, a set in which every individual solution offers a path that minimises the bi-criteria problem in both \textit{cost} and \textit{weight}.
The BOSPP is a difficult problem, technically more difficult than the WCSPP. 
Therefore, if there exists a time and space-efficient solution to the BOSPP, it is worth investigating whether it can be adapted for the WCSPP as it may help us to address our serious concerns about the applicability of recent approaches.

\textbf{Contributions:}
For our first contribution, we design two new solution approaches to the WCSPP, called {WC-EBBA*}\textsubscript{par} and {WC-A*}, by extending the recent fast algorithms available for both WCSPP and BOSPP.
We conduct an extensive analysis of our A*-based methods through a set of challenging WCSPP instances, and compare their algorithmic performance with \textit{six} recent WCSPP approaches to understand the strengths and weaknesses of the algorithms in both time and space metrics.
The results show that the proposed approaches are very effective in reducing the search time and also the total memory requirement of the constrained search in almost all levels of tightness, outperforming some recent solutions by orders of magnitude faster runtime.

As our second contribution, we empirically study variations in a fundamental component of bi-criteria search with A*: priority queue.
We briefly explain our motivations for this extension.
The bi-criteria search with A* works based on enumeration, i.e., all promising paths are evaluated until there is no feasible solution path better than the discovered optimum path.
In such search schemes, it is highly expected that the search generates dominated paths, i.e., paths for which we can find at least one other path with better $\mathit{cost}$ and $\mathit{weight}$.
Recent studies have shown that the lazy dominance test in bi-objective A* contributes to faster query times.
Nonetheless, it inevitably increases the number of suspended paths in the queue, mainly because the queue accepts dominated paths, making queue operations lengthier.
To this end, we extend our experiments and investigate the impacts of three types of priority queues on the performance of constrained search with A*, namely bucket, hybrid and binary heap queues.
We define our hybrid queue to be a two-level bucket-heap structure.
Among these types, we are interested in finding a type that appropriately addresses the inefficiencies incurred by the lazy dominance test in the exhaustive search of A*.
The results of our extensive experiments show that bucket-based queues (bucket queue and hybrid queue) perform far better than the traditional binary heap queues in managing the significant number of paths produced in difficult WCSPP instances.

For our third contribution, we investigate the impact of tie-breaking on the algorithmic performance of constrained search with A*, i.e., the total computational cost of handling cases where two paths in the priority queue have the same estimated $\mathit{cost}$.
For this purpose, we evaluate two variants of our priority queues (with and without tie-breaking) on realistic and also randomised graphs, and show how recent A*-based algorithms can expedite their search by simply ordering unexplored paths solely based on their primary cost, without a tie-breaking mechanism.
Our detailed analysis reveals that tie-breaking has a very limited ability to reduce the number of path expansions. 
Instead, it incurs significant computational overheads to break the ties among paths in the queue.
In other words, in difficult bi-criteria problems, the priority queue's effort in breaking the ties between paths is not paid off by the search, and we are better off simply ordering paths in the queue based on their primary costs.

The paper is structured as follows.
In the next section, we survey related work in the WCSPP literature and explain, at a high level, the main features of our selected algorithms.
We formally define the problem in Section~\ref{sec:notation}.
The main features of our constrained A* searches are explained in this section.
In Section~\ref{sec:WC_A}, we introduce our unidirectional search scheme {WC-A*}, followed by our parallel algorithm {WC-EBBA*}\textsubscript{par} presented in Section~\ref{sec:WC_EBBA_par}.
Section~\ref{sec:practicall} describes our practical considerations for the empirical study in Section~\ref{sec:experiment}.
Our bucket and hybrid priority queues are introduced in this section.
Section~\ref{sec:experiment} contains the experimental results and also our detailed study on the impacts of tie-breaking and priority queues on the computation time.
Finally, Section~\ref{sec:conclusion} presents conclusions and directions for future work.
%
\section{Background and Selected Algorithms}
\label{sec:literature}
This section reviews classic and recent methods proposed to address the WCSPP.
In our literature review, we also provide cases where BOSPP approaches have been extended to address the WCSPP.
We conclude this section by nominating two promising extensions from both domains, along with six recent algorithms in the literature for the WCSPP as our baseline for experiments.

The WCSPP and its extended version with more than one weight constraint, which is known as the Resource Constrained Shortest Path Problem (RCSPP), are well-studied topics in AI. 
Since WCSPP is just a special case of the RCSPP, all algorithms designed to solve the RCSPP are naturally capable of addressing the WCSPP.
This applicability can be seen in almost all works on the RCSPP in the literature, where solution approaches are also computationally evaluated on benchmark instances of the WCSPP.
Given the close relationship between the two problems, there exist various types of solution approaches in the literature designed to tackle constrained pathfinding in different settings.
\citet{pugliese2013survey} presented a summary of traditional exact solution approaches to the RCSPP (methods that solve the problem to optimality). 
In their survey paper, strategies for the gap-closing step (the main search for the optimal solution) are discussed in three categories: path ranking, dynamic programming and branch-and-bound (B\&B) approach.
In the first category (path ranking), \textit{k}-paths are obtained, usually by solving the \textit{k}-shortest path problem, and then sorted (ranked) based on their cost in ascending order. 
For the search phase, the shortest paths are successively evaluated until a feasible path (with a valid resource consumption) is obtained \cite{santos2007improved}.
As the computational effort in the first phase depends on the value of \textit{k}, and since the value of \textit{k} exponentially grows with respect to the size of the network, selecting an efficient path ranking method is crucial in this category of solutions. 
Solution approaches in the second category generally establish a dynamic programming framework to extend partial paths from origin to destination, normally via a labelling method. 
Labels in this context can be defined as tuples that contain some essential information about partial paths such as cost, resource consumption and parent pointers.
The search in labelling methods can be improved by using pruning rules, or sometimes by integrating other methods to reduce the search space, normally after a preprocessing phase \cite{DumitrescuB03}.
The solution in this case is a feasible least-cost label (with a valid resource consumption) at the destination.
The third category of algorithms (B\&B) tries to find feasible solutions by applying an enumeration procedure (e.g., via a depth-first search) along branches of the search tree.
This step usually involves adding a node to the end of the partial path while using lower-bounding information to prune infeasible paths and remove nodes from the end of the partial path (backtracking).
It is evident that the performance of B\&B approaches greatly depends on the quality of the bounds and also the pruning strategy \cite{MuhandiramgeB09}.\par
\citet{FeroneFFP20} presented a short survey on the recent state-of-the-art exact algorithms designed for constrained pathfinding.
According to their report, we can still see effective methods from each category of solution approaches.
The B\&B-like solution method presented by \citet{lozano2013exact} conducts a systematic search via the depth-first search scheme for the RCSPP.
Their so-called \textit{Pulse} algorithm was equipped with \textit{infeasible} and \textit{dominance} pruning strategies and was computationally compared with the path ranking method of \citet{santos2007improved} on their benchmark instances for the WCSPP.
\citeauthor{lozano2013exact} concluded that, regardless of the tightness of the constraint, the Pulse algorithm performs roughly 40 times faster than the path ranking method of \citet{santos2007improved}.
\citet{HorvathK16} extended the integer programming techniques presented by \citet{Garcia09} for the RCSPP and designed a B\&B solution with improved cutting and heuristic methods.
Their computational results on the WCSPP instances of \citet{santos2007improved} show that their proposed branch-and-cut procedure delivers a comparable performance to the Pulse algorithm, both showing larger computational times with increasing input size.
However, \citet{sedeno2015enhanced} later developed an enhanced path ranking approach called \textit{CSP}, which was able to exploit pruning strategies of \citet{lozano2013exact}.
They computationally tested their CSP algorithm and compared its performance against Pulse on a new set of realistic large-size instances for the WCSPP.
\citeauthor{sedeno2015enhanced} reported that although Pulse might perform better than their CSP algorithm mainly on small instances, it generally performs poorly on difficult large instances, leaving 28\% of the instances unsolved even after a two-day timeout.
Their results also show that the enhanced path ranking algorithm CSP runs generally faster across a range of constraint tightness and solves more instances than Pulse, yet still leaving 4\% of the instances unsolved.
Later in \cite{BolivarLM14}, \citeauthor{BolivarLM14} presented several acceleration strategies for Pulse to solve the WCSPP with replenishment, mainly by introducing a queuing strategy that limits the depth of the Pulse search, and also a path completion strategy that allows the search to possibly update the primal upper bound early.
However, \citeauthor{BolivarLM14} reported that the accelerated Pulse method is just about 1-5\% faster than the standard Pulse algorithm on the WCSPP instances.

Among recent solutions, the dynamic programming approach of \citet{thomas2019exact} solves the RCSPP with the help of heuristic search.
Inspired by the idea of a \textit{half-way point} utilised in the bidirectional dynamic programming approach of \citet{righini2006symmetry}, \citeauthor{thomas2019exact} developed a bidirectional A* search informed with forward and backward lower bounding information required for both pruning and the A*'s best-first search.
In their \textit{{RC-BDA*}} algorithm, the resource budget is divided equally and then allocated to the forward and backward searches. 
In other words, the algorithm stops expanding partial paths with a resource consumption larger than half of the total resource budget, allowing both searches to meet at 50\% resource half-way points.
A complete path in this method can be obtained by joining forward and backward partial paths.
\citeauthor{thomas2019exact} evaluated {RC-BDA*} on some instances of \citet{sedeno2015enhanced} for the WCSPP.
Their empirical results illustrate the effectiveness of their bidirectional search on large instances, being able to solve 99\% of the total instances within five hours.
However, it can be seen that {RC-BDA*} is dominated by both Pulse of \citet{lozano2013exact} and CSP of \citet{sedeno2015enhanced} on many small-size instances.
\citeauthor{thomas2019exact} also tested a unidirectional version of {RC-BDA*} by running a single forward constrained A* search.
Compared to the bidirectional search {RC-BDA*}, they reported weaker performance (on all levels of tightness) in both runtime and memory requirement for the unidirectional variant on the instances of \cite{santos2007improved}.
Given the success of the bidirectional search in large instances, \citet{cabrera2020exact} developed a parallel framework to execute Pulse bidirectionally. 
To prevent the search from falling into unpromising deep branches, bidirectional Pulse (\textit{BiPulse}) limits the depth of the Pulse search and employs an adapted form of the queuing strategy proposed by \citet{BolivarLM14} to store and later expand halted partial paths in the breadth-first search manner.
Similar to {RC-BDA*}, BiPulse handles collision between search frontiers by joining forward and backward partial paths.
This method in BiPulse is further extended by joining partial paths with their complementary cost/resource shortest path to obtain and update the incumbent solution early. 
\citeauthor{cabrera2020exact} evaluated BiPulse on a subset of 
large instances of \citet{sedeno2015enhanced} for the WCSPP and compared BiPulse with the unidirectional Pulse and {RC-BDA*} of \citet{thomas2019exact}.
Their results show that BiPulse delivers better performance and solves more instances than both Pulse and {RC-BDA*} on medium-size instances, while leaving 3\% of the instances unsolved after four hours of runtime.
Both {RC-BDA*} and BiPulse recently won the Glover-Klingman prize, awarded by Networks \cite{GloverPrize}.

Motivated by the applications of the WCSPP, we modified and then improved {RC-BDA*} of \citet{thomas2019exact} for the WCSPP \cite{AhmadiTHK21}.
We optimised various components of the {RC-BDA*} search and proposed an enhanced dynamic programming framework that was able to solve, for the first time, 100\% of the benchmark instances of \citet{sedeno2015enhanced}, each within just nine minutes of runtime.
We showed that our Enhanced Biased Bidirectional A* algorithm for the WCSPP, which we call \textit{WC-EBBA*}, outperforms the state-of-the-art algorithms on almost all instances by several orders of magnitude.
{WC-EBBA*} follows the two-phase search of the conventional approaches.
In the first (initialisation) phase, the search establishes both lower and upper bounds needed for the main search while reducing the search space by removing states that are out-of-bounds.
Further, compared to {RC-BDA*} where the bidirectional searches are expected to meet at the 50\% half-way point, and also other approaches with dynamic half-way point \cite{TilkRGI17}, the initialisation phase of {WC-EBBA*} decides on the budgets of each search direction, enabling the search frontiers to meet at any fraction of the resource budget (not just at the 50\% half-way point).
In the second phase, or the main search, {WC-EBBA*} employs several strategies to expedite the searches in each direction.
For the dominance checking procedure, {WC-EBBA*} borrows a fast strategy from the bi-objective search context to detect and prune dominated partial path in a constant time.
{WC-EBBA*} also takes advantage of node ordering in A* to perform more efficient partial path matching, a procedure that plays an important role in both computation time and space usage of the algorithm. \par
Given the success of the recent bidirectional search algorithms in solving difficult WCSPP problems, we still need to investigate algorithms that are more efficient in terms of space, and even sometimes algorithms that are simpler in terms of search structure.
For instance, some of the recent algorithms (BiPulse, {RC-BDA*} and {WC-EBBA*}) employ a path-matching procedure that handles frontier collision.
To fulfil this task, the search needs to store all the explored partial paths in both directions.
However, this requirement can significantly increase the memory usage of the algorithms, since the number of partial paths can grow exponentially over the course of the search.
Quite recently in \citet{AhmadiTHK22_socs}, we leveraged our bi-objective bidirectional A*-based search algorithm BOBA* \cite{AhmadiTHK21_esa, AhmadiTHK21_socs} and introduced {WC-BA*} as a bidirectional WCSPP method that does not need to handle frontier collision.
{WC-BA*} executes its forward and backward searches in parallel and on different attribute orderings (e.g., forward A* search working on $\mathit{cost}$ and backward A* search on $\mathit{weight}$).
As each individual search in {WC-BA*} is complete, partial paths are no longer stored.
Besides the early solution update method we proposed in \citet{AhmadiTHK21}, {WC-BA*} has a unique method called \textit{heuristic tuning}, which allows the search to improve its initial lower bounds during the search.
We evaluated {WC-BA*} on very large graphs and compared its performance against the recent algorithms in the literature, including {WC-EBBA*}.
The results of our experiments over 2000 new instances showed that {WC-BA*} outperforms {WC-EBBA*} by up to 60\% (35\%) in terms of computation time (space) in various constrained problem instances.
This observation motivated us to further explore other efficient BOSPP methods for their constrained variant.
We briefly describe the main features of our studied algorithms as follows.

\textbf{Selected algorithms:} 
From the existing WCSPP solutions, we consider the recent {WC-BA*} and {WC-EBBA*} algorithms and also the award-winning BiPulse algorithm as our baseline.
For our extended empirical study, we also evaluate the other award-winning algorithm {RC-BDA*}, the ranking method CSP and the B\&B method Pulse.

\textbf{Proposed algorithms:}
We target the recent fast A*-based BOSPP and WCSPP methods and develop two new algorithms for the WCSPP, namely: 
\begin{itemize}
    \item {WC-A*}: The adapted version of the bi-objective A* search algorithm (\textit{BOA*}) originally presented by \citet{ulloa2020simple} and enhanced by us in \citet{AhmadiTHK21_esa}. 
    BOA* is a simple unidirectional search method, so its WCSPP variant will help us to investigate a simple solution to problems that cannot be solved bidirectionally.
    Our adapted algorithm {WC-A*} leverages improvements proposed for BOA*.
  \item {WC-EBBA*}\textsubscript{par}: The extended version of our recent {WC-EBBA*} algorithm for the WCSPP \cite{AhmadiTHK21}, improved with parallelism. 
    In contrast to the standard {WC-EBBA*} algorithm where the search only explores one direction at a time, the new variant provides the algorithm with the opportunity of executing its forward and backward searches concurrently. 
    This new feature allows us to accelerate the {WC-EBBA*}'s bidirectional search and solve more instances in a limited time.
\end{itemize}
The suffix \textit{par} in {WC-EBBA*}\textsubscript{par} denotes that the algorithm is run on a \textit{parallel} framework, ideally with two CPU cores.
Table~\ref{table:alg_summary} summarises the main features of all eight algorithms studied in this paper, including their proposed speed-up techniques.
\begin{table}[!t]
\centering
\setlength{\tabcolsep}{3.0pt}
\renewcommand{\arraystretch}{1}
\caption{\small An overview of the main features of the studied algorithms.}
\label{table:alg_summary}
\begin{tabular}{l *{8}{c}}
\headrow
Feature / Alg.  &    WC-A* &  WC-BA* &  WC-EBBA*\textsubscript{par}  & WC-EBBA* & BiPulse & Pulse & RC-BDA* & CSP\\
Bidirectional &   No & Yes & Yes & Yes & Yes & No & Yes & No \\
Frontier Collision  &   - & No & Yes & Yes & Yes & - & Yes & -   \\
Parallel Framework  &   No & Yes & Yes & No & Yes & No & No & No   \\
Early Solution Update  &   Yes & Yes & Yes & Yes & Yes & No & No & No\\
Heuristic Tuning  &   No & Yes & No & No & No & No & No & No\\
\bottomrule
\end{tabular}
\end{table}
\section{Notation and Search Strategy}
\label{sec:notation}
Consider a directed graph $G=(S,E)$ with a finite set of states $S$ and a set of edges $E \subseteq S \times S$.
Every edge $e \in E$ has two non-negative attributes that can be accessed via the cost function ${\bf cost}:E \rightarrow \mathbb{R}^+ \times \mathbb{R}^+$.
For the sake of simplicity, in our algorithmic description, we replace the conventional $(\mathit{cost}, \mathit{weight})$ attribute representation with $(\mathit{cost_1}, \mathit{cost_2})$.
Further, in our notation, every boldface function returns a tuple, so for the edge cost function, we have ${\textbf{cost}}=(\mathit{cost_1}, \mathit{cost_2})$.
Expanding a typical state $u$ \textit{generates} a set of successor states, denoted $\mathit{Succ}(u)$.
A path is a sequence of states $u_i \in S$ with $ i \in \{1, \dots, n \}$.
The $\textbf{cost}$ of path $\pi=\{u_1,u_2,u_3,\dots,u_n\}$ is then the sum of corresponding attributes on all the edges constituting the path, namely $\textbf{cost}(\pi) = \sum_{i=1}^{n-1}{\textbf{cost}(u_i,u_{i+1})}$.
The Weight Constrained Shortest Path Problem (WCSPP) aims to find a {\bf cost}-optimal $\mathit{start}$-$\mathit{goal}$ solution path such that the secondary cost of the optimum path is within the upper bound $W$ (conventionally $\mathit{weight}$ limit).
We formally define {\bf cost}-optimal solution paths below.
\begin{definition}
$\pi^*$ is a {\bf cost}-optimal $\mathit{start}$-$\mathit{goal}$ solution path for the WCSPP if it has the lexicographically smallest ($\mathit{cost_1},\mathit{cost_2}$) among all paths from $\mathit{start} \in S$ to $\mathit{goal} \in S$ such that $\mathit{cost_2}(\pi^*) \leq W$.
\end{definition}
We abstract from the two possible attribute orderings $(1,2)$ and $(2,1)$ in our notation by using a pair $(p,s)$  (for \textit{primary} and \textit{secondary}) with $p, s \in \{1,2\}$ and $p \neq s$.
Given an attribute ordering $(p,s)$, we now define lexicographical order on ($\mathit{cost_p},\mathit{cost_s}$), followed by the definition of search \textit{directions}.
\begin{definition}
Path $\pi$ is lexicographically smaller than path $\pi'$ in the ($\mathit{cost_p},\mathit{cost_s}$) order if $\mathit{cost_p}(\pi)<\mathit{cost_p}(\pi')$, or $\mathit{cost_p}(\pi)=\mathit{cost_p}(\pi')$ and $\mathit{cost_s}(\pi)<\mathit{cost_s}(\pi')$.
Path $\pi$ is equal to path $\pi'$ if $\mathit{cost_p}(\pi)=\mathit{cost_p}(\pi')$ and $\mathit{cost_s}(\pi)=\mathit{cost_s}(\pi')$.
\end{definition}
%
%
\begin{definition}
The search is called \textit{forward} if it explores the graph from the $\mathit{start}$ state to the $\mathit{goal}$ state.
Otherwise, searching from $\mathit{goal}$ towards $\mathit{start}$ is called \textit{backward}.
\end{definition}
In our notation, we generalise both possible search directions by searching in direction~$d \in$ \{\textit{forward}, \textit{backward}\} from an \textit{initial} state to a \textit{target} state.
Therefore, the (\textit{initial}, \textit{target}) pair would be ($\mathit{start},\mathit{goal}$) in the forward search and ($\mathit{goal},\mathit{start}$) in the backward search.
In addition, we define $d'$ to always be the opposite direction of $d$.
To keep our notation consistent in the bidirectional setting, we always use the reversed graph or $\mathit{Reversed}(G)$ if we search backwards.
Compared to the original graph~$G$, $\mathit{Reversed}(G)$ has the same set of states but with all the directed edges reversed.

We follow the standard notation in the heuristic search literature and define our search objects to be \textit{nodes} (equivalent to partial paths).
A node $x$ is a tuple that contains the main information of the partial path to state $s(x) \in S$.
The node $x$ traditionally stores a value pair $\textbf{g}(x)$ which measures the {\bf cost} of a concrete path from the initial state to state $s(x)$.
In addition, $x$ is associated with a value pair $\textbf{f}(x)$ which is an estimate of the {\bf cost} of a complete path from the initial state to the target state via $s(x)$; and also a reference $\mathit{parent}(x)$ which indicates the parent node of $x$.

We consider all operations of the boldface costs to be done element-wise.
For example, we define ${\bf g}(x) + {\bf g}(y)$ as $\left({g_1}(x) + {g_1}(y), {g_2}(x) + {g_2}(y)\right)$.
We use ($\prec,\succ)$ or ($\preceq,\succeq)$ symbols in direct comparisons of boldface values, e.g., ${\bf g}(x) \preceq {\bf g}(y)$ denotes
${g_1}(x) \leq {g_1}(y)$ and ${g_2}(x) \leq {g_2}(y)$.
Analogously, if one (or both) of the relations cannot be satisfied, we use ($\nprec,\nsucc)$ or ($\npreceq,\nsucceq)$ symbols.
For instance, ${\bf g}(x) \npreceq {\bf g}(y)$ denotes ${g_1}(x) > {g_1}(y)$ or ${g_2}(x) > {g_2}(y)$.
Unless otherwise stated, if the search is bidirectional, we assume that nodes are only compared within the same direction of the search.
This means that we do not compare a forward search node with a backward search node, even if they are associated with the same state.
We now define \textit{dominance} over nodes generated in the same search direction.
\begin{definition}
For every pair of nodes $(x,y)$ associated with the same state $s(x)=s(y)$, we say node $\mathit{y}$ is dominated by $\mathit{x}$ if we have $g_1(x) < g_1(y)$ and $g_2(x) \leq g_2(y)$ or if we have $g_1(x) = g_1(y)$ and $g_2(x) < g_2(y)$.
Node $\mathit{x}$ weakly dominates $\mathit{y}$ if ${\bf g}(x) \preceq {\bf g}(y)$.
\end{definition}
With the search dominance criteria defined, we now describe state lower and upper bounds.
\begin{definition}
For every state $u \in S$, ${\bf h}^d(u)$ and ${\bf ub}^d(u)$ denote  the lower and upper bounds on the ${\bf cost}$ of paths, respectively, from state $\mathit{u}$ to the target state in the search direction~$d$.
Node $\mathit{x}$ is a terminal node in direction~$d$ if ${\bf h}^d(s(x))= {\bf ub}^d(s(x))$.
\end{definition}
Note that ${\bf h}^d$ and ${\bf ub}^d$ can be established by conducting two simple unidirectional single-objective searches from the target state in the reverse direction~$d'$, one on $\mathit{cost_1}$ and the other one on $\mathit{cost_2}$.
We define the \textit{validity} condition as follows.
\begin{definition}
A path/node/state $\mathit{x}$ is valid if its estimated cost ${\bf f}(x)$ is within the search global upper bounds defined as $\overline{ \bf f}=(\overline{f_1},\overline{f_2})$, i.e., $\mathit{x}$ is valid if ${\bf f}(x) \preceq \overline{{\bf f}}$. Otherwise, $\mathit{x}$ is invalid if ${\bf f}(x) \npreceq \overline{{\bf f}}$, i.e., if we have $f_1(x) > \overline{f_1}$ or $f_2(x) > \overline{f_2}$.
\end{definition}
%
\subsection{Constrained Pathfinding with A*}
\label{sec:constrained_PF}
The main search in A* is guided by the $\mathit{start}$-$\mathit{goal}$ cost estimates or {\bf f}-values.
These {\bf f}-values are traditionally established based on a consistent and admissible heuristic function ${\bf h}: S \rightarrow \mathbb{R}^+ \times \mathbb{R}^+$ \cite{hart1968formal}.
In other words, for every search node $x$, we have ${\bf f}(x)={\bf g}(x)+{\bf h}(s(x))$ where ${\bf h}(s(x))$ estimates lower bounds on the ${\bf cost}$ of paths from state $s(x)$ to the target state.
\begin{definition}
The heuristic function $h_p$ is admissible iff $h_p(u) \le \mathit{cost}_p(\pi)$ for every $u \in S$ where $\pi$ is the optimal path on $\mathit{cost}_p$ from state $\mathit{u}$ to the target state. 
It is also consistent if we have ${ h_p}(u) \le { cost}_p(u,v) + {h_p}(v)$ for every edge $(u,v) \in E$ \cite{hart1968formal}. 
\end{definition}
We distinguish forward and backward heuristic functions by incorporating the search direction~$d$, i.e., ${\bf h}^d$.
In A*, we perform a systematic search by \textit{expanding} nodes in best-first order.
That is, the search is led by a partial path that shows the lowest cost estimate.
To this end, we expand one (lexicographically) least-cost node in each iteration and store all the descendant (new) nodes in an $\mathit{Open}$ list.
More accurately, $\mathit{Open}^d$ is a priority queue for the A* search of direction~$d$ that contains generated (but not expanded) nodes.
To commence the search, we initialise the $\mathit{Open}^d$ list with a node associated with the initial state and ${\bf g}=(0,0)$.
For the purpose of further expansion, the $\mathit{Open}^d$ list reorders its nodes according to their $\mathit{f}$-values such that the least-cost node is at the front of the list.
In the constrained setting, since {\bf f} represents a pair of costs, there are two possible lexicographic orderings of the nodes in the $\mathit{Open}^d$ list.
Therefore, depending on the search requirement, the $\mathit{Open}^d$ list can order nodes based on $(f_1,f_2)$ or $(f_2,f_1)$ lexicographically.
For example, if a search method needs unexplored nodes to be lexicographically ordered on $(f_1,f_2)$, the $\mathit{Open}^d$ list first orders nodes based on their $f_1$-value, and then based on their $f_2$-value if two (or more) of them have the same $f_1$-value.
The latter operation (ordering based on the second element) is called \textit{tie-breaking}.
Nodes associated with the target state represent solution paths, thus A* does not need to expand such solution nodes.
Finally, A* terminates if the {\bf cost}-optimal solution path is found, or if there is no node in the $\mathit{Open}^d$ list to expand.

Section~\ref{sec:A_star_correctness} in Supplementary Material (SM) revisits the principles of constrained search with A* and discusses in detail the correctness of each component of the search, including (lazy) dominance rules, the termination criteria and optimality of the solution.
In the next parts, we present the search setup and the pruning strategies as part of our common methods.
%
\subsection{Search Setup}
The first common method in our A* searches is Setup($d$) in Procedure~\ref{alg:rc_ebba_setup}.
This procedure shows the essential data structures we initialise for our A*-based constrained search in direction $d \in $ \{\textit{forward}, \textit{backward}\}.
The procedure first initialises $\mathit{Open}^d$ as the priority queue of search.
It then initialises for every $u \in S$ the scalar $g^d_{\mathit{min}}(u)$, an important parameter that will keep track of the secondary cost of the last node successfully expanded for state $\mathit{u}$ during the search in direction~$d$.
A* uses this parameter to prune some dominated nodes.
Depending on the search direction, the procedure sets the initial state $u_i$.
If the search direction is \textit{forward}, $u_i$ is chosen to be $\mathit{start}$, otherwise, $\mathit{goal}$ is chosen as the initial state.
To commence the search, the procedure generates a new node $x$ with the initial state $u_i$ and inserts it into $\mathit{Open}^d$.
Node $x$, as the initial node, has zero actual costs (i.e., $\mathit{g}$-values) and a null pointer as its parent node, but it can use the heuristic functions ${\bf h}^d$ to establish its cost estimates, i.e., $\mathit{f}$-values.

In addition, if we want to run a constrained search via {WC-EBBA*} or {WC-EBBA*}\textsubscript{par}, the procedure initialises $\chi^d(u)$ as an empty list for every $u \in S$.
As we will see later in the paper, this list will be populated with expanded nodes (partial paths) of state $\mathit{u}$ during the search in direction~$d$.
{WC-EBBA*}\textsubscript{par} (and similarly {WC-EBBA*}) will use this list to offer complementary paths to nodes that are under exploration for state $\mathit{u}$ in the opposite direction~$d'$ (as part of partial paths matching).
\subsection{Path Expansion with Pruning}
Expanding partial paths is a key component in A*.
Procedure~\ref{alg:expansion} shows the main steps involved in Expanding and Pruning (ExP) a typical node $x$ in the traditional ($f_1,f_2$) order and in direction~$d$.
The ExP($x,d$) procedure \textit{generates} a set of new descendant nodes via $s(x)$'s successors, i.e., $\mathit{Succ}(s(x))$, and then checks new nodes against the pruning criteria.
Each successor state is denoted by $v$.
Lines~\ref{alg:expansion:2}-\ref{alg:expansion:5} of Procedure~\ref{alg:expansion} show the essential operations involved in the node expansion.
Given the partial path information carried by the current node $x$, each new node $y$ is initialised with the successor state $v$, actual costs ${\bf g} (y)$ and cost estimates ${\bf f} (y)$ of the extended path, capturing $x$ as the node $y$'s parent.
Procedure~\ref{alg:expansion} also proposes three types of pruning strategies before inserting new nodes into $\mathit{Open}^d$.
We briefly explain their type and functionality below.
\begin{itemize}[nosep]
\item Line~\ref{alg:expansion:prune1}: Prune by dominance; ignore the new node $y$ if it is dominated by the last non-dominated node expanded for $s(y)$. 
\item Line~\ref{alg:expansion:prune2}: Prune by dominance; ignore node $y$ if it is dominated by one of the preliminary shortest paths to $s(y)$.
\item Line~\ref{alg:expansion:prune3}: Prune by invalidity; ignore node $y$ if it shows cost estimates beyond the search global upper bounds in $\overline{\bf f}$.
\end{itemize}
Note that the pruning strategy employed in Line~\ref{alg:expansion:prune2} will only be used in bidirectional searches, where search in direction~$d$ has access to states' upper bound function ${\bf ub}^{d'}$.
Finally, the ExP($x,d$) procedure inserts every new node $y$ into $\mathit{Open}^d$ if $y$ is not pruned by dominance or validity tests.
%
\noindent
\begin{figure}[!t]
{
\begin{minipage}[t]{.29\textwidth}
\null 
\input{Algorithms/RC_EBBA_Setup}
\end{minipage}%
\hspace{2mm}
\begin{minipage}[t]{.34\textwidth}
\null 
\input{Algorithms/Expand}
\end{minipage}%
\hspace{2mm}
\begin{minipage}[t]{.34\textwidth}
\null
\input{Algorithms/Early_sol }
\end{minipage}
}
\end{figure}
\subsection{Early Solution Update}
The Early Solution Update (ESU) strategy allows our algorithms to update the global upper bounds and possibly establish a feasible solution $\mathit{Sol}$ before reaching the target state.
Here, we choose the ESU procedure we developed for bi-criteria search in \citet{AhmadiTHK21_esa} and adapt it to the WCSPP. 
The idea of this strategy is quite straightforward and is shown in Procedure~\ref{alg:early_sol} for the traditional objective ordering $(f_1,f_2)$ in the search direction~$d$.
We briefly describe the procedure as follows.

Consider node $x$ as a partial path, the procedure ESU($x,d$) tries to establish a complete $\mathit{start}$-$\mathit{goal}$ path by joining $x$ to its two complementary shortest paths from $s(x)$ to the target state.
In our notation, we use ($\overline{f_1},f^{sol}_2$) to keep track of the actual costs of the best-known solution path during the search.
The procedure tries two cases, as explained below.
\begin{enumerate}
\item
It first joins node $x$ with its complementary shortest path for the primary attribute, i.e., the path with costs $(h^d_1(s(x)), \mathit{ub}^d_2(s(x)))$ in the ($f_1,f_2$) order.
It then retrieves the actual costs of the joined path via $f_1(x)=g_1(x)+h^d_1(s(x))$ and $f'_2 = g_2(x)+\mathit{ub}^d_2(s(x))$.
If the resulting joined path is valid, the procedure treats $x$ as a tentative solution node.
In this case, if the tentative solution is lexicographically smaller than the best-known solution, the procedure updates the global upper bound $\overline{f_1}$ with $f_1(x)$.
This is followed by storing $f'_2$ in $f^{sol}_2$ as the secondary cost of the new solution and also capturing $x$ as a new solution node via $\mathit{Sol}$.
\item 
If the joined path in the first part is found invalid, the procedure still has a chance to improve its primal upper bound using the shortest path optimum for the second criterion, with costs $(\mathit{ub}^d_1(s(x)),h^d_2(s(x)))$.
In this case, the costs of the joined path can be retrieved as $f_2(x)=g_2(x)+h^d_2(s(x))$ and $f'_1 = g_1(x)+\mathit{ub}^d_1(s(x))$.
If the resulting joined path is valid, the procedure updates the primary upper bound $\overline{f_1}$ with $f'_1$ and then resets $f^{sol}_2$ for the upcoming solution (with cost $f'_1$ or better).
\end{enumerate}
Note that the procedure does not need to explicitly check $f_1(x)$ and $f_2(x)$.
This is mainly because our constrained A* searches prune invalid nodes before attempting the ESU strategy.
In addition, if $x$ enters Procedure~\ref{alg:early_sol} with the target state, it will definitely pass case 1 above and will be captured as a valid solution node.
This is because we always have ${\bf h}^d(s(x))={\bf ub}^d(s(x))$ for such nodes, which immediately yields $f'_s=f_s(x)$ and a valid path consequently.
There is also one important difference between the two cases: valid joined paths compute a potential solution only in case~1. (see Lemma~\ref{lemma:early_sol} for the formal proof).

\textbf{Solution path construction:}
If the search terminates with $x$ as an optimal solution node with a non-target state, we just need to join $x$ with its shortest path on the primary cost ($\mathit{cost_p}$) for solution path construction. 
Therefore, solution path recovery with the ESU strategy is a two-stage procedure: first, we follow back-pointers from $\mathit{Sol}$ to the initial state and build the partial path. Second, we recover the concrete path from the $\mathit{Sol}$'s state to the target state (via the pre-established initial shortest path on $\mathit{cost_p}$) and then concatenate this complementary path to the partial path obtained in the first stage.
%
\section{Constrained Pathfinding with {WC-A*}}
\label{sec:WC_A}
The Bi-Objective Shortest Path Problem (BOSPP) and WCSPP are two inextricably linked problems, with often shared procedures and terminologies.
Compared to the WCSPP where we look for a single optimal path, BOSPP aims to find a representative set of Pareto-optimal solution paths, i.e., a set in which every individual solution offers a path that minimises the bi-criteria problem in both \textit{cost} and \textit{weight}.
More accurately, given ($\mathit{cost}_1,\mathit{cost}_2$) as our edge attributes, Pareto optimality is a situation where we cannot improve $\mathit{cost}_2$ of point-to-point paths without worsening their $\mathit{cost}_1$ and vice versa.
It is not difficult to show that any method that generates all Pareto-optimal solutions to the BOSPP is also able to deliver an optimal solution to the WCSPP.
From the WCSPP's point of view, if a {\bf cost}-optimal solution path exists, it can always be found in a Pareto-optimal set of BOSPP.
With this introduction, we now explain how our new A*-based weight constrained search algorithm {WC-A*} can be derived from the recent Bi-objective A* algorithm (BOA*) \cite{ulloa2020simple}, a unidirectional search algorithm to find a set of cost-unique Pareto-optimal paths.

{WC-A*} is a simple solution approach specialised to solve the WCSPP and follows the search strategy of BOA* and its improvements in \citet{AhmadiTHK21_esa}.
{WC-A*} works on the basis of A* and explores the graph in the forward direction (from $\mathit{start}$ to $\mathit{goal}$) in the traditional ($f_1,f_2$) order.
Algorithm~\ref{alg:rca_high} shows the high-level design of {WC-A*} and its two essential phases.
Similar to other A*-based algorithms, {WC-A*} needs to establish its heuristic functions via an initialisation phase.
However, since the main constrained search is unidirectional, {WC-A*} needs to compute lower and upper bound functions in only one direction.
Therefore, the initialisation phase consists of two single-objective backward searches.
When the preliminary heuristic searches are complete, {WC-A*} initialises a node pair $\mathit{Sol}$ with the secondary cost $f^{sol}_2$ to keep track of the lexicographically smallest solution node during the main search.
In the next phase, {WC-A*} executes a forward constrained A* search in the ($f_1,f_2$) order.
When the second phase is also complete, the algorithm returns the node pair $\mathit{Sol}$, which corresponds to a {\bf cost}-optimal solution.
Note that since {WC-A*} is a unidirectional search algorithm, there is no backward counterpart.
Clearly, there is no feasible solution if the returned node pair is empty.
We illustrate operations in each phase of {WC-A*} as follows.

\begin{algorithm}[!t]
\small
\caption{WC-A* High-level}
\label{alg:rca_high}
  \DontPrintSemicolon 
    \SetKwBlock{DoParallel}{do in parallel}{end}
    \KwIn{A problem instance (G, {\bf cost}, $\mathit{start}$, $\mathit{goal}$) with the weight limit $W$}
    \KwOut{A node pair corresponding with the cost-optimal feasible solution $Sol$}
     ${\bf h}, {\bf ub}$, ${\bf \overline{f}}, S' \leftarrow$   Initialise WC-A* (G, {\bf cost}, $\mathit{start}$, $\mathit{goal}, W$) \Comment*[r]{Algorithm~\ref{alg:Uni_search_init}}
    $Sol \leftarrow (\varnothing,\varnothing)$, $f^{sol}_2 \leftarrow \infty$ \;
    $Sol \leftarrow$ Run a WC-A* search on (G, {\bf cost}, $\mathit{start}$, $\mathit{goal}$) in the forward direction and ($f_1,f_2$) order with global upper bounds ${\bf \overline{f}}$,\;
    \nonl \qquad \qquad heuristic functions $({\bf h}, {\bf ub})$ and initial solution $Sol$ with secondary cost $f^{sol}_2$. \Comment*[r]{Algorithm~\ref{alg:rc_ba}}
\Return{$Sol$}
\end{algorithm}

%
\subsection{Initialisation}
{WC-A*} is a unidirectional search algorithm and only requires heuristics in one direction.
Algorithm~\ref{alg:Uni_search_init} shows the main steps involved in {WC-A*}'s initialisation phase.
It starts with initialising the search global upper bounds, namely $\overline{f}$-values.
The secondary global upper bound $\overline{f_2}$ is set to be the weight limit $W$, while the primary upper bound $\overline{f_1}$ is initially unknown and will be determined by one of the heuristic searches.
The main constrained search will be run in the forward direction, so we obtain the required heuristics $h^f_1$ and $h^f_2$ by running two cost-bounded backward A* searches.
The first unidirectional A* search finds lower bounds on $\mathit{cost_2}$ and stops before expanding a state with an estimated cost larger than the global upper bound~$\overline{f_2}$.
The first search on $\mathit{cost_2}$ is also capable of initialising the global upper bound $\overline{f_1}$ using the upper bound obtained for the $\mathit{start}$ state, i.e., $\mathit{ub}^f_1(\mathit{start})$.
When the first cost-bounded A* terminates, all not-yet-expanded states in this preliminary search are guaranteed not to be part of any solution paths, reducing the graph size to obtain better-quality heuristics.
This method is known as \textit{resource-based network reduction} \cite{AnejaAN83}.
To this end, our cost-bounded A* search on $\mathit{cost_1}$ will only consider the states expanded in the first search and stops before expanding a state with a cost estimate larger than $\overline{f_1}$.

The fact that {WC-A*} only needs two cost-bounded searches to establish its required heuristic functions can be seen as {WC-A*}'s strength; easy problems will especially benefit from a speedy search setup.
However, we can also see this simple initialisation phase as a weakness, mainly because {WC-A*} will have a limited capability to reduce the graph size.
Compared to the bidirectional algorithms where both attributes effectively reduce the graph size in two rounds, as in {WC-EBBA*} and {WC-BA*}, {WC-A*} benefits from this effective feature only in one round of bounded searches.
In addition, there may be cases where starting the bounded searches with $\mathit{cost_2}$ results in limited graph reduction, mainly due to a lack of informed heuristics for $\mathit{cost_2}$.
We will investigate the impact of this possible search ordering on {WC-A*}'s performance in our extended empirical study.

\begin{algorithm}[t]
\small
\caption{Initialisation phase of WC-A*}
\label{alg:Uni_search_init}
\DontPrintSemicolon
    \SetKwBlock{DoParallel}{do in parallel}{end}
    \KwIn{The problem instance (G, {\bf cost}, $\mathit{start}$, $\mathit{goal}$) and the weight limit $W$}
    \KwOut{WCSPP's uni-directional heuristic functions ${\bf h}$ and ${\bf ub}$, also global upper bounds ${\bf \overline{f}}$}
    
        Set global upper bounds: $\overline{f_1} \leftarrow \infty$ and $\overline{f_2} \leftarrow W $ \;
    
      $ h^f_2, ub^f_1 \leftarrow$ Run $f_2$-bounded backward A* (from $\mathit{goal}$ using an admissible heuristic) on $cost_2$, use $cost_1$ as a tie-breaker,\;
      \nonl \qquad \qquad update $\overline{f_1}$ with $ub^f_1(\mathit{start})$ when $\mathit{start}$ is going to get expanded
      and stop before expanding a state with $f_2 > \overline{f_2}$.
      \label{alg:Uni_search_init:init_1} \;
      $h^f_1, ub^f_2 \leftarrow$ Run $f_1$-bounded backward A* (using an admissible heuristic) on $cost_1$, use $cost_2$ as a tie-breaker, ignore unexplored\; 
      \nonl \qquad \qquad states in the search of line~\ref{alg:Uni_search_init:init_1},
      and stop before expanding a state with $f_1 > \overline{f_1}$. \label{alg:Uni_search_init:init_2}\;
$S' \leftarrow$ Non-dominated states explored in the last bounded A* search of line \ref{alg:Uni_search_init:init_2} \;  
\Return{${\bf h}^f, {\bf ub}^f$, ${\bf \overline{f}}$, $S'$}
\end{algorithm}

%
\subsection{Constrained Search}
The constrained search in the second phase of {WC-A*} is relatively straightforward and involves in general two types of strategies: expand and prune.
Algorithm~\ref{alg:rc_ba} shows a pseudocode of {WC-A*} in the generic search direction~$d$ and in objective ordering ($f_p,f_s$).
In this notation, $f_p$ and $f_s$ denote \textit{primary} and \textit{secondary} cost of the search, respectively, generalising the two possible objective orderings ($f_1,f_2$) or ($f_2,f_1$).
Although this paper studies {WC-A*} in the standard setting of BOA*, we intentionally present {WC-A*} in a generic form to emphasise its capability in solving the WCSPP from both directions and also in both objective orderings.
One such application of {WC-A*} is in the parallel searches of {WC-BA*} (see \ref{sec:WC_BA} for more details). 
For now, let {WC-A*} be a forward constrained A* search in the ($f_1,f_2$) order, i.e, we have $d=$~\textit{forward} and ($p,s$)=($1,2$).
In addition, for the operations given in Algorithm~\ref{alg:rc_ba}, we focus on those applicable to {WC-A*} (lines in black without \textsuperscript{+} or \textsuperscript{*} symbols).

\textbf{Algorithm description:}
{WC-A*} in Algorithm~\ref{alg:rc_ba} employs Setup($d$) (Procedure~\ref{alg:rc_ebba_setup}) to initialise data structures required by the search in the forward direction.
This initialisation involves inserting a node with $\mathit{start}$ state in the priority queue to commence the constrained search.
Let $\mathit{Open}^f$ be a (non-empty) priority queue of the forward search in any arbitrary iteration of the algorithm.
{WC-A*} extracts a node $x$ from the priority queue with the lexicographically smallest ($f_1,f_2$) among all nodes in $\mathit{Open}^f$.
Note that A* essentially needs $x$ to be the least-$\mathit{cost}_p$ node, however, lexicographical ordering of nodes in $\mathit{Open}^f$ will help {WC-A*} to prune dominated nodes.
If $f_1(x)$ is out-of-bounds (line~\ref{alg:rc_ba:termination}), A* guarantees that all not-yet-expanded nodes would be out-of-bounds and the search can terminate (see Lemma~\ref{lemma:astar_terminate}).
Otherwise, the algorithm considers $x$ as a valid node and checks it for dominance at line~\ref{alg:rc_ba:prune2}.
If $x$ is not weakly dominated by the last node expanded for $s(x)$, {WC-A*} will then explore $x$ and update $g^f_{\mathit{min}}(s(x))$ with the secondary (actual) cost $g_2(x)$ at line~\ref{alg:rc_ba:min_r} of Algorithm~\ref{alg:rc_ba}.
In the next step (line~\ref{alg:rc_ba:early_sol}), node~$x$ is matched with the complementary shortest paths from $s(x)$ to the $\mathit{goal}$ state to possibly obtain a tentative solution or update the primary upper bound $\overline{f_1}$ before reaching $\mathit{goal}$.
This matching is done via the ESU procedure in the ($f_1$,$f_2$) order (Procedure~\ref{alg:early_sol}).
At this point, the ESU strategy guarantees that $x$ has been captured as a tentative solution if the joined $\mathit{start}$-$\mathit{goal}$ path via $x$ is lexicographically smaller in the ($\mathit{cost_1},\mathit{cost_2}$) order than the current solution in $\mathit{Sol}$ with costs ($\overline{f_1},f^{sol}_2$).
However, {WC-A*} can still skip expanding node $x$ if it is a terminal node via line~\ref{alg:rc_ba:early_sol}, i.e., if $h^f_1=\mathit{ub}^f_1$.
This is because terminal nodes are tentative solution nodes (already captured by the ESU strategy) and thus their expansion is not necessary (see Lemma~\ref{lemma:terminal_node} for the formal proof).
Finally, $x$ will be expanded via the ExP($x,d$) procedure in the forward direction (Procedure~\ref{alg:expansion}) if it is neither a dominated nor terminal node.
%
\begin{figure}[!t]
\noindent
\begin{minipage}[t]{.49\textwidth}
\null 
\input{Algorithms/RC_BA_Main}
\end{minipage}%
\hspace{2mm}
\begin{minipage}[t]{.49\textwidth}
\null
\input{Algorithms/Early_sol2}
\input{Algorithms/Expand2}
\end{minipage}
\end{figure}
%
\noindent
\begin{figure}[!t]
\begin{subfigure}{0.49\textwidth}
\begin{tikzpicture}[
roundnode/.style={circle, draw=black,  thick, minimum size=5mm},
roundnode2/.style={circle, draw=gray,  thick, minimum size=5mm},
scale=.8, every node/.style={scale=.9}]
\footnotesize
\node[roundnode,align=center]   (s2)        {$u_2$ \\ (2,1)\\ (2,1)};
\node[roundnode,align=center]   at (0, 2.75) (start)     {$u_{s}$\\ (3,3)\\ (7,8) };
\node[roundnode,align=center]   at (-3.25, 0) (s1)        {$u_1$\\ (2,3)\\ (3,4)};
\node[roundnode,align=center]   at (3.25, 0) (s3)        {$u_3$\\ (3,2)\\ (4,3)};
\node[roundnode,align=center]   at (0, -2.75) (goal)   {$u_{g}$\\ (0,0)\\ (0,0)};
\node[roundnode2,align=center,color=gray]   (state)     [left=of goal] {state\\ ${\bf h}^f$\\ ${\bf ub}^f$ };

\draw[->,-latex,  thick] (start) edge[auto=right] node{(1,4)} (s1);
\draw[->,-latex,  thick] (start) edge[auto=left] node{(3,4)} (s2);
\draw[->,-latex,  thick] (start) edge[auto=left] node{(3,1)} (s3);
\draw[->,-latex,  thick] (s1) edge[auto=left] node{(1,2)} (s2);
\draw[->,-latex,  thick] (s3) edge[auto=right] node{(2,1)} (s2);
\draw[->,-latex,  thick] (s1) edge[auto=left] node{(2,4)} (goal);
\draw[->,-latex,  thick] (s2) edge[auto=left] node{(2,1)} (goal);
\draw[->,-latex,  thick] (s3) edge[auto=left] node{(3,3)} (goal);

\end{tikzpicture}
\end{subfigure}
\begin{subfigure}{0.49\textwidth}
\centering
\footnotesize
\renewcommand{\arraystretch}{1}
\begin{tabular}{ c   l   l  c  c c}

    \toprule
     & \textit{Open} list \\
    It. & [${\bf f}(x), {\bf g}(x), s(x)$] & Nodes pruned & $\overline{f_1}$ & $f^{sol}_2$\\
    \midrule
    1 & $\uparrow$[(3,3), (0,0), $u_{s}$] & [(3,7), (1,4), $u_{1}$] & 7  & $\infty$ \\
   \midrule 
    2 & $\uparrow$[(5,5), (3,4), $u_2$] & & 5 & 5 \\
    & \ \ [(6,3), (3,1), $u_3$]  &  & & \\
    \midrule
    3 & $\uparrow$[(6,3), (3,1), $u_3$] &  & 5 & 5 \\
    \bottomrule
    
\end{tabular}
\vspace*{0.5 cm}
\setlength{\tabcolsep}{4.5pt}
\begin{tabular}{|l| *{5}{c|} }
    \toprule
    parent arrays & $u_s$ & $u_1$ & $u_2$ & $u_3$ & $u_g$\\
    \midrule
    \texttt{parent\_state} & [$\varnothing$] & & [$u_s$] & & \\
    \midrule
    \texttt{parent\_path\_id}& [0] & & [1] & & \\
    \bottomrule
\end{tabular}
\end{subfigure}
\caption{\small Left: An example graph with \textbf{cost} on the edges, and with (state identifier, \textbf{h}, \textbf{ub}) inside the states. Right: Status of {WC-A*} in every iteration (It.): nodes in the $\mathit{Open}$ list at the beginning of each iteration, nodes pruned by the algorithm in each iteration, the (updated) value of the global upper bound $\overline{f_1}$ and also the secondary cost of the best-known solution $f^{sol}_2$ at the end of each iteration. The search is conducted in the forward direction and in the $(f_1,f_2)$ order. The symbol $\uparrow$ beside nodes denotes the expanded min-cost node in the iteration. The second table shows the status of the parent arrays of the states when the search terminates (see Section~\ref{sec:practicall} for details).}
\label{fig:example}
\end{figure}

\textbf{Example:}
We explain the constrained search of {WC-A*} by solving a WCSPP for the graph depicted in Figure~\ref{fig:example}.
On this graph, we want to find the {\bf cost}-optimal shortest path between states $u_s$ and $u_g$ with the weight limit $W=6$.
Let us assume the initialisation phase has already been completed, and we have both lower and upper bounds from all states to $u_g$ (our target state) via two single objective backward searches.
For the states in Figure~\ref{fig:example}, we have shown ${\bf h}^f$ and ${\bf ub}^f$ as part of the states' information, and have updated the primary global upper bound with $\mathit{ub}^f_1(u_s)$ by setting $\overline{f_1}=7$ (via our initial solution).
Nonetheless, we can see that the graph size has not changed as all the lower bounds (${\bf h}$-values) are within the search global upper bounds.
Figure~\ref{fig:example} also shows a summary of changes in each iteration of {WC-A*}.
In particular: the status of the $\mathit{Open}$ list at the beginning of each iteration, nodes pruned in the iteration and also the latest value of solution costs ($\overline{f_1},f^{sol}_2$) at the end of each iteration.
We explain in three iterations below how {WC-A*} finds a {\bf cost}-optimal solution path with just one node expansion.
\begin{itemize}[nosep]
  \item \textbf{Iteration 1:} At the beginning of this iteration, the search sees only one node in $\mathit{Open}$ associated with the start state $\mathit{u_s}$. We extract this node as $x \leftarrow [(3,3), (0,0), u_{s}]$ and interpret the node's information as $[(f_1,f_2), (g_1,g_2), \mathit{state}]$. The algorithm then checks $x$ against the termination criterion and also the dominance test. $x$ is a valid and non-dominated node, so {WC-A*} updates $g^d_{\mathit{min}}(u_s) \leftarrow 0$ and proceeds with the ESU strategy.
  In this procedure, we can see that the shortest path from $u_s$ to $u_g$ on $\mathit{cost_1}$ is invalid.
    In addition, the shortest path on $\mathit{cost_2}$ does not offer a valid path better than the initial solution.
    So the procedure is unable to improve the upper bound in this iteration.
    $x$ is not a terminal node either, so {WC-A*} expands this node and generates new nodes when it discovers adjacent states $u_1$, $u_2$ and $u_3$.
    Before inserting new nodes into the priority queue, the pruning criteria indicate that the new node associated with $u_1$ is invalid and should be pruned, essentially because its secondary cost estimate is out-of-bounds ($f_2>\overline{f_2}$ or $7 > 6$).
    However, the two other nodes generated for states $u_2$ and $u_3$ will be added to the $\mathit{Open}$ list.
    \item \textbf{Iteration 2:} There are two nodes in the priority queue. {WC-A*} extracts the node associated with $\mathit{u_2}$ as it is lexicographically smaller than the other node in the queue (in the ($f_1,f_2$) order). Let $x \leftarrow [(5,5), (3,4), u_2]$ be the extracted node. This node does not meet the termination criterion and is not a dominated node, mainly because this is the first time {WC-A*} visits state $u_2$, so it updates $g^d_{\mathit{min}}(u_2) \leftarrow 4$.
    The algorithm then checks $x$ against the ESU strategy.
    In the ESU($x,d$) procedure, we realise that joining $x$ with its complementary shortest path on $\mathit{cost_1}$ yields a valid path. For this complete joined path, i.e., path $\{u_s,u_2,u_g\}$, we have ${\bf cost}=(5,5)$.
    So we can see that the joined path is valid because its primary cost is smaller than the best-known cost ($f_1(x)< \overline{f_1}$ or $5<7$).
    Therefore, the ESU strategy captures $x$ as a tentative solution node in $\mathit{Sol}$ and updates $\overline{f_1} \leftarrow 5$ and $f^{sol}_2 \leftarrow 5$.
    $x$ is also a terminal node, since we have $h^f_1(u_2)=\mathit{ub}^f_2(u_2)=2$.
    Hence, {WC-A*} does not need to expand $x$ considering the fact that the tentative solution via $u_2$ is already stored in $\mathit{Sol}$.
    \item \textbf{Iteration 3:} {WC-A*} did not insert any new node in $\mathit{Open}$ in the previous iteration, so there exists only one unexplored node in the priority queue.
    The algorithm extracts this node as $x \leftarrow [(6,3), (3,1), u_3]$. 
    The extracted node $x$ is then checked against the termination criterion, similar to other extracted nodes.
    Now, we can see that $x$ is an invalid node since its $f_1$-value is no longer within the bound.
    In other words, we have $f_1(x) > \overline{f_1}$ or $6>5$ equivalently.
    Therefore, {WC-A*} successfully terminates with a {\bf cost}-optimal solution node in $\mathit{Sol}$ with costs $(5,5)$.
\end{itemize}
We now prove the correctness of constrained pathfinding with {WC-A*} (see \ref{sec:WC_A_NoTieBreak} for the correctness without tie-breaking).
\begin{theorem}
{WC-A*} returns a node corresponding to a {\bf cost}-optimal solution path for the WCSPP.
\end{theorem}
\begin{proof}
{WC-A*} enumerates all valid partial paths from the initial state towards the target state in search of an optimal solution.
Our proposed pruning strategies ensure that {WC-A*} never removes a solution node from the search space (Lemmas~\ref{lemma:invalid}~and~\ref{lemma:dominated}).
In addition, the ESU strategy correctly keeps track of the lexicographically smallest tentative solutions discovered during the search (Lemma~\ref{lemma:early_sol}), while avoiding unnecessary expansion of terminal nodes (Lemma~\ref{lemma:terminal_node}). 
Therefore, we conclude that {WC-A*} terminates with a {\bf cost}-optimal solution, even with zero-weight cycles (Lemma~\ref{lemma:astar_terminate}).
\end{proof}
%
\section{Constrained Pathfinding with Parallel {WC-EBBA*}}
\label{sec:WC_EBBA_par}
The Enhanced Biased Bidirectional A* search algorithm for the WCSPP, or {WC-EBBA*} \cite{AhmadiTHK21}, is constructed based on the {RC-BDA*} algorithm of \citet{thomas2019exact}, both following the standard bidirectional search first presented by \citet{pohl1971bi}.
{WC-EBBA*} uses separate priority queues for its forward and backward searches, namely $\mathit{Open}^f$ and $\mathit{Open}^b$, and only explores the graph in one direction at a time.
Depending on the location of the least cost node in $\mathit{Open}^f \cup \mathit{Open}^b$, we have either $d=$~\textit{forward} or $d=$~\textit{backward}.
For example, if the least-cost node in $\mathit{Open}^f$ is smaller than the least-cost node in $\mathit{Open}^b$, 
{WC-EBBA*} attempts a forward expansion.
In this interleaved search scheme, there might be cases where the algorithm mostly performs long runs of uni-directional expansions, a likely case when the search frontier of one direction reaches a dense part of the graph.
Although {WC-EBBA*} tries to balance bidirectional search effort by changing the weight budget of each direction, there is always a chance for slower directions to become dominant with this type of queuing strategy.
In addition, {WC-EBBA*} still suffers from a lengthy initialisation phase, even with its more efficient bounded searches.
Faster than the traditional one-to-all approaches, the initialisation phase of {WC-EBBA*} obtains its necessary heuristics via four bounded A* searches in series.
However, there might be cases where the algorithm spends more time in its initialisation phase than the actual constrained search, especially in easy instances with loose weight constraints.
Section \ref{sec:WC_EBBA} revisits the standard {WC-EBBA*} algorithm, detailing all key steps.
 
In this section, we present a new variant of the {WC-EBBA*} algorithm that leverages parallelism to overcome the aforementioned shortcomings.
Algorithm~\ref{alg:rcebba_high_par} shows, at a high level, three main steps for our new parallel version {WC-EBBA*}\textsubscript{par}.
Similar to its standard variant, {WC-EBBA*}\textsubscript{par} obtains its required heuristics and also global upper bounds in the first place via two rounds of parallel searches.
It then uses a set of non-dominated states obtained from the initialisation phase to determine the forward and backward budget factors $\beta^f$ and $\beta^b$.
In the next step, the search initialises a (shared) solution node pair $\mathit{Sol}$ with an unknown secondary cost $f^{sol}_2$ to keep track of the best solution obtained in both directions.
In the last step, the algorithm runs two biased {WC-EBBA*}\textsubscript{par} searches concurrently, each capable of updating shared parameters, such as the global upper bound $ \overline{f_1}$ and the solution node pair $\mathit{Sol}$.
Comparing the structure of both algorithms at this high level, we can see that {WC-EBBA*}\textsubscript{par} differs from {WC-EBBA*} in two aspects: initialisation and search structure.
We will explain each of these in detail below.



\begin{algorithm}[!t]
\small
\caption{WC-EBBA*\textsubscript{par} High-level}
\label{alg:rcebba_high_par}
  \DontPrintSemicolon 
    \SetKwBlock{DoParallel}{do in parallel}{end}
    \KwIn{A problem instance (G, {\bf cost}, $\mathit{start}$, $\mathit{goal}$) with the weight limit $W$}
    \KwOut{A node pair corresponding with the cost-optimal feasible solution $Sol$}
     ${\bf h}, {\bf ub}, {\bf \overline{f}}, S' \leftarrow$   Initialise WC-EBBA*\textsubscript{par} (G, {\bf cost}, $\mathit{start}$, $\mathit{goal}, W$) \Comment*[r]{Algorithm~\ref{alg:Bi_search_init}}
  $\beta^f , \beta^b \leftarrow$ Obtain forward and backward budget factors using $h_1$-values of non-dominated states in $S'$. \Comment*[r]{Eq.(\ref{bias_factor})}
  $Sol \leftarrow (\varnothing,\varnothing)$, $f^{sol}_2 \leftarrow \infty$ \;
    \DoParallel{
        Run a biased WC-EBBA*\textsubscript{par} search for (G, {\bf cost}, ${start}$, ${goal}$) in the \textit{forward} direction with global upper bounds ${\bf \overline{f}}$, \;
    \nonl \ heuristic functions $({\bf h}, {\bf ub})$, budget factors ($\beta^f,\beta^b$) and initial solution $Sol$ with secondary cost $f^{sol}_2$. \Comment*[r]{Algorithm~\ref{alg:rc_ebba_par}}
        
        Run a biased WC-EBBA*\textsubscript{par} search for (G, {\bf cost}, ${start}$, ${goal}$) in the \textit{backward} direction with global upper bounds ${\bf \overline{f}}$, \;
    \nonl \ heuristic functions $({\bf h}, {\bf ub})$, budget factors ($\beta^f,\beta^b$) and initial solution $Sol$ with secondary cost $f^{sol}_2$. \Comment*[r]{Algorithm~\ref{alg:rc_ebba_par}}
        
    }
\Return$Sol$
\end{algorithm}


\subsection{Initialisation}
\label{sec:init_rcebba_par}
Our bidirectional {WC-EBBA*}\textsubscript{par} requires both forward and backward lower/upper bound functions.
To speed up the preliminary searches, following {WC-BA*}, we compute the necessary functions in two rounds of parallel searches as shown in Algorithm~\ref{alg:Bi_search_init}.

\begin{algorithm}[!t]
\small
\caption{Initialisation phase of WC-EBBA*\textsubscript{par} and  WC-BA*}
\label{alg:Bi_search_init}
  \DontPrintSemicolon 
    \SetKwBlock{DoParallel}{do in parallel}{end}
    \KwIn{The problem instance (G, {\bf cost}, $\mathit{start}$, $\mathit{goal}$) with the weight limit $W$}
    \KwOut{WCSPP's bidirectional heuristic functions ${\bf h}$ and ${\bf ub}$, global upper bounds ${\bf \overline{f}}$, also non-dominated states $S'$}
    Set global upper bounds: $\overline{f_1} \leftarrow \infty$ and 
    $\overline{f_2} \leftarrow W $ \;
    \DoParallel{
      $ h^f_2, ub^f_1 \leftarrow$ Run $f_2$-bounded backward A* (from $\mathit{goal}$ using an admissible heuristic) on $cost_2$, use $cost_1$ as a tie-breaker,\;
      \nonl \qquad \qquad update $\overline{f_1}$ with $ub^f_1(\mathit{start})$ when $\mathit{start}$ is going to get expanded
      and stop before expanding a state with $f_2 > \overline{f_2}$.
      \label{alg:Bi_search_init:init_1} \;
      
      $h^b_1, ub^b_2 \leftarrow$ Run $f_1$-bounded forward A* (from $\mathit{start}$ using an admissible heuristic) on $cost_1$, use $cost_2$ as a tie-breaker, \; 
      \nonl \qquad \qquad
       and stop before expanding a state with $f_1 > \overline{f_1}$. \label{alg:Bi_search_init:init_2}\;
      
    }
   \DoParallel{
      $h^b_2, ub^b_1 \leftarrow$ Run $f_2$-bounded forward A* (using $h^f_2$ as an admissible heuristic) on $cost_2$, use $cost_1$ as a tie-breaker,\;
      \nonl \qquad \qquad ignore unexplored states in the previous round lines~\ref{alg:Bi_search_init:init_1}-\ref{alg:Bi_search_init:init_2},
      update $\overline{f_1}$ via paths matching if feasible \;
      \nonl \qquad \qquad and stop before expanding a state with $f_2 > \overline{f_2}$. \label{alg:Bi_search_init:init_3} \;
      
      $h^f_1, ub^f_2 \leftarrow$ Run $f_1$-bounded backward A* (using $h^b_1$ an admissible heuristic) on $cost_1$, use $cost_2$ as a tie-breaker, \; 
      \nonl \qquad \qquad ignore unexplored states in the previous round lines~\ref{alg:Bi_search_init:init_1}-\ref{alg:Bi_search_init:init_2},
      update $\overline{f_1}$ via paths matching if feasible \;
      \nonl \qquad \qquad and stop before expanding a state with $f_1 > \overline{f_1}$. \label{alg:Bi_search_init:init_4}\;
      
    }
$S' \leftarrow$ non-dominated states explored in both searches of the last round (lines \ref{alg:Bi_search_init:init_3}-\ref{alg:Bi_search_init:init_4})  \;  
\Return{$({\bf h}^f, {\bf h}^b)$, $({\bf ub}^f, {\bf ub}^b)$, ${\bf \overline{f}}$, $S'$}
\end{algorithm}


\begin{enumerate}[nosep]
  \item \textit{Round one:} The algorithm performs a bounded forward search on $\mathit{cost_1}$ and, in parallel, a bounded backward search on $\mathit{cost_2}$.
As the initial global upper bound $\overline{f_1}$ is not known beforehand, the bounded search on $\mathit{cost_2}$ updates the global upper bound $\overline{f_1}$ with $\mathit{ub}^f_1(\mathit{start})$ as soon as it computes $h^f_2(\mathit{start})$.
Since both searches explore the graph concurrently and have direct access to shared parameters, the forward A* search on $\mathit{cost_1}$ will turn into a bounded search as soon as $\overline{f_1}$ gets updated by the concurrent search in the opposite direction.
When the parallel searches of the first round terminate, the initialisation phase has computed two heuristic functions, namely $h^f_2$ and $h^b_1$.
There are two cases where we can terminate the parallel searches of the first round early without needing to go into the second round of concurrent searches:
(1) the problem has no optimal solution if we have found $h^f_2(\mathit{start}) > W$;
(2) if the shortest path on $\mathit{cost_1}$ has been found feasible and $\mathit{ub}^b_2(\mathit{goal}) \leq W$.
  
  \item \textit{Round two:} 
  The algorithm runs a complementary backward A* search on $\mathit{cost_1}$ and, at the same time, a bounded forward A* search on $\mathit{cost_2}$.
Benefiting from the results of the first round, the searches of the second round use two functions $h^f_2$ and $h^b_1$ as informed heuristics to guide A*.
Hence, we can expect faster searches in round two.
The second round will also perform two tasks during the search.
Firstly, it only explores the expanded states of round one, i.e., none of the searches in the second round will explore states identified as out-of-bounds in round one.
Secondly, it tries to update the global upper bound $\overline{f_1}$ with partial path matching.
Since each state expanded in the searches of the second round has access to at least one complementary shortest path obtained via the first round, we can update the global upper bound $\overline{f_1}$ if joining partial paths with their complementary optimum paths yields a valid $\textit{start}$-$\textit{goal}$ path.
\end{enumerate}

When the initialisation phase is complete, {WC-EBBA*}\textsubscript{par} uses lower bound heuristics $h^f_1$ and $h^b_1$ to determine the search budget factors $\beta^f$ and $\beta^b$ as in {WC-EBBA*} (via Eq.~(\ref{bias_factor}) in \ref{sec:WC_EBBA}).
It then initialises a (shared) node pair $\mathit{Sol}$ with an unknown secondary cost $f^{sol}_2$ to be updated during the main search.
Note that the order of searches in the initialisation phase of {WC-EBBA*}\textsubscript{par} is slightly different from that of {WC-EBBA*}.
The sequential preliminary searches in {WC-EBBA*} allows us to first reduce the graph size via two bounded searches on $\mathit{cost_2}$, followed by two complementary searches on $\mathit{cost_1}$.
However, this ordering might not help reduce the graph size in the parallel scheme, mainly because we would not be able to use heuristics obtained in one search to inform the search in the opposite direction.
For example, if we run concurrent forward and backward searches on $\mathit{cost_2}$ in the first round, there would be no lower bound heuristics for either of the searches, and thus they would turn into less informed bounded A*.
The other possible configuration could be running parallel searches of each round in one direction, e.g., searching backwards on $\mathit{cost_1}$ and $\mathit{cost_2}$ in the first round.
The main problem with this configuration is that heuristics obtained in the direction of the second round would become much more informed than heuristics of the first round in the opposite direction.
Hence, the search in one direction would always be weaker on both $\mathit{cost_1}$ and $\mathit{cost_2}$ heuristics.

In summary, our proposed parallel search ordering ensures that:
(1)~the second round can benefit from the heuristics obtained in the first round to reduce the graph size, delivering $S'$ to the next phase as a subset of valid states explored in both rounds;
(2)~each constrained search in the next phase can benefit from a set of informed heuristics for either pruning or guiding the search in the bidirectional setting;
(3)~the initialisation time is improved as there are only two rounds of bounded A*.
\subsection{Constrained Search}
In this phase, {WC-EBBA*}\textsubscript{par} executes two constrained bidirectional searches in parallel.
The parallel framework allows both constrained searches of {WC-EBBA*} to explore the graph concurrently, so the search is no longer led by one direction at a time.
Algorithm~\ref{alg:rc_ebba_par} shows the main procedures of the constrained search in the generic direction~$d$.

\textbf{Algorithm Description:}
Given ${\bf \overline{f}}$ and $\mathit{Sol}$ as the shared data structures of the algorithm, each (constrained) search starts with initialising the search in direction~$d$ via the Setup($d$) procedure and also with identifying $d'$ as the opposite direction.
At this point, the search has one node in $\mathit{Open}^d$ associated with the initial state in direction~$d$.
In contrast to {WC-EBBA*}, each search in {WC-EBBA*}\textsubscript{par} only works with one priority queue.
As can be seen in line~\ref{alg:rc_ebba_par:exract} of Algorithm~\ref{alg:rc_ebba_par}, the constrained search in direction~$d$ is continued by exploring nodes in $\mathit{Open}^d$ in the $(f_1,f_2)$ order.
Given $x$ as the least cost node in the priority queue, the algorithm removes $x$ from $\mathit{Open}^d$ and checks it against the termination criterion via line~\ref{alg:rc_ebba_par:termination}.
If the algorithm finds $x$ (as the least cost node) out-of-bounds, the search in direction~$d$ can safely terminate (see Lemma~\ref{lemma:astar_terminate}).
Otherwise, if the node's estimated cost is within the bounds, the search will follow the standard procedures of {WC-EBBA*} to explore $x$ as a valid node.
The procedure includes matching $x$ with candidate nodes of the opposite direction stored in $\chi^{d'}(s(x))$ and then storing $x$ in $\chi^{d}(s(x))$ for future expansions with $s(x)$ in direction $d'$, only if $x$ is in the coupling area (see \ref{sec:WC_EBBA} for more details).
For the sake of clarity, we present the main steps involved in node exploration via Explore($x,d,d'$) in Procedure~\ref{alg:explore}.
%
\begin{figure}[t]
\noindent
\begin{minipage}[t]{.49\textwidth}
\null 
\input{Algorithms/RC_EBBA_Main_par}
\end{minipage}%
\hspace{2mm}
\begin{minipage}[t]{.49\textwidth}
\null
\input{Algorithms/Explore}
\end{minipage}
\end{figure}
%

Unlike {WC-EBBA*} where the search is led by the least-cost node of one direction, {WC-EBBA*}\textsubscript{par} runs two parallel searches with independent priority queues.
This means that there are no dependencies between the $f_1$-value of nodes in the forward and backward priority queues, and it is very likely for one of the directions to show faster progress on $f_1$-values.
In other words, the faster search is no longer halted by the slower search, and {WC-EBBA*}\textsubscript{par} can potentially find the optimal solution faster than {WC-EBBA*}.
Note that when {WC-EBBA*} extracts a node $x$ from $\mathit{Open}^{d}$, it ensures that all nodes in $\mathit{Open}^{d'}$ have a primary cost no smaller than $f_1(x)$.
Given this important feature in {WC-EBBA*}\textsubscript{par}, we now prove the correctness of the algorithm.
\begin{theorem}
\label{theorem:wc_ebba_par}
{WC-EBBA*}\textsubscript{par} returns a node pair corresponding with a {\bf cost}-optimal solution path for the WCSPP.
\end{theorem}
\begin{proof}
As {WC-EBBA*}\textsubscript{par} employs {WC-EBBA*}'s procedures for its node expansion, the correctness of the involved strategies are directly derived from {WC-EBBA*}'s correctness (provided in \ref{sec:WC_EBBA}).
Thus, we just discuss the correctness of the stopping criterion.
We already know that when the search in direction~$d$ terminates, there is no promising node in the corresponding priority queue with a cost estimate smaller than the best-known solution cost $\overline{f_1}$.
However, since both searches use the same stopping criterion, {WC-EBBA*}\textsubscript{par} will not terminate until both (constrained) searches terminate.
Generally, if one direction has already completed its search by surpassing the global upper bound $\overline{f_1}$, the algorithm waits for the other (active) direction to confirm the optimality of the solution stored in $\mathit{Sol}$.
In this situation, if the active direction finds a better solution, it can still update $\overline{f_1}$ and $\mathit{Sol}$ accordingly, otherwise, $\mathit{Sol}$ remains optimal.
In either case, the active search will eventually confirm the optimality of the existing solution by surpassing the established upper bound.
Furthermore, reducing the global upper bound $\overline{f_1}$ in the active search does not affect the correctness of the other (terminated) search, essentially because the algorithm never builds a solution path with invalid nodes.
Therefore, {WC-EBBA*}\textsubscript{par} guarantees the optimality of $\mathit{Sol}$ when it terminates.
\end{proof}
\textbf{Memory:}
Both {WC-EBBA*} and {WC-EBBA*}\textsubscript{par} use the same data structures to initialise their bidirectional searches, some of them shared between the searches in {WC-EBBA*}\textsubscript{par}.
Therefore, there is no difference in the minimum space requirement of the algorithms.
However, there is a minor search overhead associated with parallel search in {WC-EBBA*}\textsubscript{par}.
Recall the queuing strategy in {WC-EBBA*}. The sequential search guarantees that the entire algorithm never explores nodes with an $f_1$-value larger than the optimal cost $\overline{f_1}$.
However, this is not always the case in {WC-EBBA*}\textsubscript{par}.
As an example, the algorithm may terminate its constrained search in direction~$d$ with a tentative solution of cost $\overline{f_1}$, but later discovers an optimal solution pair $(x,y)$ with the primary cost $f'_1 < \overline{f_1}$ in the opposite direction~$d'$.
In this case, it is not difficult to see that the expansion of nodes with $f_1$-values larger than the optimal cost $f'_1$ in direction~$d$ was unnecessary.
Therefore, we expect {WC-EBBA*}\textsubscript{par} to use more space than {WC-EBBA*} due to this resulting search overhead.
%
\section{Practical Considerations}
\label{sec:practicall}
As A*-based algorithms enumerate all valid paths, the size of the $\mathit{Open}$ lists can grow exponentially during the constrained search.
This case is more serious in A* with lazy dominance tests, essentially because priority queues also contain dominated nodes.
Furthermore, we can similarly see that the number of generated nodes will show exponential growth and the search may need significant memory to store all nodes for solution path construction.
For instance, A*-algorithms can easily generate and expand billions of search nodes in hard problems.
Following the practical considerations presented in \citet{AhmadiTHK21_esa} for the bi-objective search, we now describe two techniques to handle search nodes generated in the constrained search more efficiently.
\subsection{More efficient priority queues}
\label{priority_queus}
The performance of constrained A* search with lazy dominance test can suffer when the queue size grows to very large numbers of nodes.
Contrary to eager dominance or other conservative approaches where the search rigorously removes dominated nodes in the queue, as in \citet{PulidoMP15}, none of our A* searches tries to remove the dominated nodes from the queue unless they are extracted.
There are some Dijkstra-like approaches \cite{dijkstra1959note} in the literature that only keep one best candidate node of each state in the priority queue, as in \citet{Sedeno-NodaC19}.
However, our A*-based algorithms do not work with such substitution of nodes during the search.
Therefore, we need to invest in designing efficient priority queues to effectively order significant numbers of nodes in constrained search with A*.
We describe our node queuing strategies as follows.

In all of our A*-based algorithms, the lower and upper bounds on the $\mathit{f}_p$-values of the search nodes are known prior to the constrained searches, namely via the heuristic function $h^d_p$ and the global upper bound $\overline{f_p}$.
Let $[f_{\mathit{min}},f_{\mathit{max}}]$ be the range of all possible $f_p$-values generated by A* in the ($f_p,f_s$) order.
To achieve faster operations in our $\mathit{Open}$ lists, we use fixed-size bucket-based queues.
Although there may be cases where the number of nodes in the priority queue is bounded and the bucket list is sparsely populated, for the majority of cases where the number of nodes grows exponentially in our A* searches, we expect to see almost all the buckets filled.
To this end, we investigate two types of bucket-based priority queues based on the multi-level bucket data structures in the literature \cite{Dial69,DenardoF79,CherkasskyGR96}.
Consider a bucket list with $\Delta f \in \mathbb{N}^+$ (as a fixed parameter) identifying the bucket width.
We limit the size of the bucket list such that a node with $f_{\mathit{min}}$ (resp. $f_{\mathit{max}}$) is always placed in the first (resp. last) bucket, i.e., we have
\begin{equation}
    \text{Bucket Size (\textit{BS})}=\lfloor \dfrac{f_{\mathit{max}}-f_{\mathit{min}}}{\Delta f} \rfloor +1
\end{equation}
Given \textit{BS} and $\Delta f$ denoting the bucket size and width, respectively, we now explain each queue type as follows.
\begin{enumerate}[nosep]
    \item Bucket queue: 
    a two-level queue with buckets in both levels \cite{DenardoF79,CherkasskyGR96}. 
    A high-level bucket $i \in \mathbb{N}^0_{<BS} = \{i \in \mathbb{N}^0 | i < BS \}$ contains all nodes whose $f_p$-value falls in the $[f_\mathit{min}+i\times\Delta f, f_\mathit{min}+(i+1)\times\Delta f-1]$ range except for the nonempty high-level bucket with the smallest index~$k$.
    A node $x$ with $f_p(x)$ in the $[f_\mathit{min}+k\times\Delta f, f_\mathit{min}+(k+1)\times\Delta f-1]$ range is maintained in the low-level bucket.
    The size of the low-level bucket is $\Delta f$, so we have one bucket for every distinct $f_p$-value in the range.
    In other words, a low-level bucket $j  \in \mathbb{N}^0_{<\Delta f} = \{j \in \mathbb{N}^0 | j < \Delta f\}$ contains all nodes whose $f_p$-value is equal to $f_\mathit{min}+k\times\Delta f + j$.
    To handle nodes in each bucket, we can use a linked list structure with two node extraction strategies: First-In, First-Out (FIFO) or Last-In, First-Out (LIFO).
    This queue is only able to handle integer costs.
    \item Hybrid queue: a two-level priority queue with buckets in the higher level and a binary heap in the lower level \cite{DenardoF79}. 
    Similar to the bucket queue, a high-level bucket $i \in \mathbb{N}^0_{<BS} = \{i \in \mathbb{N}^0 | i < BS \}$ contains all nodes whose $f_p$-value falls in the $[f_\mathit{min}+i\times\Delta f, f_\mathit{min}+(i+1)\times\Delta f-1]$ range except for the nonempty high-level bucket with the smallest index~$k$.
    A node $x$ with $f_p(x)$ in the range $[f_\mathit{min}+k\times\Delta f, f_\mathit{min}+(k+1)\times\Delta f-1]$ is maintained in the low-level binary heap structure.
    This queue type can handle both integer and non-integer costs.
\end{enumerate}
\begin{figure}[t]
\begin{tikzpicture} [scale=0.9, every node/.style={scale=1}]
\draw (1.5,0) -- (3.5,0);
\draw [dashed] (3.5,0) -- (5,0);
\draw (1.5,1) -- (3.5,1);
\draw [dashed] (3.5,1) -- (5,1);
\draw (1.5,0) -- (1.5,1);
\draw [dashed] (5,0) -- (5,1);

\draw [<->] (1.6,1.25) to node[above]{$\Delta f= \infty$} (4.9,1.25);
\node at (3.25,0.5) {Binary-Heap (BH)};

\draw (7.5,-0.25) rectangle (8.75,0.25);
\draw (7.5,0.5) rectangle (8.75,1);
\draw (9.75,0.5) rectangle (11,1);
\draw [dashed] (8.75,0.5) to (9.75,0.5); 
\draw [dashed] (8.75,1) to (9.75,1); 
\draw [<->] (7.5,1.25) to node[above]{$\Delta f$} (8.75,1.25);
\draw [<->] (9.75,1.25) to node[above]{$\Delta f$} (11,1.25);
\node at (8.15,0.75) {\small LL};
\node at (9.25,0.75) {$\dots$};
\node at (10.35,0.75) {\small LL};
\node at (8.15, 0) {\small BH};
\node at (6.75,0.75) {\small High-level};
\node at (6.75,0) {\small Low-level};


\draw (13.25,-0.25) rectangle (13.75,0.25);
\draw [dashed] (13.75,-0.25) to (14.25,-0.25);
\draw [dashed] (13.75,0.25) to (14.25,0.25);
\draw (14.25,-0.25) rectangle (14.75,0.25);
\node at (14,0.0) {$\dots$};
\node at (13.5,0.0) {\small LL};
\node at (14.5,0.0) {\small LL};
\draw (13.25,0.5) rectangle (14.75,1);
\draw (15.75,0.5) rectangle (17.25,1);
\draw [dashed] (14.75,0.5) to (15.75,0.5); 
\draw [dashed] (14.75,1) to (15.75,1); 
\draw [<->] (13.25,1.25) to node[above]{$\Delta f$} (14.75,1.25);
\draw [<->] (15.75,1.25) to node[above]{$\Delta f$} (17.25,1.25);
\node at (14.0,0.75) {\small LL};
\node at (15.25,0.75) {$\dots$};
\node at (16.5,0.75) {\small LL};
\node at (12.5,0.75) {\small High-level};
\node at (12.5,0) {\small Low-level};

\node at (3.25,-0.75) {\small (a)};
\node at (9.25,-0.75) {\small (b)};
\node at (15.25,-0.75) {\small (c)};
\end{tikzpicture}
\caption{\small Schematic of priority queues studied: (a) Conventional Binary Heap (BH) queue with no limit on $f_p$-values; (b) hybrid queue with buckets in the higher level and binary heap in the lower level. $\Delta f$ denotes bucket width; (c) two-level bucket queue with Linked List (LL) in buckets of both levels. The width of the low-level buckets is one.}
\label{fig:queues}
\end{figure}
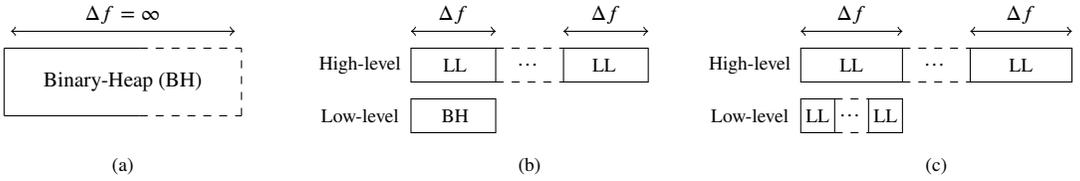
Figure~\ref{fig:queues} illustrates our priority queues and compares them against the traditional binary heap queue.
In this figure, $\Delta f$ denotes the bucket width.
We can see that the hybrid queue can be converted into a form of binary heap if we pick a large enough bucket width.
We can also convert our two-level bucket queue into a one-level bucket queue \cite{Dial69} by simply setting $\Delta f=1$.
In this case, nodes can be directly added (removed) to (from) the high-level buckets and the use of the low-level bucket is no longer necessary.

\textbf{Queue operations:}
To add new nodes to the queue, we simply insert them into high-level buckets corresponding with their $f_p$-value.
To find and extract the least-cost node, since $f_p$-values in A* are monotonically non-decreasing, we can simply scan the high-level buckets from left ($f_{\mathit{min}}$) to right ($f_{\mathit{max}}$).
The first bucket is always a non-empty bucket when our constrained searches are initialised.
Let $k$ be the smallest index of a non-empty high-level bucket.
We transfer all nodes in the nonempty high-level bucket $k$ to the low-level structure for upcoming extractions.
In the meantime, if the queue receives a new node with an $f_p$-value corresponding with index $k$, we directly add the node to the low-level data structure.
Nodes in the low-level structure are explored during the search in the order of their $f_p$-values.
Before each node extraction, if all nodes in the low-level data structure are already extracted, we increase $k$ to the index of the next non-empty (high-level) bucket and then transfer all nodes of the non-empty bucket to the low-level data structure. Nodes are extracted from the low-level structure only.

\textbf{Tie-breaking:}
The bucket queue with linked lists is obviously unable to handle tie-breaking, but it instead offers fast queue operations.
In the hybrid queue with binary heaps, however, we can break ties between $f_p$-values by simply comparing nodes on their $f_s$-values in the low-level binary heap.
\subsection{Memory efficient backtracking}
Node generation is a necessary part of all of our WCSPP algorithms with A*.
Each node occupies a constant amount of memory and represents a partial path. A typical node representation would contain information such as {\bf f}-values, the node's corresponding state, its position in the priority queue, and its parent, required for solution path construction.
Considering the huge number of generated nodes in difficult problems, we use our memory-efficient approach in \citet{AhmadiTHK21_esa} as part of solution path construction for all of our studied A* algorithms.
Since constrained search in A* only expands each node at most once (the search prunes cycles), every time a node is expanded, we store its backtracking information in compact data structures outside the node and then recycle the memory of the processed node for future node expansions.
In other words, since the majority of a node's information will no longer be required for solution path construction (parameters such as position of the node in the queue and {\bf f}-values), we only extract the minimal backtracking information and return the node to the memory manager.
Clearly, the memory occupied by nodes pruned during the search can similarly be recycled (see the example given in \ref{sec:compact_backtracing} for further illustration).
The final size of the compact structure varies from instance to instance, but it is shown that, compared to the conventional back-pointer approach, the compact approach is on average five times more efficient in terms of memory \cite{AhmadiTHK21_esa}.
%
\section{Experimental Analysis}
\label{sec:experiment}
We compare our new algorithms with the state-of-the-art solution approaches available for the WCSPP and RCSPP and evaluate them on a set of 2000 WCSPP instances \cite{AhmadiTHK22_socs} for 12 maps in the 9th DIMACS Implementation Challenge \footnote{DIMACS - Shortest Paths; 2005. http://www.diag.uniroma1.it/~challenge9/.}, with the largest map containing around 24~M nodes and 57~M edges (see \ref{sec:benchmark_SM} for the benchmark details).
Following the literature, we define the weight limit $\mathit{W}$ based on the tightness of the constraint $\delta$ as:
\begin{equation}
\delta=\dfrac{W-h_2}{ub_2-h_2} \quad \text{for} \quad \delta \in\{10\%,20\%,\dots,70\%,80\%\}  
\end{equation}
where $h_2$ and $ub_2$ are, respectively, lower and upper bounds on $\mathit{cost_2}$ of $\mathit{start}$-$\mathit{goal}$ paths.
In this setup, high (resp. low) values of $\delta$ mean that the weight limit $W$ is loose (resp. tight).

\textbf{Studied algorithms:} we consider the award-winning BiPulse algorithm of \citet{cabrera2020exact}, {WC-EBBA*} \cite{AhmadiTHK21}, and {WC-BA*} \cite{AhmadiTHK22_socs} algorithms .
For the other constrained search methods {RC-BDA*} \cite{thomas2019exact}, Pulse \cite{lozano2013exact} and CSP \cite{sedeno2015enhanced}, although shown to be slower than WC-EBBA* in \citet{AhmadiTHK21}, we evaluate their performance against the benchmark instances in \ref{sec:experment_result_SM}.

\textbf{Implementation:}
We implemented all the A*-based algorithms ({WC-A*}, {WC-BA*}, {WC-EBBA*} and {WC-EBBA*}\textsubscript{par}) in C++ and used the Java implementation of the BiPulse algorithm kindly provided to us by its authors.
For {WC-BA*}, we implement its standard variant (with the HTF method).
Our graph implementation removes duplicate edges from the DIMCAS graphs, i.e., if there are two (or more) edges between a pair of states in the graph, we only keep the lightest edge.
In addition, for the initialisation phase of the A*-based algorithms, we use spherical distance as an admissible heuristic function for both (distance and time) objectives.
All the A*-based algorithms use the same type of priority queue and solution construction approach.
In particular, since all costs in the benchmark instances are integer, we use bucket queues with linked lists (using the LIFO strategy with $\Delta f = 1$) for the priority queue of constrained A* searches, along with the compact approach discussed in Section~\ref{sec:practicall} for solution path construction.
There is no procedure for the solution path construction in the Pulse and BiPulse implementations, so the code only returns optimal costs.
All C++ code was compiled with O3 optimisation settings using the GCC7.5 compiler.
The Java code was compiled with OpenJDK version 1.8.0\_292.
We ran all experiments on an AMD EPYC 7543 processor running at 2.8~GHz and with 128~GB of RAM, under the SUSE Linux Server 15.2 environment and with a one-hour timeout.
In addition, we allocated two CPU cores to all parallel algorithms.
There are five algorithms and 2000 instances, resulting in 10000 runs.
However, we realised that BiPulse is unable to handle our two largest maps CTR and USA (400 instances altogether) due to some inefficiencies in its graph implementation.
We can similarly see that BiPulse has not been tested on these two graphs in its original paper \cite{cabrera2020exact} either.
In order to achieve more consistent computation time in the parallel setting, especially in easy instances with small runtime, we perform five consecutive runs of each algorithm and store the results of the run showing the median runtime (algorithms attempt each instance five times).
Thus, we perform $5 \times 9600 =48000$ runs.
All runtimes we report in this paper include initialisation time.
Our codes, benchmark instances and detailed results are publicly available\footnote{\url{https://bitbucket.org/s-ahmadi/biobj}}.
\subsection{Algorithmic Performance}
We now analyse the performance of the selected algorithms over the benchmark instances as provided in Table~\ref{table:Results}.
We have 160 point-to-point WCSPP instances in each map, except in the USA map, for which we have 240 instances.
As stated before, BiPulse was unable to handle instances of two maps. 
We report for each algorithm the number of solved cases $|S|$, the runtime in seconds and memory usage in MB.
Note that because of the difficulties in reporting the memory usage, we allow 1~MB tolerance in our experiments.
For the runtime of unsolved cases, we generously report a runtime of one hour (the timeout) by assuming that the algorithm would have found the optimal solution right after the timeout.
We discuss the results in three aspects as follows.
\begin{table}[ht]
\caption{\small Number of solved cases $|S|$ (out of 240 for USA and 160 for the other maps), runtime and memory use of the algorithms. Runtime of unsolved instances is assumed to be 3600~seconds. Memory is reported only for the solved cases and 1MB=10\textsuperscript{3}KB.}
\label{table:Results}
\begin{multicols}{2}
\centering
\small
\begin{adjustbox}{width=1\textwidth}
\renewcommand{\arraystretch}{1}
\begin{tabular}{| l | l | r | *{3}{r} | *{2}{r}|}
\hline
 & & & \multicolumn{3}{c|}{Runtime(s)} & \multicolumn{2}{c|}{Memory(MB)} \\ \cline{4-8}
\headrow
\textbf{Map} & \textbf{Algorithm} & \textbf{$|S|$} & \textbf{Min} & \textbf{Avg.} & \textbf{Max} & \textbf{Avg.} & \textbf{Max} \\
\hline
NY  & WC-A*                                        & 160 & \textbf{0.01} & \textbf{0.06}   & \textbf{0.15}    & \textbf{2}   & \textbf{7}    \\
    & WC-BA*                                       & 160 & \textbf{0.01} & \textbf{0.06}   & 0.19             & 3            & 8             \\
    & WC-EBBA*                                     & 160 & \textbf{0.01} & 0.08            & 0.16             & \textbf{2}   & 13            \\
    & WC-EBBA*\textsubscript{par} & 160 & \textbf{0.01} & \textbf{0.06}   & \textbf{0.15}    & 3            & 14            \\
    & BiPulse                                      & 160 & 0.46          & 0.85            & 2.35             & 13           & 126           \\
\hline
BAY & WC-A*                                        & 160 & \textbf{0.01} & 0.10            & 0.40             & \textbf{3}   & 20            \\
    & WC-BA*                                       & 160 & \textbf{0.01} & \textbf{0.09}   & \textbf{0.39}    & \textbf{3}   & \textbf{16}   \\
    & WC-EBBA*                                     & 160 & \textbf{0.01} & 0.13            & 0.46             & 4            & 35            \\
    & WC-EBBA*\textsubscript{par} & 160 & \textbf{0.01} & 0.11            & 0.49             & 5            & 20            \\
    & BiPulse                                      & 160 & 0.54          & 1.08            & 4.05             & 41           & 596           \\
\hline
COL & WC-A*                                        & 160 & \textbf{0.02} & 0.22            & 1.77             & \textbf{7}   & 67            \\
    & WC-BA*                                       & 160 & 0.03          & \textbf{0.20}   & 1.30             & 8            & \textbf{60}   \\
    & WC-EBBA*                                     & 160 & 0.04          & 0.25            & \textbf{1.01}    & 10           & 138           \\
    & WC-EBBA*\textsubscript{par} & 160 & 0.04          & 0.22            & 1.25             & 11           & 84            \\
    & BiPulse                                      & 160 & 0.73          & 2.85            & 27.46            & 160          & 1035          \\
\hline
FLA & WC-A*                                        & 160 & 0.19          & 1.45            & 15.33            & \textbf{38}  & 337           \\
    & WC-BA*                                       & 160 & \textbf{0.16} & \textbf{1.11}   & 4.92             & 40           & \textbf{183}  \\
    & WC-EBBA*                                     & 160 & 0.22          & 1.29            & 4.99             & \textbf{38}  & 199           \\
    & WC-EBBA*\textsubscript{par} & 160 & 0.17          & 1.15            & \textbf{4.57}    & 50           & 212           \\
    & BiPulse                                      & 160 & 1.47          & 35.29           & 396.44           & 287          & 1271          \\
\hline
NW  & WC-A*                                        & 160 & 0.12          & 1.83            & 12.90            & \textbf{57}  & \textbf{314}  \\
    & WC-BA*                                       & 160 & \textbf{0.10} & 2.20            & 18.50            & 86           & 585           \\
    & WC-EBBA*                                     & 160 & 0.13          & 2.11            & 16.20            & 113          & 1039          \\
    & WC-EBBA*\textsubscript{par} & 160 & 0.11          & \textbf{1.81}   & \textbf{12.54}   & 138          & 1084          \\
    & BiPulse                                      & 160 & 1.58          & 163.50          & 1440.52          & 849          & 6586          \\
\hline
NE  & WC-A*                                        & 160 & 0.17          & 2.07            & 31.51            & \textbf{53}  & \textbf{557}  \\
    & WC-BA*                                       & 160 & \textbf{0.15} & 2.19            & 31.50            & 80           & 1003          \\
    & WC-EBBA*                                     & 160 & 0.21          & 2.67            & 32.00            & 82           & 699           \\
    & WC-EBBA*\textsubscript{par} & 160 & 0.20          & \textbf{1.87}   & \textbf{27.11}   & 86           & 735           \\
    & BiPulse                                      & 156 & 2.11          & 237.58          & 3600.00          & 496          & 6350          \\
\hline
\end{tabular}
\begin{tabular}{| l | l | r | *{3}{r} | *{2}{r} |}
\hline
\hiderowcolors
 & & & \multicolumn{3}{c|}{Runtime(s)} & \multicolumn{2}{c|}{Memory(MB)} \\ \cline{4-8}
 \headrow
\textbf{Map} & \textbf{Algorithm} & \textbf{$|S|$} & \textbf{Min} & \textbf{Avg.} & \textbf{Max} & \textbf{Avg.} & \textbf{Max} \\
\showrowcolors
CAL & WC-A*                                        & 160 & 0.12          & 4.10            & 57.78            & 96           & 1295          \\
    & WC-BA*                                       & 160 & 0.10          & 1.87            & 21.08            & \textbf{57}  & \textbf{711}  \\
    & WC-EBBA*                                     & 160 & 0.12          & 1.93            & 11.71            & 60           & 460           \\
    & WC-EBBA*\textsubscript{par} & 160 & \textbf{0.09} & \textbf{1.57}   & \textbf{10.31}   & 78           & 834           \\
    & BiPulse                                      & 160 & 2.26          & 154.10          & 2640.61          & 518          & 5068          \\
\hline
LKS & WC-A*                                        & 160 & 0.09          & 25.45           & 298.96           & \textbf{507} & \textbf{4246} \\
    & WC-BA*                                       & 160 & 0.08          & 28.18           & 334.46           & 830          & 8122          \\
    & WC-EBBA*                                     & 160 & \textbf{0.07} & 29.14           & 349.12           & 872          & 5839          \\
    & WC-EBBA*\textsubscript{par} & 160 & \textbf{0.07} & \textbf{20.13}  & \textbf{244.63}  & 1022         & 8294          \\
    & BiPulse                                      & 109 & 2.76          & 1439.84         & 3600.00          & 980          & 5134          \\
\hline
E   & WC-A*                                        & 160 & \textbf{0.07} & 36.16           & 349.42           & \textbf{673} & \textbf{5180} \\
    & WC-BA*                                       & 160 & 0.09          & 35.72           & 366.17           & 984          & 7496          \\
    & WC-EBBA*                                     & 160 & 0.10          & 43.15           & 418.95           & 1388         & 10581         \\
    & WC-EBBA*\textsubscript{par} & 160 & 0.10          & \textbf{33.45}  & \textbf{310.04}  & 1522         & 13373         \\
    & BiPulse                                      & 99  & 3.66          & 1617.31         & 3600.00          & 573          & 4000          \\
\hline
W   & WC-A*                                        & 160 & \textbf{0.32} & \textbf{31.25}  & \textbf{349.29}  & \textbf{560} & \textbf{4990} \\
    & WC-BA*                                       & 160 & 0.43          & 38.57           & 412.05           & 920          & 6794          \\
    & WC-EBBA*                                     & 160 & 0.36          & 40.62           & 367.12           & 985          & 6796          \\
    & WC-EBBA*\textsubscript{par} & 160 & 0.39          & 35.66           & 390.92           & 1294         & 13354         \\
    & BiPulse                                      & 101 & 7.76          & 1620.46         & 3600.00          & 547          & 5421          \\
\hline
CTR & WC-A*                                        & 160 & 0.27          & 74.62           & 663.09           & \textbf{1063}         & \textbf{7548}          \\
    & WC-BA*                                       & 160 & \textbf{0.26} & 80.27           & 631.31           & 1738         & 11073         \\
    & WC-EBBA*                                     & 160 & 0.29          & 99.39           & 758.78           & 2223         & 17463         \\
    & WC-EBBA*\textsubscript{par} & 160 & \textbf{0.26} & \textbf{67.13}           & \textbf{536.69}           & 2257         & 17508         \\
    & BiPulse                                      & -   & -             & -               & -                & -            & -             \\
\hline
USA & WC-A*                                        & 236 & 0.20          & 471.61          & 3600.00          & 5272         & 35783         \\
    & WC-BA*                                       & 240 & \textbf{0.17} & 394.11          & 3444.40          & 8577         & 68279         \\
    & WC-EBBA*                                     & 240 & 0.26          & 379.19          & 2120.83          & 7709         & 63995         \\
    & WC-EBBA*\textsubscript{par} & 240 & 0.21          & \textbf{298.72} & \textbf{1651.07} & 8791         & 66932         \\
    & BiPulse                                      & -   & -             & -               & -                & -            & -            \\ 
\hline
\end{tabular}
\end{adjustbox}
\end{multicols}
\end{table}

\textbf{Solved cases:}
All A*-based algorithms have been able to fully solve all instances of 11 maps.
For the BiPulse algorithm, however, we can see it has been struggling with some instances from the NE, LKS, E and W maps.
Hence, we could expect even more unsolved cases in the larger maps CTR and USA if we were able to evaluate BiPulse on that set of instances.
Comparing the number of solved instances of A*-based algorithms in the USA map, we can see that {WC-EBBA*} (both sequential and parallel versions) and {WC-BA*} are the best-performing algorithms with all instances solved.
For {WC-BA*}, we can see that it solves its most difficult instance just 2.5 minutes before the timeout.
For {WC-A*}, however, we see it is the only algorithm that shows unsolved cases within the one-hour timeout when compared with its A*-based competitors.

\textbf{Runtime:}
We report the minimum, average, and maximum runtime of the algorithms in each map.
Boldface values denote the smallest runtime among all algorithms.
We can see that the runtime of all algorithms increases with the graph size, and we have obtained larger values in larger graphs.
From the results, it becomes clear that the improved bidirectional search scheme in BiPulse does not contribute to runtimes competitive with A* and the depth-first search nature of BiPulse is still a potential reason for its poor performance on large graphs.
Comparing the average runtime of BiPulse against the A*-based algorithms, we see it can be up to two orders of magnitude slower than its competitors.
Among the A*-based methods, we can nominate our {WC-EBBA*}\textsubscript{par} as the fastest algorithm due to showing smaller runtimes overall, specifically in larger maps CTR and USA by showing the maximum runtime of 28~minutes.
For more detailed runtime analyses, please see~\ref{sec:time_SM}.

\textbf{Memory:}
Table~\ref{table:Results} also presents the average and maximum memory usage of the algorithms over the solved instances.
The results show that, although BiPulse's implementation does not handle solution path construction, its memory requirement is considerably higher than its competitors.
Among the A*-based algorithms, we can see that {WC-A*} and {WC-BA*} show smaller values than {WC-EBBA*} and its parallel variant {WC-EBBA*}\textsubscript{par}.
In the E map, for example, the results show that the maximum memory requirement of {WC-EBBA*} and {WC-EBBA*}\textsubscript{par} in difficult instances can roughly be as big as 10.6~GB and 13.4~GB, respectively, but both {WC-A*} and {WC-BA*} manage such instances with about 5.2-7.5~GB of memory in nearly the same amount of time.
Although time and space are highly correlated in search strategies and slower algorithms are likely to generate more nodes, we can see how frontier collision in {WC-EBBA*} contributes to faster runtime but in turn higher memory usage due to the partial path matching procedure.
In the NW, E and LKS maps for example, the results show that {WC-A*} outperforms {WC-EBBA*} in terms of runtime while using two times less memory on average.
{WC-A*} also shows comparable performance to {WC-EBBA*}\textsubscript{par} in the NW map (only 20 milliseconds slower on average) but consumes about 2.5 times less memory on average.
Comparing the average memory usage of {WC-A*} and {WC-BA*}, we realise that
{WC-BA*} might consume up to 65\% more space than {WC-A*} on average, and in almost all of our large graphs (except the USA map), they both perform better than {WC-EBBA*} and {WC-EBBA*}\textsubscript{par} in memory use.
We have provided a more detailed memory analysis of the algorithms in~\ref{sec:time_SM}.
\subsection{Performance Impact of Constraint Tightness}
\begin{figure}[!t]
\centering
\includegraphics[width=0.8\textwidth]{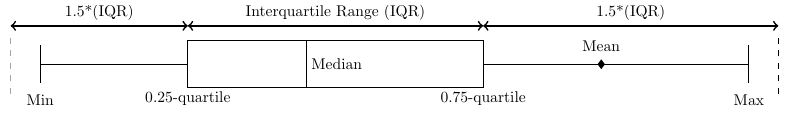}
\caption{\small A schematic of box-plot visualising the distribution of data: the plot shows the minimum, the first quartile (25\% of data), the median, the third quartile (75\% of data), the mean and the maximum (minimum and maximum within the 1.5$\times$ IQR).}
\label{fig:boxplot_def}
\end{figure}

We now analyse the performance of each algorithm with varying levels of tightness.
In the experiment setup, we used eight values for the tightness of the WCSPP constraint, namely values ranging from 10\% to 80\%.
We use box plots for all of our constraint-based analyses to show the distribution of values over the instances.
For the sake of clarity, we show in Figure~\ref{fig:boxplot_def} what type of statistical information each box plot presents.

\begin{figure}[t]
\centering
\includegraphics[width=1\textwidth]{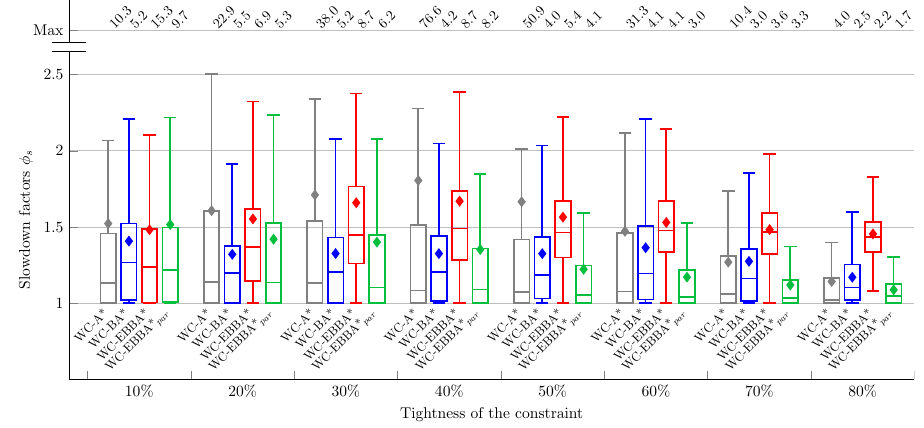}
\caption{\small The distribution of slowdown factors for all the A*-based algorithms against the virtual best oracle in each level of tightness. Values above the plots show the maximum slowdown factor observed in the experiments (outliers are not shown).}
\label{fig:boxplot_slowdown}
\end{figure}

To better study the strengths and weaknesses of the algorithms across various levels of tightness, we need to undertake a form of one-to-one comparison.
To this end, we define our baseline to be a virtual oracle that cannot be beaten by any of the algorithms in terms of runtime.
In other words, for every instance, the virtual oracle is given the best (i.e., smallest) runtime of all algorithms.
Given the virtual best oracle as the baseline, we then calculate for every runtime (across all algorithms) a slowdown factor $\phi_{s}$ with the virtual oracle's runtime as the baseline, i.e, we have $\phi_s \geq 1$ and at least one algorithm with $\phi_s =1$ for every instance.
This means that algorithms with $\phi_s$-values close to one are as good as the virtual best oracle.
For the sake of better readability, this analysis does not include BiPulse as it shows significantly larger slowdown factors compared to our A*-based algorithms, and there was no instance for which BiPulse achieves $\phi_s =1$ (it never beats the virtual oracle).
Figure~\ref{fig:boxplot_slowdown} shows the range of slowdown factors obtained for each algorithm across all levels of tightness.
We describe patterns for each algorithm.
\begin{itemize}[nosep]
    \item {WC-A*}: This algorithm works consistently well across all levels of tightness in terms of the median slowdown factor.
    However, {WC-A*}'s performance gets closer to that of the virtual best oracle when the weight constraint becomes loose.
    The main reason for this fast response in loose constraints is in {WC-A*}'s simple (unidirectional) initialisation phase. Problems with loose constraints normally appear to be easy, mainly because the optimal solution is not far from the initial solution. Therefore, there might be easy cases where the initialisation time is the dominant term in the total execution time, making algorithms with a faster initialisation phase the best performer.
    Having compared the maximum slowdown factors of {WC-A*} with other algorithms, we can observe that {WC-A*} is the only algorithm that shows the largest maximum slowdown factors in almost all levels of tightness (except in 10\%).
    This means that it will be more likely to get the worst performance with {WC-A*} than with other A*-based algorithms.
    \item {WC-BA*}: 
    Similar to {WC-A*}, {WC-BA*} performs well in loose constraints, but is relatively slow in tightly constrained instances.
    However, {WC-BA*} shows the best average factor in the 10-40\% tightness range.
    More importantly, the maximum slowdown factors of {WC-BA*} are considerably better than that of {WC-A*}, and even smaller than {WC-EBBA*} in almost all levels of tightness.
    We can nominate one potential reason for this behaviour.
    {WC-BA*} is a bidirectional algorithm, so its worst-case performance is far less severe than {WC-A*}.
    More accurately, if the search in one direction is slow in some difficult instances, the concurrent search can help {WC-BA*} in reducing the search space.
    In addition, {WC-BA*} uses bidirectional lower bounds and is also able to tune its heuristics during the search, so it effectively does more pruning in tight constraints and will consequently perform better in the runtime comparison. 
    In loose constraints, {WC-A*} and {WC-BA*} perform quite similarly as {WC-BA*}'s initialisation phase is done in parallel, but there may be cases (in the 80\% tightness) where {WC-A*} performs better mainly because the search space is very small and heuristic tuning and parallel search in {WC-BA*} only adds unnecessary overhead.
    \item
    {WC-EBBA*}: This algorithm presents comparable runtimes only in very tightly constrained instances (10\%) and is the weakest solution method in mid-range and loose constraints if we compare algorithms based on their median slowdown factors.
    However, the average performance of {WC-EBBA*} is still better than {WC-A*} in the 10-60\% range.
    In terms of the worst-case performance (maximum slowdown), {WC-EBBA*} lies between {WC-A*} and {WC-BA*}, but still performs far better than {WC-A*} mainly due to its bidirectional framework.
    However, compared to the other bidirectional algorithm {WC-BA*}, we can see that {WC-EBBA*} gradually becomes weaker when we move from tight (left) to loose constraints (right) in both median and mean slowdown factors.
    There is one potential reason for this pattern.
    {WC-EBBA*}'s preliminary heuristics are more informed than those of {WC-BA*} due to sequential heuristic searches.
    Therefore, {WC-EBBA*} prunes more nodes and performs faster in very tight constraints.
    In loose constraints, however, the sequential searches (in both initialisation and the constrained search) become the lengthy part of {WC-EBBA*} and lead to increasing the runtime.
    \item
    {WC-EBBA*}\textsubscript{par}:
    This algorithm can be seen as the best performer across the 40-80\% range.
    It also shows the best average factor in the 50-80\% tightness range.
    Compared to its standard version {WC-EBBA*}, the parallel variant significantly improves the runtime and worst-case performance (maximum factors) on almost all levels of tightness, especially in loose constraints.
    This is mainly because of faster initialisation via parallelism.
    However, {WC-EBBA*}\textsubscript{par} is slightly weaker than its sequential version in the 10\% constraint, mainly because its heuristic functions are not as informed as those of {WC-EBBA*} with sequential searches.
    Having compared {WC-EBBA*}\textsubscript{par} with the other parallel search algorithm {WC-BA*}, we can see that {WC-EBBA*}\textsubscript{par} outperforms {WC-BA*} in loosely constrained instances (the 40-80\% range), whereas {WC-BA*} performs better on average in the 10-30\% range.
    A potential reason for this behaviour is that, unlike {WC-BA*} where its backward search has limited impact on reducing the search space in loose constraints,
    bidirectional searches of {WC-EBBA*}\textsubscript{par} work in the same objective ordering, and thus they actively contribute to reducing the (already small) search space. 
    
\end{itemize}
\textbf{Summary:}
{WC-EBBA*} is a good candidate for very tight constraints, as it benefits from more informed heuristics via sequential preliminary searches.
{WC-BA*} can be seen as a great candidate for tightly constrained problem instances. Its backward search is very effective in reducing the search space in such cases.
{WC-EBBA*}\textsubscript{par} works very well on loosely constrained problems, mainly because its bidirectional searches are both capable of reducing the search space.
Finally, {WC-A*} can be seen as a good candidate for very loose constraints where the search space is already small and does not need better informed heuristics.

\textbf{A different observation:}
We focus on the performance of {WC-A*} and {WC-EBBA*}.
\citeauthor{thomas2019exact} reported that the unidirectional forward A* search is less effective than the bidirectional search of {RC-BDA*} on the WCSPP instances of \citet{santos2007improved}.
In their paper, the forward A* search is reportedly 20 times slower than {RC-BDA*} on average.
It is also outperformed by {RC-BDA*} in all tested levels of tightness.
They concluded that the bidirectional nature of {RC-BDA*} plays a critical role in {RC-BDA*}'s success.
Given {WC-EBBA*} as the enhanced weight constrained version of {RC-BDA*}, if we assume our {WC-A*} is a unidirectional variant of {WC-EBBA*}, the experimental results in Figure~\ref{fig:boxplot_slowdown} show that the bidirectional search does not always yield smaller runtimes.
As we discussed above, there are cases in which unidirectional {WC-A*} delivers outstanding performance (in both median and mean runtimes) compared to the bidirectional {WC-EBBA*} (in the 60-80\% range).
However, we can confirm that the unidirectional variant is still dominated by the bidirectional variant in tight constraints.
%
\subsection{On the Importance of Initialisation}
\label{sec:initilaisation_importance}
In order to investigate the significance of preliminary searches in the initialisation phase of our non-parallel algorithms {WC-A*} and {WC-EBBA*}, we implemented a version of these two algorithms where the order of preliminary searches is changed to:\\
(1) Run standard A* on $\mathit{cost_2}$ to obtain an initial upper bound on $\mathit{cost_1}$ and initialise $\overline{f_1}$\\
(2) Run bounded A* on $\mathit{cost_1}$ using the upper bound obtained in the previous step \\
(3) Run bounded A* on $\mathit{cost_2}$ using the weight limit as the upper bound, and only using the states expanded in the previous step 

\begin{figure}[t]
\centering
\includegraphics[width=1\textwidth]{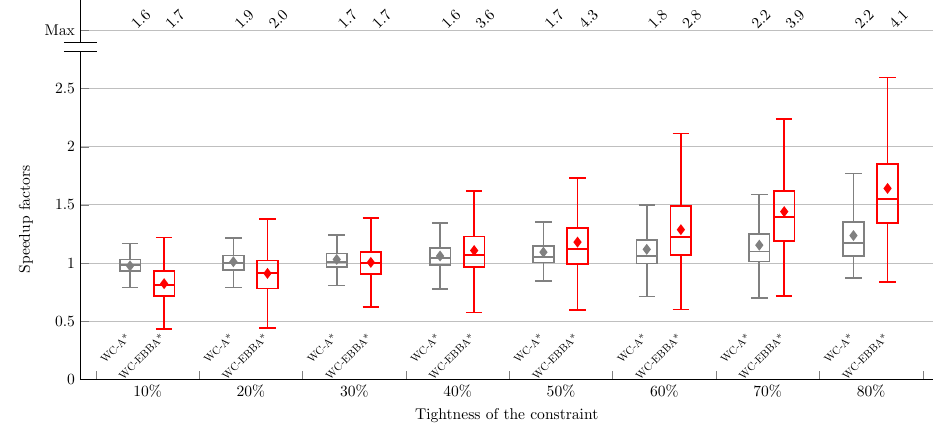}
\caption{\small The distribution of speedup factors achieved by changing the order of preliminary searches in the initialisation phase of {WC-A*} and {WC-EBBA*} (outliers are not shown).}
\label{fig:RC_A_improve}
\end{figure}

In this ordering, the bounded search on $\mathit{cost_2}$ is done in the last step.
Note that {WC-EBBA*} requires steps~(2) and (3) above to be done in both directions.
Moreover, the bounded search of step~(3), in one direction, can be seen as a continuation of step~(1).
With this extension, we aim to analyse the performance of both {WC-A*} and {WC-EBBA*} algorithms with heuristic functions of different qualities.
For this analysis, we evaluate both variants of the algorithms on all of our benchmark instances.
We perform three runs of each variant and store the run showing the median runtime, and then calculate the speedup factors achieved by using the new variant.
The speedup factor is larger than one if the new variant runs faster (with smaller runtime) than the standard version.
We have 250 instances in each level of tightness, and unsolved cases are considered to have a runtime of one hour. 
Figure~\ref{fig:RC_A_improve} shows the distribution of speedup factors achieved across all levels of tightness with respect to the standard variants.

The results in Figure~\ref{fig:RC_A_improve} highlight that {WC-EBBA*} is much more affected by the changes in the initialisation phase than {WC-A*}.
Looking at the range of speedups achieved in tight constraints (the 10-30\% range), we notice that both algorithms become weaker when we perform bounded searches on $\mathit{cost_1}$ earlier than $\mathit{cost_2}$.
In particular, the detailed results illustrate that the new strategy can make {WC-EBBA*} up to six times slower in the 10\% tightness level.
However, based on the pattern, the new search order (starting bounded searches with $\mathit{cost_1}$) becomes more effective when we move from tight (left) to loose constraints (right).
More precisely, in the 40-80\% range, we see an improvement in the performance of the algorithms (via both median and mean speedup factors) with the new search order.
In the 80\% tightness level, for example, we see {WC-EBBA*} performing 60\% better than its standard version on average.
There are two potential reasons for having such a pattern. 
Firstly, the upper bound on $\mathit{cost_2}$ (the weight limit) is smaller in tight constraints, so starting with the bounded search on $\mathit{cost_2}$ would be more effective in reducing the graph size in such constraints.
Secondly, our heuristic function (spherical distance) is more informed in the bounded search on $\mathit{cost_1}$ (distance) than $\mathit{cost_2}$ (time).
As a result, in loose constraints with larger upper bounds on $\mathit{cost_2}$, we can expect that bounded search on $\mathit{cost_1}$ performs better than $\mathit{cost_2}$ in graph reduction.
In terms of the best and worst-case performances in very loose constraints (70\% and 80\%), we see maximum speedup factors around two and four for {WC-A*} and {WC-EBBA*}, respectively, but maximum slowdown factors around 1.15-1.45 for both algorithms.
This means that the worst-case performance of the new variants is far less severe in loose constraints.

We can conclude that, depending on the quality of heuristics, both {WC-A*} and {WC-EBBA*} (with an admissible heuristic on $\mathit{cost_1}$) can benefit from the proposed (reversed) search order for their initialisation phase in loosely constrained instances to perform faster, but such a setting is less effective in tightly constrained problems. 
Based on this observation, we recommend using procedures that can change the order of bounded searches in the initialisation phase based on the tightness of the constraint.
%
\subsection{Extended Experiments: Tie-breaking and Priority Queues}
\label{sec:experiment_extended}
We now study the significance of priority queues in constrained search with A* and show how disabling tie-breaking and using bucket-based queues can together contribute to substantial improvement in algorithmic performance.
Earlier in Section~\ref{sec:practicall}, we presented two types of bucket-based queues for the WCSPP: bucket queues with linked lists, and hybrid queues with binary heaps.
For the analysis of this section, we also consider binary heaps as the conventional queuing method.
Obviously, both binary heaps and hybrid queues (with binary heaps) can order nodes with and without tie-breaking.
For both bucket and hybrid queues, we set the bucket width to be $\Delta f =1$.
Bucket queues (with linked lists), however, are not able to break the tie between nodes in cases two nodes have the same primary cost.
For this category of queues, we study LIFO and FIFO node extraction strategies instead.

For this extended set of experiments, we choose the unidirectional search algorithm {WC-A*} as it offers a more stable performance than our bidirectional algorithms in terms of runtime and the number of node expansions.
For our benchmark instances, since the objectives (distance and time) in the DIMACS road networks are highly correlated, we design a set of randomised graphs by changing the $\mathit{cost_2}$ of the edges in the DIMCAS maps (the time attribute) with random (integer) values in the $[1,10000]$ range.
We do not change instance specifications, i.e., (origin, destination) pairs and the tightness values.
The new set of (randomised) graphs will help us to investigate the queuing strategies in cases where the search generates more non-dominated nodes with random costs.
We ran all experiments on the machine described earlier in the section and with a one-hour timeout (the runtime of unsolved cases).
The runtimes we report for this extended experiment are the median of three runs for each instance.
We present the cumulative runtime of {WC-A*} with all types of studied priority queues over the instances of each map in Table~\ref{table:runtime_tie_org_map} for the original DIMACS networks and in Table~\ref{table:runtime_tie_rand_map} for the DIMACS graphs with random weights.
\begin{table}[!t]
\caption{\small Cumulative runtime of the WC-A* algorithm (in \textit{minutes}) with different types of priority queues per map (\textbf{normal weights}). We report total time needed to solve all cases of the instance and consider the timeout (1-hour) as the runtime of unsolved cases.  }
\label{table:runtime_tie_org_map}
\small
\begin{adjustbox}{width=1\textwidth}
\renewcommand{\arraystretch}{1}
\begin{tabular}{lrrrrrrrrrrrr}
\headrow
Queue / Map           & \multicolumn{1}{c}{NY} & \multicolumn{1}{c}{BAY} & \multicolumn{1}{c}{COL} & \multicolumn{1}{c}{FLA} & \multicolumn{1}{c}{NW} & \multicolumn{1}{c}{NE} & \multicolumn{1}{c}{CAL} & \multicolumn{1}{c}{LKS} & \multicolumn{1}{c}{E} & \multicolumn{1}{c}{W} & \multicolumn{1}{c}{CTR} & \multicolumn{1}{c}{USA} \\
Bucket-LIFO      & 0.16 & 0.24 & 0.57 & 3.54 & 4.38  & 4.36  & 10.85 & 68.29  & 107.25 & 89.50  & 208.19 & 2025.00 \\
Bucket-FIFO      & 0.16 & 0.25 & 0.59 & 4.01 & 4.82  & 4.81  & 14.55 & 72.82  & 113.27 & 109.36 & 222.44 & 2205.61 \\
Hybrid w/o tie   & 0.16 & 0.26 & 0.64 & 4.24 & 5.28  & 5.22  & 13.87 & 80.88  & 130.08 & 103.65 & 248.62 & 2331.18 \\
Hybrid w tie     & 0.17 & 0.28 & 0.71 & 5.04 & 6.49  & 6.48  & 16.64 & 114.37 & 174.41 & 146.05 & 313.83 & 2875.61 \\
Bin-Heap w/o tie & 0.21 & 0.37 & 1.06 & 8.43 & 11.25 & 11.24 & 31.87 & 225.36 & 348.82 & 278.86 & 602.25 & 4189.57 \\
Bin-Heap w tie   & 0.21 & 0.38 & 1.07 & 9.09 & 12.00 & 12.63 & 35.62 & 257.55 & 395.29 & 303.87 & 681.50 & 4600.02\\
\end{tabular}
\end{adjustbox}
\end{table}
\begin{table}[t]
\caption{\small Cumulative runtime of the WC-A* algorithm (in \textit{minutes}) with different types of priority queues per map (\textbf{random weights}). We report total time needed to solve all cases of the instance and consider the timeout (1-hour) as the runtime of unsolved cases.}
\label{table:runtime_tie_rand_map}
\small
\begin{adjustbox}{width=1\textwidth}
\renewcommand{\arraystretch}{1}
\begin{tabular}{lrrrrrrrrrrrr}
\headrow
Queue / Map           & \multicolumn{1}{c}{NY} & \multicolumn{1}{c}{BAY} & \multicolumn{1}{c}{COL} & \multicolumn{1}{c}{FLA} & \multicolumn{1}{c}{NW} & \multicolumn{1}{c}{NE} & \multicolumn{1}{c}{CAL} & \multicolumn{1}{c}{LKS} & \multicolumn{1}{c}{E} & \multicolumn{1}{c}{W} & \multicolumn{1}{c}{CTR} & \multicolumn{1}{c}{USA} \\
          
Bucket-LIFO      & 0.46 & 0.39 & 1.51 & 44.61  & 13.13 & 41.75  & 46.98  & 763.38  & 324.08  & 683.44  & 2800.15 & 6476.14 \\
Bucket-FIFO      & 0.49 & 0.42 & 1.54 & 42.45  & 13.86 & 44.63  & 46.05  & 757.41  & 328.41  & 693.51  & 2807.20  & 6557.63 \\
Hybrid w/o tie   & 0.52 & 0.43 & 1.73 & 49.18  & 16.46 & 47.63  & 54.44  & 873.85  & 377.23  & 759.35  & 2953.02 & 6685.09 \\
Hybrid w tie     & 0.63 & 0.48 & 2.09 & 65.48  & 20.41 & 68.16  & 69.86  & 1174.14 & 529.29  & 1067.34 & 3194.13 & 7138.32 \\
Bin-Heap w/o tie & 0.94 & 0.66 & 3.40  & 127.65 & 37.91 & 137.14 & 130.24 & 1724.87 & 929.88  & 1761.51 & 3666.08 & 7951.32 \\
Bin-Heap w tie   & 1.03 & 0.69 & 3.51 & 132.23 & 37.36 & 150.44 & 133.88 & 1870.27 & 1050.45 & 1894.05 & 3837.63 & 8217.28

\end{tabular}
\end{adjustbox}
\end{table}
\textbf{Tie-breaking impact:}
Comparing the performance of {WC-A*} in Tables~\ref{table:runtime_tie_org_map} and \ref{table:runtime_tie_rand_map} in both settings, i.e., with and without tie-breaking, we can see that the algorithm performs better if we simply do not break ties in the hybrid and binary heap priority queues.
For the one-level bucket queue with linked lists, the results show that {WC-A*} with the LIFO strategy outperforms {WC-A*} with the FIFO strategy in almost all maps.
There is one potential reason for this observation.
In the LIFO strategy, recent insertions (in each bucket) appear earlier in the queue.
Since recent insertions are normally more informed than earlier insertions, the search potentially prunes more dominated nodes in the LIFO strategy.
Interestingly, the detailed results show that {WC-A*} with worst-case tie-breaking (in bucket-queue with the FIFO strategy) leads to up to 100\% extra expansions but still runs faster than the variant with tie-breaking (hybrid queue and binary heap with tie-breaking).
Further, hybrid queues without tie-breaking are not as effective as bucket queues with linked lists, mainly due to overheads in the two-level bucket-heap data structure.
We have provided more detailed analyses on the impacts of tie-breaking in \ref{sec:PQ_impact_extended}.

\textbf{Queue type impact:}
Comparing the performance of {WC-A*} based on the priority queues in Tables~\ref{table:runtime_tie_org_map} and \ref{table:runtime_tie_rand_map}, we see that bucket queues are consistently faster than the other queue types across all maps, whereas the hybrid queues are ranked second in the head-to-head comparison between the three priority queue types in both tables.
Nonetheless, {WC-A*} with the hybrid queue is still significantly faster than {WC-A*} with the conventional binary heap queue.
In particular, even the hybrid queue with tie-breaking shows better cumulative runtimes than the binary heap without tie-breaking.
This observation highlights the effectiveness of using bucket-based queues in the exhaustive search of {WC-A*}.

\textbf{Memory:}
We did not observe any significant difference in the memory usage of {WC-A*} with the three queue types, as they normally contain nearly the same number of nodes over the course of the search.
Nonetheless, hybrid queues can be seen as slightly more efficient than binary heap queues in terms of memory use, as the low-level binary heap in the hybrid queue handles a portion of the total nodes and thus is less likely to grow into a big list that contains all nodes (as in conventional binary heaps). 

\begin{figure}[!t]
\begin{subfigure}{0.49\textwidth}
\includegraphics[width=1\textwidth]{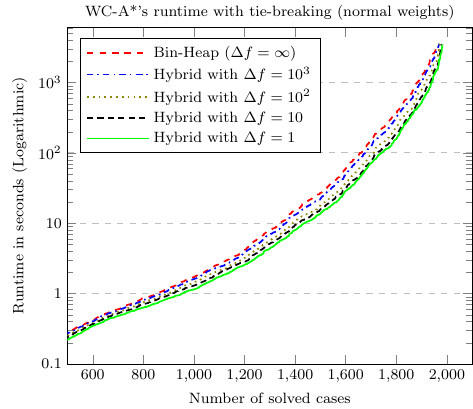} 
\end{subfigure}
\vspace{0.5 \baselineskip}
\hfill
\begin{subfigure}{0.49\textwidth}
\includegraphics[width=1\textwidth]{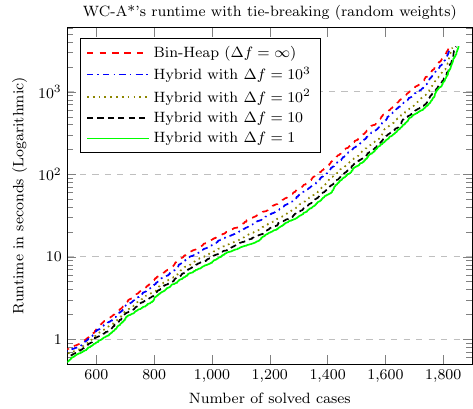}
\end{subfigure}
\begin{subfigure}{0.49\textwidth}
\includegraphics[width=1\textwidth]{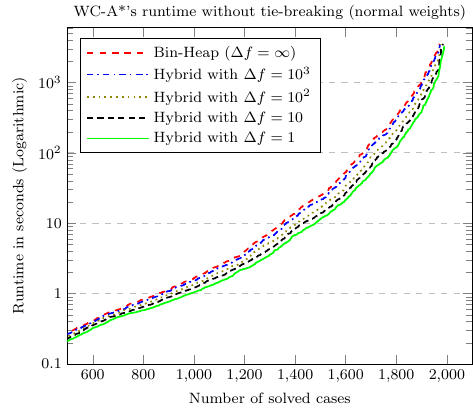} 
\end{subfigure}
\hfill
\begin{subfigure}{0.49\textwidth}
\includegraphics[width=1\textwidth]{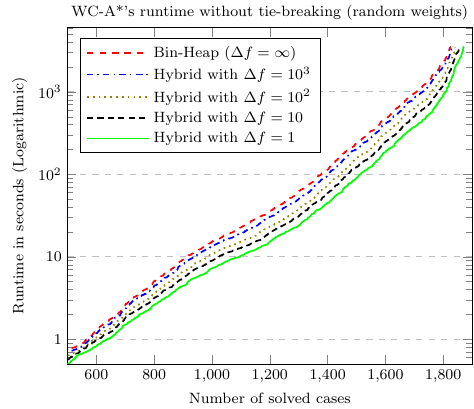}
\end{subfigure}
\caption{\small Cactus plots of {WC-A*}'s performance with hybrid queues of different widths with and without tie-breaking on the original DIMACS graphs (left) and randomised graphs (right). The first 500 easy instances are not shown.}
\label{fig:pqueue_hybrid_width}
\end{figure}

\textbf{Bucket width impacts:}
For our last experiment in this paper, we study the impact of the bucket width on the search performance.
In particular, we are interested in cases where buckets are required to order nodes based on their primary cost, as in problem instances with non-integer costs via hybrid queues.
To this end, we evaluated the performance of {WC-A*} on both realistic and randomised graphs using bucket widths $\Delta f \in \{10,100,1000\}$ in the hybrid queue, with and without tie-breaking.
Figure~\ref{fig:pqueue_hybrid_width} shows the results for the given (increased) bucket widths, along with the two extreme cases $\Delta f=1$ and $\Delta f = \infty$ (i.e., binary heap).
To better see the difference between the plots, we do not show the first 500 instances.
We can see from the results in Figure~\ref{fig:pqueue_hybrid_width} that increasing the bucket width adversely affects the algorithmic performance of {WC-A*} on both graph types.
In both variants (with and without tie-breaking), {WC-A*} solves fewer instances if we increase the bucket width to 10, 100 or 1000.
The pattern also shows that {WC-A*}'s performance with the hybrid queue becomes closer to that of the binary heap queue if we further increase the bucket width $\Delta f$.
A potential reason for this behaviour is related to the number of nodes in each bucket of the hybrid queue.
Increasing the bucket width means covering a larger range of $f_p$-values, and consequently, dealing with more nodes in each bucket.
As a result, high-level buckets become more populated and the low-level binary heap operations would take longer.
\section{Conclusion and Future work}
\label{sec:conclusion}
This paper presented two new solution approaches to the hard Weight Constrained Shortest Path Problem (WCSPP): {WC-A*}  and {WC-EBBA*}\textsubscript{par}.
{WC-A*} is a unidirectional algorithm derived from the techniques used in recent bi-objective search algorithms BOA* and BOBA*.
Our second algorithm, {WC-EBBA*}\textsubscript{par}, is the extended version of the state-of-the-art bidirectional constrained search algorithm {WC-EBBA*}, enhanced with parallelism.
We evaluated our WCSPP algorithms {WC-A*} and {WC-EBBA*}\textsubscript{par} on very large graphs through a set of 2000 realistic instances and compared their performance against \textit{six} recent WCSPP algorithms in the literature
, namely CSP \cite{sedeno2015enhanced}, the award-winning algorithms {BiPulse} \cite{cabrera2020exact} and {RC-BDA*} \cite{thomas2019exact}, and our recent algorithms {WC-EBBA*} \cite{AhmadiTHK21} and {WC-BA*} \cite{AhmadiTHK22_socs}.
The results show that the new A*-based algorithms of this paper are effective in improving the runtime over the state-of-the-art algorithms.
In particular, they both outperform BiPulse in almost all instances by showing up to two orders of magnitude faster runtime on average.
We summarise the strengths and weaknesses of each algorithm as follows.
\begin{itemize}[nosep]
    \item {WC-A*}: Very fast on loose constraints and excellent in memory efficiency.
    This algorithm uses a unidirectional search strategy and can be considered an effective solution approach for problems that cannot be solved bidirectionally.
    {WC-A*} outperforms the state-of-the-art {WC-EBBA*} algorithm on problems with relatively small search space.
    In addition, it is very space efficient and consumes up to three times less memory than the other studied algorithms.
    However, compared to bidirectional algorithms, it performs poorly on tightly constrained instances and shows more critical worst cases. 
    \item {WC-EBBA*}\textsubscript{par}: Very fast on almost all levels of tightness.
    This algorithm can be a great choice for applications that need fast solution approaches.
    {WC-EBBA*}\textsubscript{par} considerably improves the standard version {WC-EBBA*} in both runtime and worst-case performance, especially in instances with non-tight constraints, and solves more instances in a limited time than its fast competitors.
    However, it shows the worst performance in terms of memory usage among the other A*-based algorithms due to its space-demanding path-matching procedure.
\end{itemize}

We also investigated the importance of the initialisation phase in the performance of constrained search with A*.
By changing the order of preliminary heuristic searches, we realised that prioritising the attribute with more informed heuristic can contribute to performing several factors faster in a specific range of constraints, but likely slower in the remaining tightness levels.
Therefore, a better initialisation procedure can be the one that decides search ordering based on the tightness of the constraint.

Furthermore, we studied two main components of the constrained search with A*: priority queue and lexicographical ordering (tie-breaking).
We showed via extensive experiments on realistic and randomised graphs that bi-criteria A* algorithms with lazy dominance procedures can take advantage of bucket-based queues to expedite the queue operations.
In addition, we empirically proved that although lexicographical ordering of search objects in the priority queue is generally supposed to be a standard method for node expansion in the literature, we can achieve far better runtimes if we just order search objects based on their primary cost (without tie-breaking).
The results of our experiments on large graphs show that the priority queues' effort in breaking ties among search objects in the queue is not paid off by delivering smaller runtimes, as the tie-breaking overhead is normally far higher than the number of pruned objects via tie-breaking.
We also studied the impact of bucket width on search performance and observed that, regardless of tie-breaking, hybrid queues with sparsely distributed nodes (via smaller bucket widths) can be more effective than hybrid queues with fewer but (densely) populated buckets.

\textbf{Future work:}
We already observed that improving the quality of search heuristics can potentially lead to faster runtime for {WC-A*} and {WC-EBBA*} in loose constraints.
An interesting direction for future work can be applying the graph reduction technique of branch-and-bound methods to the initialisation phase of our A*-based algorithms.
For example, we can run several rounds of forward-backward bounded search to remove more nodes from the graph and obtain better quality heuristics for the main search.
We currently limit the number of preliminary heuristic searches to at most two rounds.
As a further optimisation, we can improve the bidirectional heuristics of {WC-EBBA*} and {WC-EBBA*}\textsubscript{par} during the search, like heuristic tuning in {WC-BA*}.
However, as both searches of these algorithms are conducted on the same objective ordering, we can tune primary heuristics instead.
In such a setting, we need to make sure that the consistency requirement of the A* heuristic function is still satisfied.

\bibliography{References}

\begin{thebibliography}{35}
\providecommand{\natexlab}[1]{#1}
\providecommand{\url}[1]{\texttt{#1}}
\providecommand{\urlprefix}{}

\bibitem[{Handler and Zang(1980)Gabriel Y. Handler and Israel
  Zang}]{handler1980dual}
Handler GY, Zang I.
\newblock A dual algorithm for the constrained shortest path problem.
\newblock Networks 1980;10(4):293--309.
\newblock \urlprefix\url{https://doi.org/10.1002/net.3230100403}.

\bibitem[{Laporte and Pascoal(2011)Gilbert Laporte and Marta M. B.
  Pascoal}]{LaporteP11}
Laporte G, Pascoal MMB.
\newblock Minimum cost path problems with relays.
\newblock Comput Oper Res 2011;38(1):165--173.
\newblock \urlprefix\url{https://doi.org/10.1016/j.cor.2010.04.010}.

\bibitem[{Zhu and Wilhelm(2012)Xiaoyan Zhu and Wilbert E.
  Wilhelm}]{zhu2012three}
Zhu X, Wilhelm WE.
\newblock A three-stage approach for the resource-constrained shortest path as
  a sub-problem in column generation.
\newblock Comput Oper Res 2012;39(2):164--178.
\newblock \urlprefix\url{https://doi.org/10.1016/j.cor.2011.03.008}.

\bibitem[{Storandt(2012)Sabine Storandt}]{Storandt12}
Storandt S.
\newblock Quick and energy-efficient routes: computing constrained shortest
  paths for electric vehicles.
\newblock In: Winter S, M{\"{u}}ller{-}Hannemann M, editors. 5th {ACM}
  {SIGSPATIAL} International Workshop on Computational Transportation Science
  2011, CTS'12, November 6, 2012, Redondo Beach, CA, {USA} {ACM}; 2012. p.
  20--25.
\newblock \urlprefix\url{https://doi.org/10.1145/2442942.2442947}.

\bibitem[{Ahmadi et~al.(2021)Saman Ahmadi and Guido Tack and Daniel Harabor and
  Philip Kilby}]{AhmadiTHK21_cp}
Ahmadi S, Tack G, Harabor D, Kilby P.
\newblock Vehicle Dynamics in Pickup-And-Delivery Problems Using Electric
  Vehicles.
\newblock In: Michel LD, editor. 27th International Conference on Principles
  and Practice of Constraint Programming, {CP} 2021, Montpellier, France
  (Virtual Conference), October 25-29, 2021, vol. 210 of LIPIcs Schloss
  Dagstuhl - Leibniz-Zentrum f{\"{u}}r Informatik; 2021. p. 11:1--11:17.
\newblock \urlprefix\url{https://doi.org/10.4230/LIPIcs.CP.2021.11}.

\bibitem[{Pugliese and Guerriero(2013)Luigi Di Puglia Pugliese and Francesca
  Guerriero}]{pugliese2013survey}
Pugliese LDP, Guerriero F.
\newblock A survey of resource constrained shortest path problems: Exact
  solution approaches.
\newblock Networks 2013;62(3):183--200.
\newblock \urlprefix\url{https://doi.org/10.1002/net.21511}.

\bibitem[{Santos et~al.(2007)Luis Santos and João Coutinho-Rodrigues and John
  R. Current}]{santos2007improved}
Santos L, Coutinho-Rodrigues J, Current JR.
\newblock An improved solution algorithm for the constrained shortest path
  problem.
\newblock Transportation Research Part B: Methodological 2007;41(7):756 -- 771.
\newblock
  \urlprefix\url{http://www.sciencedirect.com/science/article/pii/S0191261507000124}.

\bibitem[{Dumitrescu and Boland(2003)Irina Dumitrescu and Natashia
  Boland}]{DumitrescuB03}
Dumitrescu I, Boland N.
\newblock Improved preprocessing, labeling and scaling algorithms for the
  Weight-Constrained Shortest Path Problem.
\newblock Networks 2003;42(3):135--153.
\newblock \urlprefix\url{https://doi.org/10.1002/net.10090}.

\bibitem[{Muhandiramge and Boland(2009)Ranga Muhandiramge and Natashia
  Boland}]{MuhandiramgeB09}
Muhandiramge R, Boland N.
\newblock Simultaneous solution of Lagrangean dual problems interleaved with
  preprocessing for the weight constrained shortest path problem.
\newblock Networks 2009;53(4):358--381.
\newblock \urlprefix\url{https://doi.org/10.1002/net.20292}.

\bibitem[{Ferone et~al.(2020)Daniele Ferone and Paola Festa and Serena Fugaro
  and Tommaso Pastore}]{FeroneFFP20}
Ferone D, Festa P, Fugaro S, Pastore T.
\newblock On the Shortest Path Problems with Edge Constraints.
\newblock In: 22nd International Conference on Transparent Optical Networks,
  {ICTON} 2020, Bari, Italy, July 19-23, 2020 {IEEE}; 2020. p. 1--4.
\newblock \urlprefix\url{https://doi.org/10.1109/ICTON51198.2020.9203378}.

\bibitem[{Lozano and Medaglia(2013)Leonardo Lozano and Andr{\'{e}}s L.
  Medaglia}]{lozano2013exact}
Lozano L, Medaglia AL.
\newblock On an exact method for the constrained shortest path problem.
\newblock Comput Oper Res 2013;40(1):378--384.
\newblock \urlprefix\url{https://doi.org/10.1016/j.cor.2012.07.008}.

\bibitem[{Horv{\'{a}}th and Kis(2016)Mark{\'{o}} Horv{\'{a}}th and Tam{\'{a}}s
  Kis}]{HorvathK16}
Horv{\'{a}}th M, Kis T.
\newblock Solving resource constrained shortest path problems with LP-based
  methods.
\newblock Comput Oper Res 2016;73:150--164.
\newblock \urlprefix\url{https://doi.org/10.1016/j.cor.2016.04.013}.

\bibitem[{Garcia(2009)Renan Garcia}]{Garcia09}
Garcia R.
\newblock Resource constrained shortest paths and extensions.
\newblock PhD thesis, Georgia Institute of Technology, Atlanta, GA, {USA};
  2009.

\bibitem[{Sede{\~{n}}o{-}Noda and Alonso{-}Rodr{\'{\i}}guez(2015)Antonio
  Sede{\~{n}}o{-}Noda and Sergio
  Alonso{-}Rodr{\'{\i}}guez}]{sedeno2015enhanced}
Sede{\~{n}}o{-}Noda A, Alonso{-}Rodr{\'{\i}}guez S.
\newblock An enhanced {K-SP} algorithm with pruning strategies to solve the
  constrained shortest path problem.
\newblock Appl Math Comput 2015;265:602--618.
\newblock \urlprefix\url{https://doi.org/10.1016/j.amc.2015.05.109}.

\bibitem[{Bol{\'{\i}}var et~al.(2014)Manuel A. Bol{\'{\i}}var and Leonardo
  Lozano and Andr{\'{e}}s L. Medaglia}]{BolivarLM14}
Bol{\'{\i}}var MA, Lozano L, Medaglia AL.
\newblock Acceleration strategies for the weight constrained shortest path
  problem with replenishment.
\newblock Optim Lett 2014;8(8):2155--2172.
\newblock \urlprefix\url{https://doi.org/10.1007/s11590-014-0742-x}.

\bibitem[{Thomas et~al.(2019)Barrett W. Thomas and Tobia Calogiuri and Mike
  Hewitt}]{thomas2019exact}
Thomas BW, Calogiuri T, Hewitt M.
\newblock An exact bidirectional A{\(\star\)} approach for solving
  resource-constrained shortest path problems.
\newblock Networks 2019;73(2):187--205.
\newblock \urlprefix\url{https://doi.org/10.1002/net.21856}.

\bibitem[{Righini and Salani(2006)Giovanni Righini and Matteo
  Salani}]{righini2006symmetry}
Righini G, Salani M.
\newblock Symmetry helps: Bounded bi-directional dynamic programming for the
  elementary shortest path problem with resource constraints.
\newblock Discret Optim 2006;3(3):255--273.
\newblock \urlprefix\url{https://doi.org/10.1016/j.disopt.2006.05.007}.

\bibitem[{Cabrera et~al.(2020)Nicol{\'{a}}s Cabrera and Andr{\'{e}}s L.
  Medaglia and Leonardo Lozano and Daniel Duque}]{cabrera2020exact}
Cabrera N, Medaglia AL, Lozano L, Duque D.
\newblock An exact bidirectional pulse algorithm for the constrained shortest
  path.
\newblock Networks 2020;76(2):128--146.
\newblock \urlprefix\url{https://doi.org/10.1002/net.21960}.

\bibitem[{Golden and Shier(2021)Golden, Bruce L. and Shier, Douglas
  R.}]{GloverPrize}
Golden BL, Shier DR.
\newblock 2019–2020 Glover-Klingman Prize Winners.
\newblock Networks 2021;n/a(n/a).
\newblock
  \urlprefix\url{https://onlinelibrary.wiley.com/doi/abs/10.1002/net.22072}.

\bibitem[{Ahmadi et~al.(2021)Saman Ahmadi and Guido Tack and Daniel Damir
  Harabor and Philip Kilby}]{AhmadiTHK21}
Ahmadi S, Tack G, Harabor DD, Kilby P.
\newblock A Fast Exact Algorithm for the Resource Constrained Shortest Path
  Problem.
\newblock In: Thirty-Fifth {AAAI} Conference on Artificial Intelligence, {AAAI}
  2021, Thirty-Third Conference on Innovative Applications of Artificial
  Intelligence, {IAAI} 2021, The Eleventh Symposium on Educational Advances in
  Artificial Intelligence, {EAAI} 2021, Virtual Event, February 2-9, 2021
  {AAAI} Press; 2021. p. 12217--12224.
\newblock
  \urlprefix\url{https://ojs.aaai.org/index.php/AAAI/article/view/17450}.

\bibitem[{Tilk et~al.(2017)Christian Tilk and Ann{-}Kathrin Rothenb{\"{a}}cher
  and Timo Gschwind and Stefan Irnich}]{TilkRGI17}
Tilk C, Rothenb{\"{a}}cher A, Gschwind T, Irnich S.
\newblock Asymmetry matters: Dynamic half-way points in bidirectional labeling
  for solving shortest path problems with resource constraints faster.
\newblock Eur J Oper Res 2017;261(2):530--539.
\newblock \urlprefix\url{https://doi.org/10.1016/j.ejor.2017.03.017}.

\bibitem[{Ahmadi et~al.(2022)Saman Ahmadi and Guido Tack and Daniel Harabor and
  Philip Kilby}]{AhmadiTHK22_socs}
Ahmadi S, Tack G, Harabor D, Kilby P.
\newblock Weight Constrained Path Finding with Bidirectional {A*}.
\newblock In: Chrpa L, Saetti A, editors. Proceedings of the Fifteenth
  International Symposium on Combinatorial Search, {SOCS} 2022, Vienna,
  Austria, July 21-23, 2022 {AAAI} Press; 2022. p. 2--10.
\newblock
  \urlprefix\url{https://ojs.aaai.org/index.php/SOCS/article/view/21746}.

\bibitem[{Ahmadi et~al.(2021{\natexlab{a}})Saman Ahmadi and Guido Tack and
  Daniel Harabor and Philip Kilby}]{AhmadiTHK21_esa}
Ahmadi S, Tack G, Harabor D, Kilby P.
\newblock Bi-Objective Search with Bi-Directional {A*}.
\newblock In: Mutzel P, Pagh R, Herman G, editors. 29th Annual European
  Symposium on Algorithms, {ESA} 2021, September 6-8, 2021, Lisbon, Portugal
  (Virtual Conference), vol. 204 of LIPIcs Schloss Dagstuhl - Leibniz-Zentrum
  f{\"{u}}r Informatik; 2021. p. 3:1--3:15.
\newblock \urlprefix\url{https://doi.org/10.4230/LIPIcs.ESA.2021.3}.

\bibitem[{Ahmadi et~al.(2021{\natexlab{b}})Saman Ahmadi and Guido Tack and
  Daniel Harabor and Philip Kilby}]{AhmadiTHK21_socs}
Ahmadi S, Tack G, Harabor D, Kilby P.
\newblock Bi-Objective Search with Bi-directional A* (Extended Abstract).
\newblock In: Ma H, Serina I, editors. Proceedings of the Fourteenth
  International Symposium on Combinatorial Search, {SOCS} 2021, Virtual
  Conference [Jinan, China], July 26-30, 2021 {AAAI} Press; 2021. p. 142--144.
\newblock
  \urlprefix\url{https://ojs.aaai.org/index.php/SOCS/article/view/18563}.

\bibitem[{Ulloa et~al.(2020)Carlos Hern{\'{a}}ndez Ulloa and William Yeoh and
  Jorge A. Baier and Han Zhang and Luis Suazo and Sven
  Koenig}]{ulloa2020simple}
Ulloa CH, Yeoh W, Baier JA, Zhang H, Suazo L, Koenig S.
\newblock A Simple and Fast Bi-Objective Search Algorithm.
\newblock In: Beck JC, Buffet O, Hoffmann J, Karpas E, Sohrabi S, editors.
  Proceedings of the Thirtieth International Conference on Automated Planning
  and Scheduling, Nancy, France, October 26-30, 2020 {AAAI} Press; 2020. p.
  143--151.
\newblock
  \urlprefix\url{https://aaai.org/ojs/index.php/ICAPS/article/view/6655}.

\bibitem[{Hart et~al.(1968)Peter E. Hart and Nils J. Nilsson and Bertram
  Raphael}]{hart1968formal}
Hart PE, Nilsson NJ, Raphael B.
\newblock A Formal Basis for the Heuristic Determination of Minimum Cost Paths.
\newblock {IEEE} Trans Syst Sci Cybern 1968;4(2):100--107.
\newblock \urlprefix\url{https://doi.org/10.1109/TSSC.1968.300136}.

\bibitem[{Aneja et~al.(1983)Yash P. Aneja and V. Aggarwal and K. P. K.
  Nair}]{AnejaAN83}
Aneja YP, Aggarwal V, Nair KPK.
\newblock Shortest chain subject to side constraints.
\newblock Networks 1983;13(2):295--302.
\newblock \urlprefix\url{https://doi.org/10.1002/net.3230130212}.

\bibitem[{Pohl(1971)Pohl, Ira}]{pohl1971bi}
Pohl I.
\newblock Bi-directional search.
\newblock Machine intelligence 1971;6:127--140.

\bibitem[{Pulido et~al.(2015)Francisco Javier Pulido and Lawrence Mandow and
  Jos{\'{e}}{-}Luis P{\'{e}}rez{-}de{-}la{-}Cruz}]{PulidoMP15}
Pulido FJ, Mandow L, P{\'{e}}rez{-}de{-}la{-}Cruz J.
\newblock Dimensionality reduction in multiobjective shortest path search.
\newblock Comput Oper Res 2015;64:60--70.
\newblock \urlprefix\url{https://doi.org/10.1016/j.cor.2015.05.007}.

\bibitem[{Dijkstra(1959)Edsger W. Dijkstra}]{dijkstra1959note}
Dijkstra EW.
\newblock A note on two problems in connexion with graphs.
\newblock Numerische Mathematik 1959;1:269--271.
\newblock \urlprefix\url{https://doi.org/10.1007/BF01386390}.

\bibitem[{Sede{\~{n}}o{-}Noda and Colebrook(2019)Antonio Sede{\~{n}}o{-}Noda
  and Marcos Colebrook}]{Sedeno-NodaC19}
Sede{\~{n}}o{-}Noda A, Colebrook M.
\newblock A biobjective Dijkstra algorithm.
\newblock Eur J Oper Res 2019;276(1):106--118.
\newblock \urlprefix\url{https://doi.org/10.1016/j.ejor.2019.01.007}.

\bibitem[{Dial(1969)Robert B. Dial}]{Dial69}
Dial RB.
\newblock Algorithm 360: shortest-path forest with topological ordering {[H]}.
\newblock Commun {ACM} 1969;12(11):632--633.
\newblock \urlprefix\url{https://doi.org/10.1145/363269.363610}.

\bibitem[{Denardo and Fox(1979)Eric V. Denardo and Bennett L. Fox}]{DenardoF79}
Denardo EV, Fox BL.
\newblock Shortest-Route Methods: 1. Reaching, Pruning, and Buckets.
\newblock Oper Res 1979;27(1):161--186.
\newblock \urlprefix\url{https://doi.org/10.1287/opre.27.1.161}.

\bibitem[{Cherkassky et~al.(1996)Boris V. Cherkassky and Andrew V. Goldberg and
  Tomasz Radzik}]{CherkasskyGR96}
Cherkassky BV, Goldberg AV, Radzik T.
\newblock Shortest paths algorithms: Theory and experimental evaluation.
\newblock Math Program 1996;73:129--174.
\newblock \urlprefix\url{https://doi.org/10.1007/BF02592101}.

\bibitem[{Miettinen(1998)Kaisa Miettinen}]{Miettinen98Nonlinear}
Miettinen K.
\newblock Nonlinear multiobjective optimization, vol.~12 of International
  series in operations research and management science.
\newblock Kluwer; 1998.

\end{thebibliography}
\clearpage
\setcounter{page}{1}

\appendix
\def\appendix{\setcounter{section}{0}}
\renewcommand\thefigure{SM.~\arabic{figure}} 
\renewcommand\thetable{SM.~\arabic{table}} 
\def\thesection{SM.~\arabic{section}}
\setcounter{figure}{0}  
\setcounter{table}{0}
\setcounter{algocf}{0}
\renewcommand{\thealgocf}{SM.~\arabic{algocf}}

\section*{Supplementary Material}
\section{On the Correctness of Constrained Search with~A*}
\label{sec:A_star_correctness}
Given the notation and basics of constrained search presented in Section~\ref{sec:notation}, we discuss in more detail the correctness of our A*-based constrained search method in this section.
We start with one of the main properties of expanded nodes in A*.
\begin{lemma}
\label{lemma:astar}
Suppose A* is led by (smallest) $f_p$-values.
Let $(x_1,x_2,...,x_t)$ be the sequence of nodes expanded by A*. 
Then, if $h_p$ is consistent, $i \leq j$ implies $f_p(x_i) \leq f_p(x_j)$ \cite{hart1968formal}.
\end{lemma}
\begin{corollary}
\label{corollary:astar}
Under the premises of Lemma~\ref{lemma:astar}, if A* expands a node $x$ in direction~$d$, then every node $y$ expanded after $x$ (in the same search direction) would have $f_p(x) \leq f_p(y)$.
\end{corollary}
The non-decreasing order of expanded nodes in A* allows the search to detect \textit{dominated} nodes in constant time.
To achieve this fast dominance check, for each $u \in S$, A* in direction~$d$ needs to keep track of the secondary cost (i.e., $g_s$-value) of the last non-dominated node expanded for state $\mathit{u}$ via a parameter called $g^d_{\mathit{min}}(u)$.
In other words, $g^d_{\mathit{min}}(u)$ maintains the smallest $g_s$-value of previously expanded nodes for state $\mathit{u}$.
$g^d_{\mathit{min}}(u)$ is initialised to infinity, so the first node expanded with $\mathit{u}$ is considered non-dominated.
We formalise this dominance rule in Lemma~\ref{lemma:last_sol}.
\begin{lemma}
\label{lemma:last_sol}
Suppose nodes in $\mathit{Open}^d$ are ordered based on $f_p$-values.
Let $y$ be a node generated with state $\mathit{u}$ in direction~$d$ and let $g^d_{\mathit{min}}(u)$ be the $g_s$-value of the last node expanded by A* with state $\mathit{u}$.
$y$ is a weakly dominated node if $g^d_{\mathit{min}}(u) \leq g_s(y)$.
\end{lemma}
\begin{proof}
Let $x$ be the last node expanded with state $\mathit{u}$ by A* in direction~$d$, i.e., we have $s(y)=s(x)=u$ and $g^d_{\mathit{min}}(u)=g_s(x)$.
Since node $x$ has been expanded before node $y$, according to Lemma~\ref{lemma:astar}, A* guarantees that $f_p(x) \leq f_p(y)$.
In addition, since both nodes are associated with the same state $\mathit{u}$, they have used the same heuristic estimate (from state $\mathit{u}$ to the target state) to establish their $f_p$-values and we consequently have $g_p(x) \leq g_p(y)$.
Therefore, node $x$ weakly dominates node $y$ if $g_s(x) \leq g_s(y)$, or equivalently, if $g^d_{\mathit{min}}(u) \leq g_s(y)$.
\end{proof}

\textbf{Heuristic Functions:} 
Each heuristic function ${\bf h}^d$ estimates lower bounds on {\bf cost} of paths from each $u \in S$ to a target state in direction~$d$, so we need one such function for each constrained A* search if the algorithm is bidirectional.
Conventionally, ${\bf h}^d$ can be established by conducting two simple unidirectional single-objective searches from the target in the reverse direction~$d'$, one on $\mathit{cost_1}$ and the other one on $\mathit{cost_2}$.
For example, if a forward (constrained) A* search is intended, we can find lower bounds from all states to $\mathit{goal}$ in graph $G$ by simply running two individual one-to-all backward searches (from $\mathit{goal}$ on $\mathit{Reversed}(G)$), using Dijkstra's algorithm \cite{dijkstra1959note} for example, to establish $h^f_1(u)$ and $h^f_2(u)$ for every $u \in S$. 
We formally prove that the resulting heuristic functions are admissible and consistent.\par
\begin{lemma}
\label{lemma:admissible}
Let $d$ be the search direction of A*.
Computing the shortest paths on $\mathit{cost_p}$ (either of the edge attributes) via a unidirectional one-to-all search algorithm in the opposite direction~$d'$ yields a consistent and admissible heuristic function $h^d_p$.
\end{lemma}
\begin{proof}
We prove this lemma by assuming the contrary, namely that $h^d_p$ is not admissible or not consistent.
Assume that the Dijkstra's algorithm, as a correct one-to-all shortest path algorithm, has established our heuristic function $h^d_p$ via a unidirectional search on the attribute $\mathit{cost_p}$ in direction~$d'$.
If $h^d_p$ is not admissible, then there exists an optimal path $\pi$ from state $\mathit{u}$ to target in the graph $G$ for which we have $h^d_p(u) > \mathit{cost_p}(\pi)$.
However, the existence of such a path contradicts the correctness of Dijkstra's algorithm, mainly because one could use the path $\pi$ to further improve the Dijkstra's path (reduce the cost) to state $\mathit{u}$.
If $h^d_p$ is not consistent, then there exists an edge $(u,v) \in E$ for which we have $h^d_p(u) > \mathit{cost_p}(u,v) + h^d_p(v)$.
However, the existence of such an edge again contradicts our assumption of the correctness of Dijkstra's algorithm, mainly because one can use the edge $(u,v)$ to further reduce the cost of the shortest path to state $\mathit{u}$ via the intermediate state $\mathit{v}$.
Therefore, computing the shortest paths via a (correct) one-to-all search algorithm always yields an admissible and consistent heuristic function.
\end{proof}

\textbf{Upper Bound Functions:} 
The unidirectional searches we run from the target state not only give us the required heuristics of A*, but they also compute another set of useful costs, called \textit{state upper bounds}.
Recall our forward (constrained) A* search in the example above.
For a typical state $u \in S$, when we find its complementary shortest path $\pi$ (between $\mathit{u}$ and $\mathit{goal}$ on $G$) via a backward search solely on $\mathit{cost_1}$, we set $h^f_1(u) \leftarrow \mathit{cost_1}(\pi)$.
However, the $\mathit{cost_1}$-optimal path $\pi$ has a (potentially non-optimal) cost component $\mathit{cost_2}(\pi)$, which is the (secondary) cost of the path $\pi$ measured with the second attribute $\mathit{cost_2}$. While it may not be optimal, it is certainly an upper bound on the secondary cost of an optimal path.
We store this cost for state $\mathit{u}$ as $\mathit{ub}^f_2(u) \leftarrow \mathit{cost_2}(\pi)$. 
The same applies in the other one-to-all search on $\mathit{cost_2}$ when the single-objective search computes the $\mathit{ub}^f_1$-values.
Therefore, we can establish our upper bound functions ${\bf ub}^d: S \rightarrow \mathbb{R}^+ \times \mathbb{R}^+$ at the same time we build the lower bound functions ${\bf h}^d$.

But how do these non-optimal costs establish upper bounds?
Assume a bidirectional search scheme, where we aim to run two A* searches in directions $d$ and $d'$.
For the A* search in direction~$d'$, we can obtain ${\bf h}^{d'}$ and ${\bf ub}^{d'}$ essentially by running two one-to-all unidirectional searches in direction~$d$.
Now, we can see $d$ as the direction in which we run one-to-all heuristic searches and later one of the bidirectional A* searches.
Therefore, if A* generates a node $x$ in direction~$d$, we already know that the costs of concrete paths to $s(x)$ (i.e., ${\bf g(x)}$) is at least the costs of state $s(x)$'s shortest paths previously obtained via the preliminary heuristic searches in direction~$d'$, so we always have ${\bf h}^{d'}(s(x)) \preceq {\bf g(x)}$.
With this introduction, we now formally prove that nodes violating these upper bounds are dominated nodes.
\begin{lemma}
\label{lemma:state_ub}
The search node $x$ generated in direction~$d$ is a dominated node if ${\bf g}(x) \npreceq {\bf ub}^{d'}(s(x))$.
\end{lemma}
\begin{proof}
Let $x$ be a search node generated in the A* search of direction~$d$.
Let ${\bf h}^{d'}$ and ${\bf ub}^{d'}$ be the lower and upper bound functions already obtained by running two one-to-all searches in direction~$d$.
Given ${\bf h}^{d'}$ as a pair of admissible heuristic functions, under the premises of Lemma~\ref{lemma:admissible}, we always have ${\bf h}^{d'}(s(x)) \preceq {\bf g}(x)$.
There are two possible cases for the actual costs in ${\bf g}(x)$ if we compare them against the proposed upper bounds, namely ${\bf g}(x) \preceq {\bf ub}^{d'}(s(x))$ or ${\bf g(x)} \npreceq {\bf ub}^{d'}(s(x))$.
We are interested in the second relation ${\bf g(x)} \npreceq {\bf ub}^{d'}(s(x))$.
We decompose this case into two possible scenarios: either ${ub}^{d'}_1(s(x)) < g_1(x)$ or ${ub}^{d'}_2(s(x)) < g_2(x)$.
For the first scenario ${ub}^{d'}_1(s(x)) < g_1(x)$, we can see that node $x$ is dominated by the preliminary shortest path on $\mathit{cost_2}$ as we already know ${h}^{d'}_2(s(x)) \leq g_2(x)$.
Similarly, for the second scenario ${ub}^{d'}_2(s(x)) < g_2(x)$, node $x$ is dominated by the preliminary shortest path on $\mathit{cost_1}$ as we already have ${h}^{d'}_1(s(x)) \leq g_1(x)$.
Therefore, we can conclude that a node $x$ in search direction~$d$ is always dominated by one of the preliminary shortest paths obtained by one-to-all searches of the same direction if ${\bf g}(x) \npreceq {\bf ub}^{d'}(s(x))$. 
\end{proof}
Lemma~\ref{lemma:state_ub} shows that nodes with established costs strictly larger than the state's $\mathit{ub}$-values are already dominated, and thus the ${\bf ub}^{d'}$ function defines upper bounds on {\bf cost} of nodes generated in direction~$d$.
Hence, our A* searches can use this upper bound function to prune dominated nodes in the bidirectional setting (when these upper bounds exist in both directions).
In summary, we can see from Lemmas~\ref{lemma:admissible} and \ref{lemma:state_ub} that running one-to-all searches in a generic direction~$d$ not only establishes the heuristics of A* in direction~$d'$ but also helps bidirectional A* algorithms in detecting dominated nodes in the search direction~$d$. \par
\textbf{Global Upper Bounds:} Having a pair of global upper bounds is usually essential to appropriately navigate the constrained search in A* to an optimal solution.
Our algorithms use two global upper bounds for the search, namely $\overline{\bf f}=(\overline{f_1},\overline{f_2})$, to distinguish invalid (out-of-bounds) nodes.
The second element~$\overline{f_2}$, which is the global upper bound on $\mathit{cost_2}$ of complete solution paths, can be initialised with the weight limit $W$.
Clearly, any node $x$ with the estimated cost $f_2(x) > W$ is out-of-bounds and considered invalid for the problem.
However, the initial value of $\overline{f_1}$ is not known beforehand and should be obtained by the search.
As a potential candidate, once preliminary heuristic functions are computed, we can use the $\mathit{cost_2}$-optimum path between $\mathit{start}$ and $\mathit{goal}$ to initialise $\overline{f_1}$.
We briefly explain one such initialisation.
Assume an A* search in the forward direction is intended.
We run two one-to-all backward searches on $\mathit{cost_1}$ and $\mathit{cost_2}$ and obtain the lower and upper bound functions ${\bf h}^f$ and ${\bf ub}^f$.
In general, there may be three possible cases for the upcoming A* search after establishing the required functions:
\begin{enumerate}[nosep]
  \item $\mathit{ub}^f_2(\mathit{start})$ is within the global upper bound $\overline{f_2}$ (the weight limit). In this case, the algorithm can immediately terminate and return the shortest path on $\mathit{cost_1}$ as an optimal solution path.
  \item $h^f_2(\mathit{start})$ is strictly larger than the global upper bound $\overline{f_2}$. In this case, the algorithm can immediately terminate, as there would not be any feasible solution to the problem.
  \item $h^f_2(\mathit{start})$ is within the global upper bound $\overline{f_2}$. In this case, the shortest $\mathit{start}$-$\mathit{goal}$ path on $\mathit{cost_2}$ can be considered as an initial solution path. Thus, we can use $ub^f_1(\mathit{start})$ to initialise $\overline{f_1}$.
\end{enumerate}
Updating the global upper bound $\overline{f_1}$ via the condition of case three above is always safe because the shortest path on $\mathit{cost_2}$ would trivially become the first complete $\mathit{start}$-$\mathit{goal}$ path within the bound.
However, one can shrink the gap between the initial and optimal solution by updating the global upper bound $\overline{f_1}$ with every (lexicographically) shorter $\mathit{start}$-$\mathit{goal}$ path observed during the search, as in \citet{AhmadiTHK21}.
To this end, the pair ($\overline{f_1},f^{sol}_2$) keeps track of the actual costs of the best-known solution path during the search in A*.
More accurately, we update ($\overline{f_1},f^{sol}_2$) of the current solution path only if ($\mathit{cost_1},\mathit{cost_2}$) of the new solution path is lexicographically smaller than the current solution path.

So far, we have discussed two types of unpromising nodes: \textit{(weakly) dominated} and \textit{invalid} nodes.
In general, nodes can be (weakly) dominated by the state's last expanded node or the state's upper bound (Lemmas~\ref{lemma:last_sol}~and~\ref{lemma:state_ub}).
Further, nodes can be invalidated by global upper bounds, either via the weight limit or via the current solution path.
We now formally prove in Lemmas~\ref{lemma:invalid} and \ref{lemma:dominated} why invalid and dominated nodes never lead to a {\bf cost}-optimum solution path for the WCSPP. 
\begin{lemma}
\label{lemma:invalid}
Invalid nodes are not part of any {\bf cost}-optimal $\mathit{start}$-$\mathit{goal}$ solution path.
\end{lemma}
\begin{proof}
Let $\overline{f_1}$ be $\mathit{cost_1}$ of the best known valid $\mathit{start}$-$\mathit{goal}$ path.
Let $x$ be an invalid node generated in the search direction~$d$ with $f_1(x) > \overline{f_1}$ or $f_2(x) > \overline{f_2}$.
If $f_1(x) > \overline{f_1}$, A* guarantees that extending $x$ towards the target state would not lead to a solution path better than the current solution path with cost $\overline{f_1}$ (see Lemma~\ref{lemma:astar}).
Hence, $x$ cannot be part of the final {\bf cost}-optimal solution path.
If $f_2(x) > \overline{f_2}$, the admissible heuristic function $h^d_2$ guarantees that extending $x$ towards the target state would never lead to a solution path within the upper bound requirement of the WCSPP.
Therefore, we can conclude that invalid nodes are not part of any {\bf cost}-optimal solution path.
\end{proof}
\begin{lemma}
\label{lemma:dominated}
Dominated nodes are not part of any {\bf cost}-optimal $\mathit{start}$-$\mathit{goal}$ solution path.
\end{lemma}
\begin{proof}
We prove this lemma by assuming the contrary, namely by claiming that dominated nodes can be part of a {\bf cost}-optimal $\mathit{start}$-$\mathit{goal}$ solution paths.
Let $x$ and $y$ be two nodes associated with the same state where $y$ is dominated by $x$.
Suppose that path~$\pi^*$ is a {\bf cost}-optimal $\mathit{start}$-$\mathit{goal}$ solution path containing the dominated node $y$.
Since $x$ dominates $y$, we have either $g_1(x) < g_1(y)$ or $g_2(x) < g_2(y)$.
In either of the cases, one can replace node $y$ with $x$ on $\pi^*$ to further reduce the cost of $\mathit{start}$-$\mathit{goal}$ optimum path for at least one attribute.
However, being able to reduce the cost of the established optimal solution path would contradict our assumption on the optimality of the solution path $\pi^*$.
Therefore, we conclude that dominated nodes cannot be part of any {\bf cost}-optimal $\mathit{start}$-$\mathit{goal}$ solution path. 
\end{proof}
Lemmas~\ref{lemma:invalid}~and~\ref{lemma:dominated} imply that constrained A* search can ignore invalid or dominated nodes, given the fact that such nodes would never be part of a {\bf cost}-optimal solution path.
Therefore, we can use the aforementioned dominance and validity tests to prune unpromising nodes during the search.
However, we still need to show why A* can safely prune weakly dominated nodes.
Lemma~\ref{lemma:weakly_dominated} will help us to show the correctness of this type of pruning. 
\begin{lemma}
\label{lemma:weakly_dominated}
Let $y$ be a node weakly dominated by node $x$ and $s(x)=s(y)$.
If $y$ is part of a {\bf cost}-optimal solution path, $x$ is also part of a {\bf cost}-optimal solution path.
\end{lemma}
\begin{proof}
We prove this lemma by assuming the contrary, namely that $x$ is not part of any {\bf cost}-optimal solution path.
Since $y$ is weakly dominated by $x$, we have ${\bf g}(x) \preceq {\bf g}(y)$, or $g_1(x) \leq g_1(y)$ and $g_2(x) \leq g_2(y)$ equivalently.
We distinguish two cases:
(1) either $g_1(x) <g_1(y)$ or $g_2(x) < g_2(y)$; (2) both $g_1(x) = g_1(y)$ and $g_2(x) = g_2(y)$.\\
Case (1): node $x$ is lexicographically smaller than node $y$ and is not part of any optimal solution path, so one can replace the partial path of node $y$ with the partial path of node $x$ and nominate a path lexicographically smaller than the current solution path via node $y$. 
However, the existence of such smaller solution path would contradict our assumption on the optimality of the solution path via node $y$.\\
Case (2): nodes $x$ and $y$ both offer the same pair of costs. 
$x$ is not part of any {\bf cost}-optimum solution path, so there should exist an optimum path $\pi^*$ lexicographically smaller than any solution path via $x$.
However, since $y$ also has the same costs as $x$, and since they are associated with the same state, there would consequently not be any optimum path via $y$ lexicographically smaller than the optimum path $\pi^*$, otherwise, we could have made an optimal path via $x$. 
However, this contradicts our assumption on the optimality of the solution path via node $y$.\\
Therefore, node $x$ will definitely be part of a {\bf cost}-optimum solution path if we have established an optimal solution via node $y$.
\end{proof}
Lemma~\ref{lemma:weakly_dominated} shows that if one node is weakly dominated by another node, it is always safe to prune the weakly dominated node and keep the other node, because:
if the weakly dominated node does not lead to a {\bf cost}-optimum solution, its pruning has been correct;
otherwise, Lemma~\ref{lemma:weakly_dominated} guarantees that the unpruned node would be able to produce a {\bf cost}-optimum solution in the absence of the weakly dominated node.
Now looking back to our dominance test in Lemma~\ref{lemma:last_sol}, we can see that A* has already processed and stored one such candidate node (the last expanded node) and weakly (not-yet-expanded) dominated nodes can be safely pruned.

\textbf{All Optimum Paths:}
The fact that A* can safely prune weakly dominated nodes does not prevent us from being able to obtain all (equally) optimum solution paths.
Recall the proof in Lemma~\ref{lemma:weakly_dominated}.
We can see that if a weakly dominated node $y$ is part of a {\bf cost}-optimal solution path, the node that weakly dominates $y$ must also be part of an optimum path with costs equal to ${\bf g}(y)$.
In other words, both nodes are equally good in terms of optimal costs, and thus we can replace the non-dominated node with $y$ and obtain another optimum path.
Therefore, if we aim to obtain all {\bf cost}-optimal solution paths, we just need to capture non-dominated nodes with equal costs, knowing that expanding one of them is enough to establish at least one optimum path.
For example, if $x$ is our last expanded node for state $\mathit{u}$, and $y$ is the next node generated for $\mathit{u}$ with ${\bf g}(x)={\bf g}(y)$, we do not expand $y$ but link it to $x$ as an alternative (equal) path.
If $x$ appears on an optimal path, we can simply replace $x$ with $y$ to generate a different (yet optimal) path.
In such cases, all possible permutations of equal nodes on the optimal path would yield all {\bf cost}-optimum paths.

We now describe one of the other benefits of pruning weakly dominated nodes,
which is enabling A* to produce cycle-free paths.
Lemma~\ref{lemma:cycle} formally states this observation.

\begin{lemma}
\label{lemma:cycle}
Suppose A* prunes weakly dominated nodes.
Let $\pi=\{u_1,u_2,u_3,\dots,u_n\}$ be the sequence of states on any arbitrary path $\pi$ extended in A*.
We have $u_i \neq u_j$ for any $i,j \in 1\dots n, i \neq j$.
\end{lemma}
\begin{proof}
We need to show that there is no cycle in any of the paths extended by A* via node expansion.
Assume A* has expanded node $x$ with state $\mathit{u}$.
Let $z$ be any arbitrary node whose ancestor is $x$.
Node $z$'s state is also $\mathit{u}$, so we will definitely have a cycle if A* expands $z$.
This is because we would visit the state $\mathit{u}$ twice if we backtrack from $z$, once at $z$ and once again at $x$.
Now consider the dominance test in Lemma~\ref{lemma:last_sol}.
Since $x$ is the ancestor of $z$, we know that $x$ has already been expanded.
In addition, since the edge costs are non-negative, node $z$ cannot improve its ancestor's costs upon revisiting state $\mathit{u}$, 
thus we have $f_p(x) \leq f_p(z)$ and $f_s(x) \leq f_s(z)$, or $f_p(x) \preceq f_p(z)$ equivalently, which means $z$ is weakly dominated by $x$.
Therefore, $z$ will be pruned as a weakly dominated node and A* with dominance pruning will never extend a path with repeated states.
\end{proof}
Given the important observation in Lemma~\ref{lemma:cycle}, we now formally prove that our constrained A* search terminates.
In addition, we present a condition under which our A* algorithms can terminate early, without needing to explore (not-yet-expanded) nodes in $\mathit{Open}^d$.
\begin{lemma}
\label{lemma:astar_terminate}
Suppose nodes in $\mathit{Open}^d$ are ordered based on $f_p$-values.
Also, suppose A* prunes weakly dominated nodes.
Let $x$ be the least cost node extracted from $\mathit{Open}^d$.
A* can terminate early if $f_p(x) > \overline{f_p}$.
\end{lemma}
\begin{proof}
Under the premises of Lemma~\ref{lemma:cycle}, we know that A* does not generate (even zero-weight) cycles if weakly dominated nodes are pruned.
In addition, A* does not expand the target state.
Hence, without cycles, the number of $\mathit{start}$-$\mathit{goal}$ paths becomes finite.
Given the finite number of paths, we can guarantee that A* eventually terminates after enumerating all paths, or when $\mathit{Open}^d$ is empty.
Now consider the early termination criterion, i.e., $f_p(x) > \overline{f_p}$.
The termination criterion denotes that $x$ is an invalid node.
When A* extracts the least cost node $x$ from $\mathit{Open}^d$, it guarantees that all future expanded nodes will have a primary cost of $f_p(x)$ or more (see Corollary~\ref{corollary:astar}).
Thus, we can see that all nodes that could get expanded after $x$ (generated or not) are
invalid, and according to Lemma~\ref{lemma:invalid}, their expansion will not lead to a {\bf cost}-optimal solution path.
Therefore, A* can terminate early as soon as it extracts an invalid node from $\mathit{Open}^d$.
\end{proof}
Given the search termination criterion elaborated above, we finally describe two situations in which our constrained search can establish a tentative solution path.
\begin{lemma}
\label{lemma:early_sol}
Suppose nodes in $\mathit{Open}^d$ are ordered based on $f_p$-values.
Let $x$ be a valid search node extracted from $\mathit{Open}^d$ in direction~$d$.
If $g_s(x) + \mathit{ub}^{d}_s(s(x)) \leq \overline{f_s}$, the search has found a tentative solution path with the primary cost of $f_p(x)$.
\end{lemma}
\begin{proof}
Let $\pi$ be the complete $\mathit{start}$-$\mathit{goal}$ path that resulted from joining node $x$ with its complementary shortest path on primary attribute $\mathit{cost_p}$, so we have $\mathit{cost_p}(\pi)=f_p(x)$ and $\mathit{cost_s}(\pi)=g_s(x) + \mathit{ub}^{d}_s(s(x))$.
Since all complementary paths are already optimum, A* guarantees that extending $x$ in ($f_p,f_s$) order towards the target state via states on the shortest path for $\mathit{cost_p}$ will lead to a complete $\mathit{start}$-$\mathit{goal}$ path with the same primary cost $f_p(x)$.
Therefore, given $x$ as a valid node, path $\pi$ is tentatively a solution path if $\mathit{cost_s}(\pi) \leq \overline{f_s}$.
\end{proof}
\begin{lemma}
\label{lemma:terminal_node}
Suppose nodes in $\mathit{Open}^d$ are ordered based on $f_p$-values, and let $x$ be the least cost node extracted from $\mathit{Open}^d$.
Suppose $x$ is a valid terminal node. 
$x$ is a tentative solution node.
\end{lemma}
\begin{proof}
Let $\mathit{cost_s}$ be the tie-breaker in the preliminary heuristic searches on $\mathit{cost_p}$ (cases where we break the tie for paths with equal $\mathit{cost_p}$).
State $s(x)$ with $h^d_p(s(x))= \mathit{ub}^d_p(s(x))$ offers one complementary path which is optimum for both attributes $\mathit{cost_1}$ and $\mathit{cost_2}$.
Thus, we must have $h^d_s(s(x))= \mathit{ub}^d_s(s(x))$.
Otherwise, if $h^d_s(s(x)) < \mathit{ub}^d_s(s(x))$, the shortest path on $\mathit{cost_s}$ dominates the shortest path on $\mathit{cost_p}$ but such cases are avoided by the tie-breaker.
Therefore, by joining $x$ with its complementary shortest path we have $f_p(x)=g_p(x)+h^d_p(s(x))$ and $f'_s=g_s(x)+\mathit{ub}^d_s(s(x))=f_s(x)$.
Since $x$ is a valid node and has not been pruned by the global upper bounds,
we can conclude that ${f}_p(x) \leq \overline{{ f}_p}$ and ${f}_s(x) \leq \overline{{ f}_s}$, which yields $f'_s \leq \overline{f_s}$.
Therefore, $x$ is a tentative solution node as the joined path is valid.
\end{proof}

\section{Constrained Pathfinding with WC-EBBA*}
\label{sec:WC_EBBA}
This section revisits the key components of weight-constrained search with {WC-EBBA*} (Enhanced Biased Bidirectional A*) presented in \citet{AhmadiTHK21}, with its high-level description described in Algorithm~\ref{alg:rcebba_high}.
The algorithm runs in two phases: initialisation and constrained search.
In the initialisation phase, {WC-EBBA*} establishes its bidirectional heuristic functions via Algorithm~\ref{alg:init_ebba}.
Next, it uses forward and backward lower bounds accessible via the lower bound function {\bf h} to determine a pair of budget factors for the main search of the next phase.
{WC-EBBA*} then uses the budget factors to conduct its biased A* search via Algorithm~\ref{alg:rc_ebba}.
When {WC-EBBA*}'s biased search terminates, $\mathit{Sol}$ contains a pair of cost-optimal solutions nodes.
Clearly, there would not be any feasible solution to the problem if $\mathit{Sol}$ does not contain a search node, or if $\mathit{Sol}=(\varnothing,\varnothing)$ equivalently.
We explain the main procedures in each phase as follows.
%


\begin{algorithm}[!t]
\small
\caption{WC-EBBA* Higher level}
\label{alg:rcebba_high}
  \DontPrintSemicolon 
    \SetKwBlock{DoParallel}{do in parallel}{end}
    \KwIn{A problem instance (G, {\bf cost}, ${start}$, ${goal}$) with the weight limit $W$}
    \KwOut{A node pair corresponding with the cost-optimal feasible solution $Sol$}
     ${\bf h}, {\bf ub}, {\bf \overline{f}}, S' \leftarrow$   Initialise WC-EBBA* (G, {\bf cost}, ${start}$, ${goal}, W$) \Comment*[r]{Algorithm~\ref{alg:init_ebba}}
  $\beta^f , \beta^b \leftarrow$ Obtain forward and backward budget factors using $h_1$-values of non-dominated states in $S'$. \Comment*[r]{Eq.(\ref{bias_factor})}
     $Sol \leftarrow$ Run a biased WC-EBBA* search for (G, {\bf cost}, ${start}$, ${goal}$) with global upper bounds ${\bf \overline{f}}$, \;
    \nonl \qquad \qquad heuristic functions $({\bf h}, {\bf ub})$, and budget factors ($\beta^f,\beta^b$). \Comment*[r]{Algorithm~\ref{alg:rc_ebba}}
\Return{$Sol$}
\end{algorithm}

\subsection{Initialisation} 
This phase is responsible for computing the lower and upper bound functions required by the forward and backward (constrained) A* searches of phase two, i.e, (${\bf h}^f, {\bf h}^b$) and (${\bf ub}^f, {\bf ub}^b$).
To compute the heuristic functions (${\bf h}^f, {\bf h}^b$), RC-BDA* \cite{thomas2019exact} conventionally runs two one-to-all searches in each direction, using Dijkstra's algorithm for example.
This means that RC-BDA*'s runtime is at least four times a single one-to-all Dijkstra's search, even for simple problems with very small search space.
{WC-EBBA*} improves this serious inefficiency by adapting a classic technique called \textit{resource-based network reduction}.
Originally presented by \citet{AnejaAN83} for solving the RCSPP, the reduction technique tries to reduce the size of the network by removing vertices violating the problem limits, normally via an iterative lower-bounding procedure.
However, the use case of the \textit{network reduction} technique in {WC-EBBA*} is much simpler than other iterative methods.
{WC-EBBA*} replaces each one-to-all search of RC-BDA* with a cost-bounded (single-objective) A* search, reducing the graph size via four consecutive heuristic searches.
Algorithm~\ref{alg:init_ebba} shows all steps in detail.

\begin{algorithm}[!t]
\small
\caption{Initialisation phase of WC-EBBA*}
\label{alg:init_ebba}
\DontPrintSemicolon
    \SetKwBlock{DoParallel}{do in parallel}{end}
    \KwIn{The problem instance (G, {\bf cost}, $\mathit{start}$, $\mathit{goal}$) and the weight limit $W$}
    \KwOut{WCSPP's bidirectional heuristic functions ${\bf h}$ and ${\bf ub}$, global upper bounds ${\bf \overline{f}}$, also non-dominated states $S'$}
    
    Set global upper bounds: $\overline{f_1} \leftarrow \infty$ and 
    $\overline{f_2} \leftarrow W $ \;
    
      $ h^f_2, ub^f_1 \leftarrow$ Run $f_2$-bounded backward A* (from $\mathit{goal}$ using an admissible heuristic) on $cost_2$, use $cost_1$ as a tie-breaker,\;
      \nonl \qquad \qquad update $\overline{f_1}$ with $ub^f_1(\mathit{start})$ when $\mathit{start}$ is going to get expanded
      and stop before expanding a state with $f_2 > \overline{f_2}$.
      \label{alg:ebba_init:init_1} \;
      
      $h^b_2, ub^b_1 \leftarrow$ Run $f_2$-bounded forward A* (from $\mathit{start}$ using $h^f_2$ as an admissible heuristic) on $cost_2$, use $cost_1$ as a tie-breaker,\;
      \nonl \qquad \qquad  update $\overline{f_1}$ via partial paths matching if feasible and stop before expanding a state with $f_2 > \overline{f_2}$. \label{alg:ebba_init:init_2} \;
      
      $h^b_1, ub^b_2 \leftarrow$ Run $f_1$-bounded forward A* (using an admissible heuristic) on $cost_1$, use $cost_2$ as a tie-breaker, ignore unexplored\; 
      \nonl \qquad \qquad states in the search of line~\ref{alg:ebba_init:init_2},
      update $\overline{f_1}$ via paths matching if feasible and stop before expanding a state with $f_1 > \overline{f_1}$. \label{alg:ebba_init:init_3}\;
      
      $ h^f_1, ub^f_2 \leftarrow$ Run $f_1$-bounded backward A* (using $h^b_1$ as an admissible heuristic) on $cost_1$, use $cost_2$ as a tie-breaker, only use states\;
      \nonl \qquad \qquad explored in the search of line \ref{alg:ebba_init:init_3}, update $\overline{f_1}$ via paths matching if feasible and stop before expanding a state with $f_1 > \overline{f_1}$. \label{alg:ebba_init:init_4} \;
$S' \leftarrow$ Non-dominated states explored in the last bounded A* search of line \ref{alg:ebba_init:init_4}  \;  
\Return{$({\bf h}^f, {\bf h}^b)$, $({\bf ub}^f, {\bf ub}^b)$, ${\bf \overline{f}}$, $S'$}
\end{algorithm}
The initialisation phase always starts with setting global upper bounds.
Lines \ref{alg:ebba_init:init_1} and \ref{alg:ebba_init:init_3} in Algorithm~\ref{alg:init_ebba} require an admissible heuristic function to establish their $f$-values.
However, as with other A* searches, each search in those steps would turn into a cost-bounded Dijkstra's algorithm (A* with zero heuristic) if such heuristic does not exist.
Another interesting observation in this initialisation procedure is that the network reduction technique improves the quality of the heuristic functions. 
For example, heuristics obtained in the fourth search are much more informed than those of the first search.
One could iterate this four-step procedure to further inform the (constrained) search heuristics, but that indeed increases the time {WC-EBBA*} would spend in the initialisation phase, which is not ideal for easy problems.
Note that the order of preliminary searches can change depending on the quality of heuristics and even tightness of the constraint.
For example, if our heuristic function for $\mathit{cost_1}$ (used in line~\ref{alg:ebba_init:init_3}) is more informed than that of $\mathit{cost_2}$ (used in line~\ref{alg:ebba_init:init_1}), we can start our preliminary bounded searches with $\mathit{cost_1}$ as it can potentially invalidate more states for the other searches on $\mathit{cost_2}$ (see Section~\ref{sec:initilaisation_importance} for empirical analysis).

There is another optimisation that {WC-EBBA*} employs in its initialisation phase.
After the first bounded search, {WC-EBBA*} tries to reduce the global upper bound $\overline{f_1}$ by applying a form of path matching strategy to all partial paths explored (similar to the ESU strategy).
Every time the search computes a lower bound in lines \ref{alg:ebba_init:init_2}-\ref{alg:ebba_init:init_4}, at the same time, it joins the corresponding partial path with a complementary (optimum) path already obtained in the opposite direction.
If the resulting $\mathit{start}$-$\mathit{goal}$ path is valid, the initial search can update $\overline{f_1}$ with $\mathit{cost_1}$ of the joined path.
This technique helps {WC-EBBA*} to further reduce the graph size in steps 3-4, obtain better quality heuristics and start the constrained search with an initial solution better than $\mathit{ub}^f_1(\mathit{start})$.
We showed in \citet{AhmadiTHK21} how this technique can effectively shrink the gap between the initial and optimal solutions, and sometimes solve the problem in the initialisation phase by fully closing the gap.
For the final reduced graph (after line \ref{alg:ebba_init:init_4}), we obtain $S' \subseteq S$ as a set of valid states.
\subsection{Biased Search}
We now discuss the constrained search of {WC-EBBA*} in a different notation than the original method in \citet{AhmadiTHK21}.
We also provide a complete proof of algorithm correctness based on the procedures presented in the previous parts.
The initialisation phase in {WC-EBBA*} can help the main (constrained) search in biasing the search effort.
In contrast to RC-BDA* where each search direction is allocated 50\% of the \textit{weight} budget, the search in {WC-EBBA*} generally works with a budget factor $\beta^d \in [0,1]$ for each search in direction~$d$.
$\beta^f$ is considered as the budget factor of the forward search, while $\beta^b$ is the budget factor of the backward search.
$\beta^f$ and $\beta^b$ are chosen in a way that they both always cover the entire range of the budget with $\beta^f+\beta^b=1$.
We can see that setting $\beta^f=\beta^b=0.5$ models the budget allocation of {RC-BDA*}.
However, {WC-EBBA*} allocates more budget to a direction that will use smaller primary lower bounds on average, i.e., direction~$d$ with $ \sum_{u \in S'} h^d_1(u) < \sum_{u \in S'} h^{d'}_1(u)$.
Note that $d'$ is the opposite direction of $d$ and $S' \subseteq S$ is the set of valid states explored in the last step of the initialisation phase.
{WC-EBBA*} employs Eq.~(\ref{bias_factor}) to determine the budget factors in directions $d$ and $d'$ based on the condition above.
\begin{equation} 
\label{bias_factor}
\beta^d= \min\left(1, 0.5 \times \dfrac{\sum_{u \in S'} h^{d'}_1(u)} {\sum_{u \in S'} h^{d}_1(u)}\right)
\qquad 
\text{and}
\qquad 
\beta^{d'}=1-\beta^d
\end{equation}
After calculating the budget factors $\beta^f$ and $\beta^b$, the main (constrained) A* search starts. 
Algorithm~\ref{alg:rc_ebba} shows the pseudocode of the biased search in {WC-EBBA*}.
The algorithm first initialises necessary data structures in both directions by calling the Setup($d$) procedure described in Algorithm~\ref{alg:rc_ebba_setup}.
This includes preparing the node set $\chi^d(s(x))$ for every $u \in S$ and inserting a node associated with the initial state into each priority queue.
After this procedure, we have one node in the priority queue of each direction, i.e., $\mathit{Open}^f$ and $\mathit{Open}^b$.
The algorithm then initialises a solution node pair $\mathit{Sol}$ and a scalar $f^{sol}_2$ to keep track of the $\mathit{cost_2}$ of the last solution path.
\begin{figure}[!t]
\noindent
\begin{minipage}[t]{.49\textwidth}
\null 
\input{Algorithms/RC_EBBA_Main}
\end{minipage}%
\hspace{2mm}
\begin{minipage}[t]{.49\textwidth}
\null
\input{Algorithms/join}
\input{Algorithms/store}
\end{minipage}
\end{figure}

\textbf{Node Selection and Stopping Criterion:} The algorithm enumerates all nodes iteratively until there is no node in the priority queues to explore, or the stopping criterion is met.
{WC-EBBA*} can perform only one iteration at a time.
Therefore, to guarantee optimally, it needs to start each iteration with a direction that offers a smaller cost estimate in the ($f_1,f_2$) order.
Let $x$ be a node with the lexicographically smallest $(f_1,f_2)$ among all nodes in $\mathit{Open}^f$ and $\mathit{Open}^d$.
If this node belongs to $\mathit{Open}^f$, {WC-EBBA*} conducts a forward search, or a backward search if $x \in \mathit{Open}^b$.
Note that A* only needs $x$ to be the least-cost node, however, extracting the lexicographically smallest node from $\mathit{Open}^f \cup \mathit{Open}^b$ will prevent the search expanding dominated nodes.
Let $d$ be the direction the search wants to extend $x$.
For this direction, the search removes $x$ from $\mathit{Open}^d$ to avoid re-expansions of $x$.
In the next step, the algorithm checks if the stopping criterion is met, i.e., whether the node's estimated cost is violating the primal upper bound $\overline{f_1}$ (line~\ref{alg:rc_ebba:termination}).
If this is the case, we immediately stop the search and return the node pair $\mathit{Sol}$ as the optimal solution nodes.
This termination criterion is correct because of two reasons: (1) we have chosen the least cost node in $\mathit{Open}^f \cup \mathit{Open}^d$, so all other nodes in both priority queues would also violate the global upper bound if we wanted to expand them (see Lemma~\ref{lemma:astar_terminate}), (2) expansion of the current node $x$ does not generate a valid node as we always use consistent and admissible heuristic functions {\bf h} (see Lemma~\ref{lemma:astar}).
Therefore, we can guarantee that ($\overline{f_1},f^{sol}_2$) are the optimal costs of the problem when the algorithm surpasses $\overline{f_1}$ and terminates with non-empty solution nodes in $\mathit{Sol}$.
\par
Note that {WC-EBBA*}'s criterion here is different from that of RC-BDA*.
In RC-BDA*, the algorithm generates a new node for every tentative solution and inserts them into the $\mathit{Open}$ lists.
Consequently, the search in RC-BDA* can terminate as soon as it extracts a solution node from the priority queue.
This approach causes unnecessary overhead, mainly because the priority queues would also need to order solution nodes. 
Our stopping criterion, however, directly works with upper bounds and captures solution nodes on the fly.
Therefore, {WC-EBBA*} does not need to handle solution nodes in the priority queue.

\textbf{Pruning with Lazy Dominance Check:}
If $x$ is within the global bound $\overline{f_1}$, the algorithm performs a constant time but \textit{lazy} dominance check via line~\ref{alg:rc_ebba:prune2}.
We already discussed the correctness of this pruning in Lemma~\ref{lemma:weakly_dominated}, but we briefly explain pruning with lazy dominance test in the ($f_1,f_2$) order for the sake of clarity.
Let $\mathit{x}$ and $\mathit{y}$ be two nodes associated with the same state, both generated in direction~$d$ and available in $\mathit{Open}^d$.
Since {WC-EBBA*} checks nodes for dominance lazily, the priority queue may contain dominated nodes.
We conduct the search in the ($f_1,f_2$) order and expand $y$ first, i.e., we have $f_1(y) \leq f_1(x)$.
Now, we want to expand node~$x$.
Since both nodes $\mathit{x}$ and $\mathit{y}$ have used the same heuristic function to establish their ${\bf f}$-values, we already have $h^d_1(s(y))=h^d_1(s(x))$ and consequently $g_1(y) \leq g_1(x)$.
Therefore, we can see that node $x$ is weakly dominated by node $y$ if $g_2(y) \leq g_2(x)$.
Based on this observation, as nodes of each state are explored in the non-decreasing order of their $\mathit{cost_1}$ (see Lemma~\ref{lemma:astar}), before expanding every new node $x$ with state $s(x)$, we just need to compare the secondary cost of $x$ with that of the last expanded node for $s(x)$.

All of our A* algorithms take advantage of this constant time dominance test by systematically keeping track of the secondary cost of the last successfully expanded node in $g^d_{\mathit{min}}(s(x))$.
For the {WC-EBBA*} algorithm, this operation can be seen in line~\ref{alg:rc_ebba:min_r}.
At this point, we can see $x$ as a non-dominated node.

\textbf{ESU and Terminal Nodes:} In the next step (line~\ref{alg:rc_ebba:early_sol}), the algorithm uses node $x$ to search for a tentative solution and to possibly improve the global upper bounds via the ESU($x,d$) procedure or Procedure~\ref{alg:early_sol}.
However, the algorithm can still skip expanding $x$ if it is a terminal node (see Lemma~\ref{lemma:terminal_node}).
In such cases, we have $h^d_1(s(x))= \mathit{ub}^d_1(s(x))$ as shown at line~\ref{alg:rc_ebba:unique_path} of Algorithm~\ref{alg:rc_ebba}.
However, such nodes are already captured by the ESU strategy, so the search can safely prune partial paths arriving at states with a unique complementary path given the fact that the tentative solution is already stored in $\mathit{Sol}$ and expanding such nodes is not necessary.\par
\textbf{Expand and Prune:} 
At line~\ref{alg:rc_ebba:expand}, if $g_2(x)$ is within the weight budget of direction~$d$, i.e., we have $\beta^d \times \overline{f_2}$, {WC-EBBA*} expands node $x$ via procedure ExP($x,d$) described in Procedure~\ref{alg:expansion} (note that $\overline{f_2}$ has already been initialised with the weight limit $W$).
The ExP procedure in {WC-EBBA*} undertakes all the proposed pruning strategies, i.e., pruning by dominance, by states' upper bounds and by global upper bounds.
For pruning by dominance, we again perform a lazy dominance check, that is, for the new node $y$, we check $g_2(y)$ against the $g_2$-value of the last expanded node (for state $s(y)$) stored in $g^d_{\mathit{min}}(s(y))$.
For pruning by bound strategies, we simply check ${\bf g}(y)$ against ${\bf ub}^{d'}(s(y))$ and ${\bf f}(y)$ against ${\bf \overline{f}}$.

\textbf{The Coupling Area:}
In bidirectional search, every node $x$ associated with the state $s(x)$ in one direction can be coupled with any of the partial paths successfully expanded for the same state in the opposite direction to construct a complete $\textit{start}$-$\textit{goal}$ path.
We say a node $x$ is in the coupling area if there already exists at least one candidate path in the opposite direction, or if 
the existence of such a path is expected according to lower bounds.
The former can be simply checked by measuring the size of the node list in the opposite direction, i.e., if $|\chi^{d'}(s(x))| > 0$, and the latter can be tested by measuring the minimum distance of $s(x)$ to the target state on $\mathit{cost_2}$, i.e., $h^{d}_2(s(x))$.
Note that $h^{d}_2(s(x))$ is at the same time a lower bound on $\mathit{cost_2}$ of all paths reaching $s(x)$ in the opposite direction~$d'$.
Therefore, if $h^{d}_2(s(x))$ is not within the weight budget of direction~$d'$, i.e., if $h^{d}_2(s(x)) > \beta^{d'} \times \overline{f_2}$, it then becomes clear that {WC-EBBA*} will never expand any node with state $s(x)$ in direction~$d'$.
More accurately, {WC-EBBA*} predicts that all such nodes with a $g_2$-value larger than the weight budget in direction~$d'$ will not get expanded, so
we can save time and space by not storing (and matching) nodes outside the coupling area.

\textbf{Partial Path Matching:} 
If $x$ has reached the coupling area, the search tries to match $x$ with candidates of the opposite direction~$d'$ via the Match($x,d'$) procedure in line~\ref{alg:rc_ebba:join}.
Procedure~\ref{alg:path_match} shows the main steps involved in matching $x$ with expanded paths in the opposite direction~$d'$.
For node $x$ with state $s(x)$, $\chi^{d'}(s(x))$ contains all the candidate nodes expanded for $s(x)$ in the opposite direction~$d'$.
The procedure is supposed to match $x$ with all candidate nodes in $\chi^{d'}(s(x))$ in order.
Let $y$ be one such node in the opposite direction from the list.
The procedure first calculates the total costs of the complete $\textit{start}$-$\textit{goal}$ path via $f'_1 = g_1(x)+g_1(y)$ and $f'_2 = g_2(x)+g_2(y)$.
It then performs a simple check to avoid matching all items in the list.
If $\mathit{cost_1}$ of the complete path ($f'_1$-value) is out-of-bounds, the procedure terminates the loop early because the next candidate nodes in the list will also be out-of-bounds.
Otherwise, if $\mathit{cost_2}$ of the complete path ($f'_2$-value) is within the global upper bound, the procedure takes both $x$ and $y$ as a pair of solution nodes and updates the solution costs $\overline{f_1}$ and $f^{sol}_2$ if the cost pair of the new joined path ($f'_1,f'_2$) is lexicographically smaller than that of the last solution path in the ($\mathit{cost_1},\mathit{cost_2}$) order.
We will discuss the correctness of these techniques in Lemma~\ref{lemma:path_matching}.\par

\textbf{Storing Partial Paths:}
Given $x$ as a node in the coupling area, the Store($x,d$) procedure in line~\ref{alg:rc_ebba:store} captures $x$ and adds it at the end of 
$\chi^{d}(s(x))$, the list of expanded nodes of state $s(x)$ in direction~$d$.
This procedure is essential for the fulfilment of partial path matching, however, the search would not be able to backtrack and construct the optimal solution if one wants to permanently delete nodes outside the coupling area.
We explain our memory efficient method for solution path construction later as part of our practical considerations in Section~\ref{sec:practicall}.
The Store($x,d$) procedure is a simple \textit{insert} operation, as shown in Procedure~\ref{alg:path_store}, but we later introduce a simple refinement to this procedure to adapt it for cases when there is no tie-breaking in place in the priority queues.
\begin{lemma}
\label{lemma:path_matching}
Consider the Match($x,d'$) procedure in {WC-EBBA*}.
The matching loop can terminate early if $f'_1 > \overline{f_1}$.
\end{lemma}
\begin{proof}
Under the premises of Lemma~\ref{lemma:astar}, we know that A* in {WC-EBBA*} explores nodes in the non-decreasing order of their $f_1$-values.
Let $\mathit{u}$ be a typical state for which we have expanded nodes in direction~$d'$ of the search.
Since all nodes with state $\mathit{u}$ have used the same heuristic function $h^{d'}_1(u)$ to determine their $f_1$-values, we consequently have a non-decreasing order for the $g_1$-value of nodes expanded with state $\mathit{u}$ in direction~$d'$.
The procedure Store($x,d$) simply adds nodes at the end of the node list $\chi^{d'}(u)$ if they are in the coupling area.
Therefore, we can see that the order of nodes in $\chi^{d'}(u)$ is analogously non-decreasing.
Let $x$ be a node with state $\mathit{u}$ in the coupling area and in direction~$d$.
The procedure Match($x,d'$) tries to join $x$ with all candidates in $\chi^{d'}(u)$ in order.
The first candidate node would then be the top node in the list $\chi^{d'}(u)$ with the smallest $g_1$-value.
Let $y$ be any arbitrary candidate node in the list.
The procedure calculates $\mathit{cost_1}$ and $\mathit{cost_2}$ of the joined path via $f'_1=g_1(x)+g_1(y)$ and $f'_2=g_2(x)+g_2(y)$, respectively. \\
Case (1): $f'_1 \leq \overline{f_1}$ and $f'_2 \leq \overline{f_2}$; we can see that the joined path is a tentative solution. In this case, the procedure is able to capture $(x,y)$ as a new solution pair.\\
Case (2): $f'_1 > \overline{f_1}$; we know that all nodes after $y$ in the node list $\chi^{d'}(u)$ already have a $g_1$-value no smaller than $g_1(y)$ and matching remaining nodes with $x$ would not yield a path with a smaller primary cost than $f'_1$.
Therefore, it is not possible to obtain a valid joined path with nodes that appear after $y$ in the node list, and the procedure can terminate the loop early.
\end{proof}
We now prove the correctness of the weight constrained search in {WC-EBBA*} with our proposed procedures.
\begin{theorem}
\label{theorem:wc_ebba}
{WC-EBBA*} returns a node pair corresponding with a {\bf cost}-optimal solution path for the WCSPP.
\end{theorem}
\begin{proof}
{WC-EBBA*} enumerates all valid paths from both directions.
We proved in Lemma~\ref{lemma:astar_terminate} that {WC-EBBA*}'s search terminates, even with zero-weight cycles.
In Lemmas~\ref{lemma:invalid}~and~\ref{lemma:dominated}, we showed that the proposed pruning strategies are correct.
We showed in Lemma~\ref{lemma:path_matching} that the Match strategy skips candidate nodes correctly.
In Lemma~\ref{lemma:early_sol}, we proved that the ESU strategy correctly tracks tentative solutions.
We proved in Lemma~\ref{lemma:terminal_node} that terminal nodes are always treated as a tentative solution and thus their expansion is not necessary.
Therefore, we just need to show that the search is still able to obtain a {\bf cost}-optimal solution path via the defined budget factors and also the Match($x,d'$) and Store($x,d$) procedures.

We know that there is no limit on the $g_1$-value of nodes during node expansion, but {WC-EBBA*} does not expand nodes if their $g_2$-value is not within the secondary upper bound determined by the budget factors.
Let $x$ with state $\mathit{u}$ be part of an optimal solution path, not expanded in direction~$d$ by violating the budget limit $g_2(x) > \beta^{d} \times \overline{f_2}$.
Without loss of generality, we prove that if the search in direction~$d$ does not expand the solution node $x$, the search in direction~$d'$ is able to join $x$ with a complementary solution node $y$ (associated with state $\mathit{u}$) and return a solution pair ($x,y$).
We first show that the valid node $x$ will be matched with the candidate nodes in $\chi^{d'}(u)$ and then stored in $\chi^d(u)$.
Let $\beta^d$ and $\beta^{d'}$ be the budget factors such that $\beta^d+\beta^{d'}=1$.
Node $x$ has not been expanded because $g_2(x) > \beta^{d} \times \overline{f_2}$.
In this case, we have $g_2(x) > (1-\beta^{d'}) \times \overline{f_2}$, or equivalently $\overline{f_2} - g_2(x) < \beta^{d'} \times \overline{f_2}$.
Since $x$ has been a valid node, we have $f_2(x) \leq \overline{f_2}$ which yields $f_2(x) - g_2(x) \leq \beta^{d'} \times \overline{f_2}$ and consequently $h^d_2(u) \leq \beta^{d'} \times \overline{f_2}$.
Therefore, $x$ will be processed by both the Match($x,d'$) and Store($x,d$) procedures.
In this case, if the complementary solution node $y$ has already been stored in $\chi^{d'}(u)$, then Match($x,d'$) detects it and the optimal solution pair will be established.
Assume that the solution node $y$ (as a counterpart) has not been discovered in direction~$d'$ yet.
We now show that $y$ is guaranteed to get expanded in direction~$d'$ if it belongs to an optimal solution path.
$(x,y)$ is our solution node pair with $s(x)=s(y)$, so we must have $g_2(x)+g_2(y) \leq \overline{f_2}$, otherwise the solution path will be invalid.
This relation is equivalent to $g_2(y) \leq \overline{f_2}-g_2(x)$.
We also know that $x$ was not expanded because $g_2(x) > \beta^{d} \times \overline{f_2}$.
Hence, we will have $\overline{f_2} - g_2(x) < \beta^{d'} \times \overline{f_2}$ and consequently $g_2(y) \leq \beta^{d'} \times \overline{f_2}$.
We can now see that node $y$ is within the budget limit of direction~$d'$ and will be expanded.
In addition, since the algorithm has already stored $x$ in the node list of direction~$d$, i.e., in $\chi^{d}(u)$, this list is no longer empty and node $y$ will go through the partial path matching procedure via line~\ref{alg:rc_ebba:match} of Algorithm~\ref{alg:rc_ebba}.
Therefore, the solution node pair ($x,y$) will be matched, and the optimal solution path is always discoverable.
\end{proof}
Note that the existence of a counterpart is necessary in the proof of Theorem~\ref{theorem:wc_ebba} because of the bidirectional nature of the algorithm.
A trivial case for this necessary component of the search is setting $\beta^d=0$.
In this case, the search in direction~$d$ only expands one node associated with the initial state and inserts it into $\chi^d(u_i)$.
Therefore, the search in direction~$d'$ can use the entire budget via $\beta^{d'}=1$ and match all nodes arriving at $u_i$ with the sole candidate node in $\chi^d(u_i)$.
\subsection{{WC-EBBA*} without Tie-breaking}
Breaking the ties between large numbers of nodes in the priority queue of A* may slow down the search.
We study a variant of {WC-EBBA*} that computes an optimal solution of the problem without performing tie-breaking in $\mathit{Open}$ lists.
More accurately, in every iteration of {WC-EBBA*}, we set $x$ to be a node with the minimum $f_1$-value (no condition on $f_2$) among all other nodes in $\mathit{Open}^f \cup \mathit{Open}^b$.
In the case of {WC-EBBA*}\textsubscript{par}, we extract from $\mathit{Open}^d$ a node with minimum $f_1$-value.
Note that both variants (sequential and parallel) originally pick a node with the lexicographically smallest ($f_1,f_2$) among all nodes in the priority queue.
Disabling tie-breaking in {WC-EBBA*} and {WC-EBBA*}\textsubscript{par} does not affect the correctness of the pruning strategies and the termination criterion, essentially because none of them requires nodes in the priority queue to be lexicographically ordered (see Lemmas~\ref{lemma:last_sol},~\ref{lemma:state_ub}~and~\ref{lemma:astar_terminate}).
In addition, the correctness of constrained search with budget factors in {WC-EBBA*} and {WC-EBBA*}\textsubscript{par} does not depend on expanding nodes in a lexicographical order (see Theorems~\ref{theorem:wc_ebba}~and~\ref{theorem:wc_ebba_par}).
However, disabling tie-breaking may change the order of nodes expanded during the search and possibly expanding dominated nodes as a consequence.
To this end, we focus on the procedures that involve solution update and show that, even without tie-breaking, $\mathit{Sol}$ still contains the optimal node pair when the algorithms terminate.

There are two procedures that involve updating $\mathit{Sol}$: ESU($x,d$) and Match($x,d'$).
Let $z$ and $x$ be two solution nodes associated with the same state where $f_1(z)=f_1(x)$ and $z$ is dominated by $x$, i.e., we have $f_2(x) < f_2(z)$.
Firstly, the termination criterion guarantees that both nodes will be explored, mainly because the search only terminates if $f_1(x)$ is \textit{strictly} larger than the optimal cost $\overline{f_1}$.
Secondly, both nodes are in the same priority queue, but in the absence of tie-breaking, the algorithm may explore and then find (dominated) node $z$ as a solution first.
We briefly discuss the situation of $x$ in each procedure as follows.\par

\textbf{ESU($x,d$) without Tie-breaking:}
Node $z$ as a solution has already updated $\overline{f_1}$ and $f^{sol}_2$ with $f_1(x)$ and $g_2(z)+\mathit{ub}^d_2(s(z))$ via the ESU strategy, respectively.
However, when $x$ is extracted from the priority queue as a (new) tentative solution, it will pass the upper bound tests in ESU($x,d$) and will be captured as a new solution.
This is mainly because $f_1(x)={\overline{f_1}}$ and $g_2(x)+\mathit{ub}^d_2(s(x)) < f^{sol}_2$ based on the assumption.
Note that we have $f_1(z)=f_1(x)$ and $f_2(x) < f_2(z)$ from the assumption which immediately follows $g_1(z)=g_1(x)$ and $g_2(x) < g_2(z)$ (both nodes are associated with the same state, i.e., $s(z) = s(x)$, and use the same heuristics).

\textbf{Match($x,d'$) without Tie-breaking:}
Node $z$ was successfully matched with $y$ and the Match procedure has already updated the global upper bound ${\overline{f_1}}$ with $g_1(z)+g_1(y)$ and solution cost $f^{sol}_2$ with $g_2(z)+g_2(y)$.
In the next iterations, when (non-dominated) node $x$ is being explored, the procedure will also find $x$ as a tentative solution because $g_1(x)+g_1(y)={\overline{f_1}}$ and $g_2(x)+g_2(y) \leq f^{sol}_2$ based on the assumption.
Therefore, we can see that the Match procedure is able to detect and replace dominated solutions with non-dominated ones.

Our discussion above showed that disabling tie-breaking in the $\mathit{Open}$ lists does not affect the ESU and Match strategies, and both {WC-EBBA*} and {WC-EBBA*}\textsubscript{par} still return an optimal solution path when they terminate.
However, expanding dominated nodes in the absence of tie-breaking would impact the number of nodes in node lists.
To this end, we discuss one possible refinement below.

\textbf{Store($x,d$) without Tie-breaking:}
When no tie-breaking is happening in the $\mathit{Open}$ lists, the lazy dominance check of {WC-EBBA*} would not be able to prune all dominated nodes arriving at the same state.
So there may be cases where the algorithm explores dominated nodes before non-dominated nodes of the same state.
Therefore, such nodes would inevitably be stored in the node list of corresponding states.
As shown in lines \ref{alg:path_store:1}-\ref{alg:path_store:2} of Procedure~\ref{alg:path_store}, Store($x,d$) can employ a constant time technique to refine dominated nodes in $\chi^d$ lists.
We explain this technique below.

Let $z$ and $x$ be two nodes associated with the same state, $s(z)=s(x)$, where $f_1(z)=f_1(x)$ and $z$ is dominated by $x$, i.e., we have $f_2(x) < f_2(z)$.
Both nodes are in the priority queue of direction~$d$.
Without tie-breaking, the algorithm processes the dominated node $z$ first and stores it at the end of the list $\chi^d((s(z))$.
In the next iterations, (non-dominated) node $x$ is processed and $g^d_{min}(s(x))$ gets updated with $g_2(x)$ (the lazy dominance check ensures $g_2(x) < g_2(z)$).
However, in the Store($x,d$) procedure, before adding a new node at the end of the list $\chi^d((s(x))$, we can perform a constant time dominance test to figure out whether the new node $x$ dominates the previously added node $z$.
This can be done by comparing the $f_1$-value of both nodes.
Note that nodes in $\chi$ lists are stored in the same order they get expanded and are already ordered by their primary cost $g_1$-values, so we already know $g_1(z) \leq g_1(x)$.
If both (new and last) nodes show the same primary cost, we can conclude that $z$ is dominated by $x$ since we then have $g_1(x)=g_1(z)$ and $g_2(x) < g_2(z)$.
Therefore, we just need to remove the last (dominated) node of $\chi^d((s(x))$ before inserting a new (non-dominated) node.
With this refinement, we can guarantee that there is at most one dominated node in the node list of each state during the search.
It is also worth mentioning that matching a dominated node of one list with a dominated node in the opposite direction might yield a feasible solution, but as explained above, such cases will be fixed via the ESU and Match strategies when their non-dominated nodes are explored.
In addition, {WC-EBBA*} and {WC-EBBA*}\textsubscript{par} are still correct even without the proposed refinement method, however, they would both consume less memory if the dominated nodes are removed from the node lists.
\section{Constrained Pathfinding and Bi-objective Search}
\label{sec:WC_BA}
\begin{figure}[t]
\centering
\includegraphics[width=0.4\textwidth]{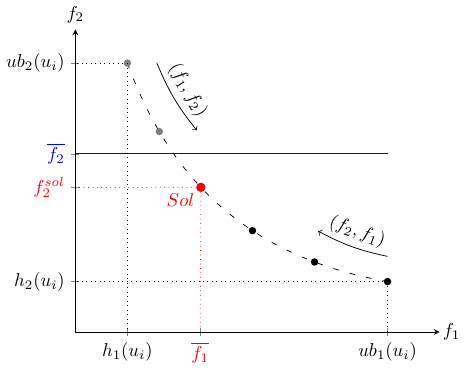} 
\caption{\small Schematic of a sample Pareto-optimal solution set (black, grey and red nodes) and also a solution $\mathit{Sol}$ in the set (the red node) with the optimal cost pair of ($\overline{f_1}$,$f^{Sol}_2$) for the WCSPP with the weight limit $\overline{f_2}$. Nodes in grey are solutions out-of-bounds for the WCSPP. $h$ and $ub$-values are the lower and upper bounds on the {\bf cost} of the initial state $u_i$, respectively.}
\label{fig:Pareto}
\end{figure}

The bi-objective shortest path problem (BOSPP) is considered conceptually closest to the Weight Constrained Shortest Path Problem (WCSPP) in terms of number of criteria involved.
This section investigates the relationship between the BOSPP and the WCSPP by revisiting the A*-based WCSPP method {WC-BA*}, adapted from the state-of-the-art {BOBA*} algorithm \cite{AhmadiTHK21_esa} in the BOSPP.
Compared to the WCSPP where we look for a single optimal path, methods to the BOSPP aim to find a representative set of Pareto-optimal solution paths, i.e., a set in which every individual (non-dominated) solution offers a path that minimises the bi-criteria problem in both \textit{cost} and \textit{weight}.
More formally, the typical $\mathit{start}$-$\mathit{goal}$ path $\pi$ is a Pareto-optimal solution if it is not dominated by any other $\mathit{start}$-$\mathit{goal}$ path, i.e., there is no path $\pi'$ between $\mathit{start}$ and $\mathit{goal}$ such that $\mathit{cost}_i(\pi') \leq \mathit{cost}_i(\pi)$ for $i=1,2$ and $\mathit{cost}_j(\pi') < \mathit{cost}_j(\pi)$ for at least one index $j$ \citep{Miettinen98Nonlinear}.
In the WCSPP, however, there is only one {\bf cost}-optimal solution path which is already non-dominated by our optimality requirement.
Figure~\ref{fig:Pareto} depicts the relationship between a sample set of Pareto-optimal solutions for the BOSPP (nodes in black, grey, and red) and the {\bf cost}-optimal solution of the WCSPP within the weight limit $\overline{f_2}$ (the red node marked $\mathit{Sol}$).
Note that Pareto-optimal solutions with $f_2$-values strictly larger than the upper bound $\overline{f_2}$ are out-of-bounds for the WCSPP (nodes in grey).
As we can see in this figure, $\mathit{Sol}$ is also Pareto-optimal in the BOSPP. Indeed, it is easy to see that this must always be the case, since the cost-optimal solution is non-dominated by definition.
Lemma~\ref{lemma:wcspp_bospp} formally states this observation.
\begin{lemma}
\label{lemma:wcspp_bospp}
Let $\Pi$ be the set of Pareto-optimal solution paths for the BOSPP between $\mathit{start}$ and $\mathit{goal}$.
Let $\pi^* \neq \emptyset$ be a {\bf cost}-optimal solution path for the WCSPP between $\mathit{start}$ and $\mathit{goal}$.
We always have $\pi^* \in \Pi$.
\end{lemma}
\begin{proof}
We prove this lemma by assuming the contrary, namely that $\pi^* \notin \Pi$.
The solution set $\Pi$ contains all (non-dominated) Pareto-optimal solution paths.
Hence, if $\pi^* \notin \Pi$, there should be a path in $\Pi$ that dominates $\pi^*$, otherwise, $\pi^*$ would have been in $\Pi$.
Let this path be $\pi' \in \Pi$.
Since $\pi'$ dominates $\pi^*$, we have $\mathit{cost_1}(\pi') < \mathit{cost_1}(\pi^*)$ or $\mathit{cost_1}(\pi') = \mathit{cost_1}(\pi^*)$ and $\mathit{cost_2}(\pi') < \mathit{cost_2}(\pi^*)$, which means $\pi'$ is lexicographically smaller than $\pi^*$.
However, the existence of $\pi'$ as a lexicographically smaller path would contradict the optimality of our solution path $\pi^*$.
Therefore, the {\bf cost}-optimal solution path $\pi^*$ is always one of the solution paths in the optimal set $\Pi$.
\end{proof}
The Pareto-optimality criteria above do not exclude equally optimum solutions, so we can refine solutions by making them \textit{cost-unique}, i.e., if there exist two (or more) solution paths with the same costs, the cost-unique criterion ensures that only one of the paths takes part in the solution set and there are no pairs of paths that can have identical $cost_1$ and $cost_2$.
$\Pi_{cu}$ in our notation refers to such cost-unique Pareto sets, so we always have $\Pi_{cu} \subseteq \Pi$.

In this paper, we consider two algorithms for the BOSPP that are able to discover a set of cost-unique Pareto-optimal solution paths.
Let $\Pi$ be the set of Pareto-optimal solution paths to a particular BOSPP.
The first algorithm, BOA* \cite{ulloa2020simple}, is a unidirectional A*-based search scheme that explores the graph in the forward direction (from $\mathit{start}$ to $\mathit{goal}$) in the traditional ($f_1,f_2$) order.
BOA* is able to find a set of cost-unique Pareto-optimal solution paths $\Pi_{cu}$.
The second algorithm, BOBA* \cite{AhmadiTHK21_esa}, is a bidirectional A*-based search scheme that explores the graph in both forward and backward directions concurrently.
In the forward search, like BOA*, BOBA* explores nodes in ($f_1,f_2$) order, and grows a subset of cost-unique Pareto-optimal solutions during the search.
Let this subset be $\Pi^f_{cu} \subseteq \Pi$.
In the backward search, BOBA* employs another BOA*-like search in the other possible objective ordering ($f_2,f_1$), and grows in parallel a complementary subset of the Pareto-optimal solutions.
Let this subset be $\Pi^b_{cu} \subseteq \Pi$.
BOBA* terminates with a set of cost-unique Pareto-optimal solution paths $\Pi_{cu}=\Pi^f_{cu} \cup \Pi^b_{cu}$ such that there is no equal solution path between $\Pi^f_{cu}$ and $\Pi^b_{cu}$ \cite{AhmadiTHK21_esa}.

From the WCSPP's point of view, if a {\bf cost}-optimal solution path $\mathit{Sol}$ exists, it will always be in the full optimal set $\Pi$ of BOA*, but only in one of the subsets of BOBA* ($\Pi^f$ or $\Pi^b$).
Figure~\ref{fig:Pareto} highlights that $\mathit{Sol}$ can be reached via two possible objective orderings: $\mathit{Sol}$ is the first valid solution in the ($f_1,f_2$) order, whereas it is the last valid solution in the ($f_2,f_1$) order.
Therefore, WCSPP's {\bf cost}-optimal solution will always be the first valid Pareto-optimal solution discovered by BOA*, and the first (resp. last) valid solution in the forward (resp. backward) search of BOBA*.
With this introduction, we now discuss how BOBA* can be adapted for the WCSPP, namely by revisiting its weight constrained variants {WC-BA*} presented in \citet{AhmadiTHK22_socs}.

\subsection{Constrained Pathfinding with WC-BA*}
{WC-BA*} is a bidirectional search algorithm that explores the search space in both objective orderings concurrently.
Algorithm~\ref{alg:rcba_high} shows the high-level design of {WC-BA*}.
Following BOBA*, {WC-BA*} performs parallel searches in both the initialisation phase and the bi-objective search.
In the initialisation phase, like other A*-based algorithms, {WC-BA*} needs to establish its heuristic functions in the forward and backward directions.
It then initialises a solution node pair $\mathit{Sol}$ and starts the constrained search by running two {WC-A*}-like searches in parallel: one in the forward direction and ($f_1,f_2$) order and the other in the backward direction and ($f_2,f_1$) order.
The algorithm shares search parameters such as best known solution in $\mathit{Sol}$ and the global upper bounds $\overline{\bf f}$ between the concurrent searches.
As an example, if the backward search finds a tentative solution lexicographically smaller than the current solution, both searches will have access to the updated value of the upper bound $\overline{f_1}$.
{WC-BA*} terminates as soon as one of the searches is complete.
We now explain each phase of {WC-BA*} as follows.

\begin{algorithm}[!t]
\small
\caption{WC-BA* High-level}
\label{alg:rcba_high}
  \DontPrintSemicolon 
    \SetKwBlock{DoParallel}{do in parallel}{end}
    \KwIn{A problem instance (G, {\bf cost}, $\mathit{start}$, $\mathit{goal}$) with the weight limit $W$}
    \KwOut{A node pair corresponding with the cost-optimal feasible solution $Sol$}
    ${\bf h}, {\bf ub}$, ${\bf \overline{f}}, S' \leftarrow$   Initialise WC-BA* (G, {\bf cost}, $\mathit{start}$, $\mathit{goal}, W$) \Comment*[r]{Algorithm~\ref{alg:Bi_search_init}}
    $Sol \leftarrow (\varnothing,\varnothing)$, $f^{sol}_2 \leftarrow \infty$ \;
    
    \DoParallel{
        Run a WC-A* search on (G, {\bf cost}, $\mathit{start}$, $\mathit{goal}$) in the forward direction and ($f_1,f_2$) order with global upper bounds ${\bf \overline{f}}$,\;
    \nonl \qquad heuristic functions $({\bf h}, {\bf ub})$ and initial solution $Sol$ with secondary cost $f^{sol}_2$. \Comment*[r]{Algorithm~\ref{alg:rc_ba}}
    
    \nonl \qquad Terminate the parallel search when the current search is complete. \;
        Run a WC-A* search on (G, {\bf cost}, $\mathit{start}$, $\mathit{goal}$) in the backward direction and ($f_2,f_1$) order with global upper bounds ${\bf \overline{f}}$,\;
    \nonl \qquad heuristic functions $({\bf h}, {\bf ub})$ and initial solution $Sol$ with secondary cost $f^{sol}_2$. \Comment*[r]{Algorithm~\ref{alg:rc_ba}}
    
    \nonl \qquad Terminate the parallel search when the current search is complete. \;
    }
\Return{$Sol$}
\end{algorithm}

%
\subsubsection{Initialisation}
The constrained search in {WC-BA*} is bidirectional, so the algorithm needs to conduct a set of bounded one-to-all searches in both directions to establish heuristic functions ${\bf h}$ and ${\bf ub}$.
As the main constrained search will be parallel, like BOBA*, we can take advantage of parallelism to compute bidirectional lower bounds faster.
Nevertheless, BOBA*'s initialisation phase has been designed to explore the entire search space in the BOSPP and is typically less informed than other bidirectional initialisation approaches for the WCSPP, like {WC-EBBA*}\textsubscript{par}.
To this end, {WC-BA*} follows the two-round search of Algorithm~\ref{alg:Bi_search_init} (presented for {WC-EBBA*}\textsubscript{par}) to establish its heuristic functions.
Note that both {WC-BA*} and {WC-EBBA*}\textsubscript{par} algorithms run two constrained A* searches in parallel, and thus the proposed search ordering in the initialisation phase of {WC-EBBA*}\textsubscript{par} will still result in consistent and admissible heuristics for {WC-BA*}.

Algorithm~\ref{alg:Bi_search_init} runs an $f_1$-bounded backward search and, in parallel, an $f_2$-bounded forward search to establish $h^f_1$ and $h^b_2$, respectively.
Therefore, the second phase of {WC-BA*} will have more informed heuristics for its forward search in the ($f_1,f_2$) order via $h^f_1$, and also for its backward search in the ($f_2,f_1$) order via $h^b_2$. 
There are two major differences if we compare the initialisation phase of BOBA* in \citet{AhmadiTHK21_esa} with the that of {WC-BA*}'s in Algorithm~\ref{alg:Bi_search_init}.
Firstly, the preliminary search on $\mathit{cost_2}$ in the first round of parallel searches in Algorithm~\ref{alg:Bi_search_init} is bounded by the weight limit $W$.
This limit is generally looser in the BOSPP (it is $\mathit{ub}^b_2(\mathit{goal})$).
Therefore, Algorithm~\ref{alg:Bi_search_init} will perform better in reducing the graph size. 
Secondly, the last round of the bounded searches in Algorithm~\ref{alg:Bi_search_init} tries to improve the global upper bound $\overline{f_1}$ by matching each partial path with its complementary shortest paths already obtained in the first round.
This strategy cannot be employed in the BOBA*'s initialisation phase, essentially because the initial upper bound on the primary global upper bound is fixed at the beginning (it is $\mathit{ub}^b_1(\mathit{goal})$) but will be updated over the course of BOBA*'s search with every new solution path.
Consequently, the second round of searches in Algorithm~\ref{alg:Bi_search_init} will be faster than that of BOBA*, and more importantly, the constrained search in the second phase of {WC-BA*} will start with a tighter upper bound on $\mathit{cost_1}$ of the optimal path, i.e., $\overline{f_1}$.
For more details, please see Section~\ref{sec:init_rcebba_par} where we discuss in detail the bounded searches involved in Algorithm~\ref{alg:Bi_search_init} as part of the initialisation phase of {WC-EBBA*}\textsubscript{par}.
\subsubsection{Constrained Search}
The {WC-BA*} algorithm executes two concurrent constrained searches for its second phase.
Following the search structure in BOBA*, we run one forward (constrained) A* search in the ($f_1,f_2$) order, and simultaneously, one backward (constrained) A* search in the ($f_2,f_1$) order, as shown in Algorithm~\ref{alg:rcba_high}.
To conduct these two constrained searches, we can use our {WC-A*} algorithm presented in Algorithm~\ref{alg:rc_ba} in the generic ($f_p,f_s$) order.
In this notation, we can achieve the backward search of {WC-BA*} by simply replacing $(p,s)=(2,1)$, denoting $\mathit{cost_2}$ and $\mathit{cost_1}$ as the \textit{primary} and \textit{secondary} attributes in the backward search, respectively.
The forward search, however, will be run in the traditional objective ordering in {WC-A*} with $(p,s)=(1,2)$.
Both searches communicate with each other and have access to shared parameters of the search including heuristic functions ${\bf h}$ and ${\bf ub}$, global upper bounds $\overline{\bf f}$, and best known solution $\mathit{Sol}$ with the costs $(\overline{f_1},f^{sol}_2)$.
Since the forward (constrained) search in {WC-BA*} is very similar to the constrained search of {WC-A*}, we focus on the operations that need refinements for the backwards direction.

\textbf{Algorithm Description:}
Consider Algorithm~\ref{alg:rc_ba} for a backward search in the ($f_2,f_1$) order, so we have $d=$~\textit{backward}.
The algorithm first calls the Setup(\textit{backward}) procedure (Procedure~\ref{alg:rc_ebba_setup}) to initialise essential data structures needed by the search, mainly the priority queue $\mathit{Open}^b$ and a search node associated with the $\mathit{goal}$ state.
To be able to communicate with the concurrent search, the algorithm then sets $d'$ to be the opposite direction \textit{forward}.
At this point, the algorithm can commence its constrained A* search with a non-empty priority queue.
In every iteration of the constrained search, the algorithm extracts the lexicographically smallest node from $\mathit{Open}^b$ in the ($f_2,f_1$) order.
Let the extracted node be $x$.
Similar to other A*-based algorithms presented in this paper, the constrained search of Algorithm~\ref{alg:rc_ba} can terminate if $f_2(x)$ is strictly larger than the global upper bound $\overline{f_2}$ (the early termination criterion).
However, since the algorithm never inserts invalid nodes into the priority queue, i.e., we always have $f_2(x) \leq \overline{f_2}$, the early termination criterion in the ($f_2,f_1$) is never satisfied and can be simply ignored.
In other words, line~\ref{alg:rc_ba:termination} of Algorithm~\ref{alg:rc_ba} can be removed from the backward search of {WC-BA*}.
Nonetheless, there is still a chance for $x$ to be an invalid node if it violates the global upper bound $\overline{f_1}$.
Note that $\overline{f_1}$ is a shared parameter and can be updated by both (forward and backward) searches.
In this case, if the algorithm finds $x$ out-of-bounds with $f_1(x) > \overline{f_1}$ via line~\ref{alg:rc_ba:prune1} of Algorithm~\ref{alg:rc_ba}, $x$ will be pruned safely (see Lemma~\ref{lemma:invalid}).
Note that we use this pruning only in the ($f_2,f_1$) order as it would not be effective in the other ordering ($f_1,f_2$).
This is mainly because both (parallel) searches update $\overline{f_1}$ with every tentative solution and, in the meantime, invalidate some of the (old) nodes in the priority queues with every update on $\overline{f_1}$.
The termination criterion in the ($f_1,f_2$) order automatically accounts for skipping such invalid nodes, however, this needs to be done manually in the ($f_2,f_1$) order as nodes (in the backward search) are no longer ordered based on their $f_1$-value.

Let $x$ be a valid node in the backward search, i.e., $x$ has successfully passed validity tests, so we know ${\bf f}(x) \leq \overline{\bf f}$.
In the next step, $x$ is tested for dominance via line~\ref{alg:rc_ba:prune2} of Algorithm~\ref{alg:rc_ba}.
In other words, we compare $x$ with the last node successfully expanded for state $s(x)$ in the backward direction.
Since we explore nodes in the non-decreasing order of their $f_2$-values in the ($f_2,f_1$) order, we can observe that nodes expanded for $s(x)$ are also ordered based on their $g_2$-values.
Therefore, if we keep track of the $g_1$-value of the last expanded node for $s(x)$ via the parameter $g^b_{\mathit{min}}(s(x))$, we can guarantee that $x$ is a weakly dominated node if $g_1(x) \leq g^b_{\mathit{min}}(s(x))$. 
Thus, $x$ can be safely pruned if it is weakly dominated by the last expansion of $s(x)$ (see Lemma~\ref{lemma:weakly_dominated} for the formal proof).
Otherwise, $x$ is a non-dominated node, and we can update $g^b_{\mathit{min}}(s(x)) \leftarrow g_1(x)$ via line~\ref{alg:rc_ba:min_r} of Algorithm~\ref{alg:rc_ba}. But before that, {WC-BA*} undertakes a strategy called \textit{Heuristic Tuning with the First expanded path (HTF)}, originally presented for BOBA* in \citet{AhmadiTHK21_esa}.
This tuning involves improving the (secondary) heuristic function of the opposite direction, here $h^f_1$, upon the first expansion of each state.
These steps are shown in lines~\ref{alg:rcba:tuning1}-\ref{alg:rcba:tuning2} of Algorithm~\ref{alg:rc_ba}.
Interested readers are refer to the WC-BA* paper \cite{AhmadiTHK22_socs} for the other tuning methods applicable to the algorithm (tuning with the last/all expanded paths).
In the next step (line~\ref{alg:rc_ba:early_sol}), the backward A* search tries to obtain a tentative solution by matching $x$ with its complementary shortest paths.
Following the ESU strategy in BOBA*, we only update $\mathit{Sol}$ if joining $x$ with its complementary shortest path on $\mathit{cost_p}$ (here $\mathit{cost_2}$) is valid.
Therefore, the ESU strategy in the ($f_2,f_1$) order is slightly different from ESU in the traditional ($f_1,f_2$) order.
We present the main steps in Procedure~\ref{alg:early_sol2} for the sake of clarity.
In the next step (line~\ref{alg:rc_ba:terminal}), according to Lemma~\ref{lemma:terminal_node}, the search skips expanding node $x$ if it is a terminal node, i.e., if we have $h^b_2(s(x))=\mathit{ub}^b_2(s(x))$.
Otherwise, if $x$ is not a terminal node, the algorithm expands $x$ in the ($f_2,f_1$) order for its last step via line~\ref{alg:rc_ba:expand}.
The node expansion procedure is identical between the two possible objective orderings, except for pruning by the last expanded node, where we use $g_1$-values to check new nodes for dominance.
We show all steps of the ExP($x,d$) procedure in the ($f_2,f_1$) order in Procedure~\ref{alg:expansion2}.
Finally, the backward search terminates if there is no node in $\mathit{Open}^b$ to explore.
We now discuss the correctness of the procedures involved in {WC-BA*}.

\textbf{The HTF method:}
Bidirectional search in different objective orderings provides {WC-BA*} with a great opportunity to improve the quality of its heuristic functions ${\bf h}$ and ${\bf ub}$.
Since the constrained searches of {WC-BA*} always prune dominated nodes and frequently update the global upper bound $\overline{f_1}$, it has more information about non-dominated paths to states than its preliminary heuristic searches in the initialisation phase.
Therefore, we can tune part of the initial heuristics to empower the pruning strategies during the concurrent constrained searches.
BOBA* performs this tuning in constant time by updating the secondary heuristics of the reverse direction.
We now explain the HTF method for {WC-BA*}.
Let $x$ be a non-dominated valid node extracted from the priority queue of the backward direction.
For this node, if $g^b_{\mathit{min}}(s(x))=\infty$, it means that there has not been any (successful) node expansion for $s(x)$ yet and $x$ will represent the first valid path from $\mathit{goal}$ to $s(x)$.
Since backward search explores nodes in the ($f_2,f_1$) order, A* guarantees that there would not be any path from $\mathit{goal}$ to $s(x)$ with a better $\mathit{cost_2}$ than $g_2(x)$.
With this observation, it is always safe to use this shortest path to update heuristics of the opposite direction, in our case via $h^f_2(s(x)) \leftarrow g_2(x)$ and $\mathit{ub}^f_1(s(x)) \leftarrow g_1(x)$.
We show one example of such updates during the search and formally prove the correctness of heuristic tuning of {WC-BA*} in Lemma~\ref{lemma:tuning}.

\textbf{Example:} Recall our {WC-A*} (forward) search on the graph of Figure~\ref{fig:example}.
If we aim to solve the problem using {WC-BA*} bidirectionally, we also need to find backward heuristics ${\bf h}^b$ and ${\bf ub}^b$.
Imagine we have obtained these heuristics simply via two one-to-all searches from $u_s$ (without any graph reduction technique).
Now consider state $u_2$ on this graph.
For this state, we can see that path $\{u_s,u_1,u_2\}$ is the shortest path between $u_s$ and $u_2$ on $\mathit{cost_1}$.
This shortest path initially sets $(h^b_1(u_2),\mathit{ub}^b_2(u_2)) \leftarrow (2,6)$.
In the next phase, we just run two iterations of the forward search (all iterations were given in Figure~\ref{fig:example}).
When the main constrained search starts, the forward search of {WC-BA*} (which is a {WC-A*} search) realises that the node arriving at $u_1$ is invalid and should be pruned (see iteration~1 in Figure~\ref{fig:example}).
This means that state $u_2$ is no longer reachable from its shortest path (via $u_1$) during the forward constrained search.
However, {WC-BA*} (still in its forward search) explores a node associated with $u_2$ in the second iteration with costs ${\bf g}(x)=(3,4)$.
At this point, since this is the first time we see state $u_2$ getting expanded, the heuristic tuning allows the constrained search to improve the initial lower and upper bounds by updating $(h^b_1(u_2),\mathit{ub}^b_2(u_2)) \leftarrow (3,4)$.
As a result, whenever the backward search wants to explore state $u_2$, it would be able to make stronger assumptions on the costs of paths via $u_2$ and prune more nodes.
\begin{lemma}\label{lemma:tuning}
Let ($f_p,f_s$) be the objective ordering in the A* search of direction~$d$.
Let $d'$ be the opposite direction of $d$ in which we run another A* search in ($f_s,f_p$) order.
Tuning heuristics $h^{d'}_p$ and $\mathit{ub}^{d'}_s$ 
during the search of direction~$d$ maintains the correctness of A* searches in both directions.
\end{lemma}
\begin{proof}
Let $h^{d}_p$ and $h^{d'}_s$ be admissible and consistent heuristic functions.
The A* search in direction~$d$ is led by $f_p$-values using the heuristic function $h^{d}_p$.
This heuristic function is not tuned by the search, so the A* search in direction~$d$ remains correct.
The A* search in direction~$d'$ is led by $f_s$-values using the heuristic function $h^{d'}_s$.
The search in direction~$d$ does not tune the heuristic function $h^{d'}_s$ (it only updates $h^{d'}_p$).
Therefore, the A* search in direction~$d'$ is still correct.
\end{proof}

It is worth mentioning that updating non-primary lower bounds may change node ordering in the priority queue in terms of tie-breaking.
Imagine an A* search in the ($f_p,f_s$) order in direction~$d$ with an online tuning procedure on $h^d_s$ (the non-primary heuristic) via the opposite search.
A* has generated two nodes $x$ and $y$ associated with the same state: node $x$ before tuning and node $y$ after tuning.
Both nodes are residing in the priority queue.
Let $g_p(y) = g_p(x)$ and $g_s(y) < g_s(x)$, which means we expect $y$ gets expanded earlier than $x$ if we do tie-breaking on $f_s$ in $\mathit{Open}^d$.
However, since $y$ has used more informed (tuned) heuristics to establish its $f_s$-value, we might have $f_s(x) < f_s(y)$, which means $x$ would appear for expansion earlier than~$y$.
Therefore, the search might inevitably expand such dominated nodes before expanding the non-dominated one.
Although we later show that {WC-BA*} returns a {\bf cost}-optimal solution even without tie-breaking, we briefly explain two approaches that avoid expanding such dominated nodes after heuristic tuning:
(1) tie-breaking on $g_s$-value; 
(2) evaluating nodes on their $f_s$-value upon their extraction from the priority queue.
In the first approach, we ensure that non-dominated nodes (with smaller $g_s$-values) will always appear first.
In the second approach, we update $f_s$-value of the node and return it into the priority queue if the original $h_s$-value of the node (retrievable via $f_s(x)-g_s(x)$ for a typical node $x$) is smaller than the (possibly tuned) lower bound in the $h^d_s$ function.
In both cases, {WC-BA*} does not expand dominated nodes after heuristic tuning.

\textbf{The ESU strategy:}
Lemma~\ref{lemma:early_sol} guarantees that if matching $x$ with the complementary shortest path on $\mathit{cost_p}$ (primary objective) yields a valid $\mathit{start}$-$\mathit{goal}$ path, A* has found a tentative solution.
This means that, if the objective ordering is ($f_2,f_1$), only complementary shortest paths on $\mathit{cost_2}$ can take part in solution path construction.
However, the constrained search can still use complementary shortest paths on $\mathit{cost_1}$ to construct joined paths and possibly improve (reduce) the global upper bound $\overline{f_1}$.
Procedure~\ref{alg:early_sol2} shows in detail how the backward search of {WC-BA*} handles solution updates.
ESU($x,d$) in the ($f_2,f_1$) order first joins $x$ with the complementary shortest path on $\mathit{cost_2}$.
For this joined path we have $f'_1=g_1(x)+\mathit{ub}^b_1(s(x))$ and $f_2(x)=g_2(x)+h^b_2(s(x))$.
Since $x$ is already a valid node (it has been checked for validity before the ESU strategy), we know that $f_2(x) \leq \overline{f_2}$.
Therefore, the joined path is a tentative solution path if $f'_1 \leq \overline{f_1}$.
In this case, the procedure updates the current solution $\mathit{Sol}$ if the joined path is lexicographically smaller than the best known path with costs ($\overline{f_1},f^{sol}_2$).
Otherwise, if joining $x$ with the complementary $\mathit{cost_2}$-optimal path does not yield a valid $\mathit{start}$-$\mathit{goal}$ path, the procedure matches $x$ with the other available optimum path on $\mathit{cost_1}$.
For this joined path we have $f_1(x)=g_1(x)+h^b_1(s(x))$ and $f'_2=g_2(x)+\mathit{ub}^b_2(s(x))$.
Therefore, the procedure can improve the global upper bound $\overline{f_1}$ if the recent joined path is valid and (strictly) shorter than the current solution path, i.e., if $f'_2 \leq \overline{f_2}$ and $f_1(x) < \overline{f_1}$.
In this case, the procedure also sets $f^{sol}_2 \leftarrow \infty$ to allow a tentative solution with cost $f_1(x)$ to be chosen freely in the next iterations of the search.

\begin{figure}[t]
\centering
\includegraphics[width=0.40\textwidth]{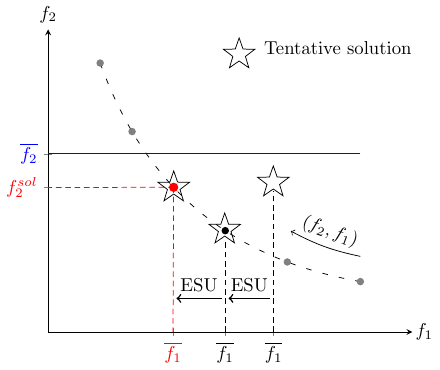} 
\caption{\small A schematic of three consecutive updates to the global upper bound $\overline{f_1}$ via partial paths matching and the ESU strategy in the backward search of {WC-BA*} and in the ($f_2,f_1$) order.}
\label{fig:Pareto_ESU}
\end{figure}

\textbf{Suboptimal solutions:}
There is one important difference between the optimality of tentative solutions obtained in the two possible orderings (Procedures~\ref{alg:early_sol}~and~\ref{alg:early_sol2}).
Consider two independent {WC-A*} searches in different orders, one in the ($f_1,f_2$) order and the other in ($f_2,f_1$).
If {WC-A*} finds a tentative solution via ESU in the ($f_1,f_2$) order, we can guarantee that $\mathit{cost_1}$ of the optimal solution is determined.
If {WC-A*} finds a tentative solution via ESU in the ($f_2,f_1$) order, since nodes are not ordered based on $f_1$-values, we cannot guarantee the $\mathit{cost_1}$-optimality until the end of the search.
Figure~\ref{fig:Pareto_ESU} shows such case in the presence of Pareto-optimal solutions.
Consider the backward search of {WC-BA*} in the ($f_2,f_1$) order (no operation in the forward direction).
When {WC-BA*} completes its initialisation phase, it has probably defined a tighter upper bound $\overline{f_1}$ via path matching during the preliminary heuristics searches.
The rightmost $\medwhitestar$ symbol in Figure~\ref{fig:Pareto_ESU} depicts one such initial solution.
Note that since the initial solution is not necessarily Pareto-optimal, we can show it as a dominated node above the Pareto-front.
Given $\overline{f_1}$ (resulting from the initial solution) as the best known upper bound on $cost_1$ of the optimal path, we can invalidate some of the Pareto-optimal solutions showing $f_1$-values larger than the initial upper bound $\overline{f_1}$.
We show these invalid solution nodes on the right side of the rightmost $\overline{f_1}$ in grey.
Note that solution nodes above $\overline{f_2}$ are already invalid, as their $f_2$-value is out-of-bounds.
During the constrained search in the ($f_2,f_1$) order, the ESU strategy finds a tentative solution better than the initial solution, but unfortunately without any guarantee on $\mathit{cost_1}$-optimality when the search is still in progress (the $\medwhitestar$ symbol in the middle).
{WC-BA*} eventually terminates after one more solution update (the leftmost $\medwhitestar$ symbol) via ESU.
At this point (termination), {WC-BA*} guarantees that the final solution (the red node) is {\bf cost}-optimal.
Interestingly, we can see that other solutions discovered in the ($f_2,f_1$) order have been Pareto-optimal (after tie-breaking).
Therefore, in contrast to {WC-A*} where the algorithm returns an optimal solution only after termination, one can stop {WC-BA*} (in difficult problems for example) when the discovered solution in $\mathit{Sol}$ (via the backward search) is enough.
This anytime behaviour can be valuable in practice, where a trade-off between speed and solution quality must be found.
Note that when the search on ($f_2,f_1$) finds a tentative solution with cost $f^{sol}_2$, we can guarantee that the next potential node has a $f_2$-value no smaller than $f^{sol}_2$, which means that the next solution will be closer to $\overline{f_2}$ than the current solution.
Therefore, one such stopping criterion for the suboptimal search can be finding a solution node at some distances of $\overline{f_2}$ during the backward search in the ($f_2,f_1$) order.
The returned solution node is guaranteed to be Pareto-optimal.

\textbf{Expand and Prune}:
There is no difference in expanding nodes in either of the possible objective orderings, however, there is one minor difference in their dominance tests.
A node $x$ in the ($f_1,f_2$) order is weakly dominated by the last expanded node for state $s(x)$ if $g_2(x) \leq g^d_{\mathit{min}}(s(x))$.
In the ($f_2,f_1$) order, nodes associated with the same state are expanded based on their $g_2$-value whereas $g^d_{\mathit{min}}(s(x))$ keeps track of the (non-primary) $g_1$-value of the last expanded node for $s(x)$.
Therefore, as shown in Procedure~\ref{alg:expansion2}, the backward node $y$ will be weakly dominated by the last node expanded for $s(y)$ if $g_1(y) \geq g^b_{\mathit{min}}(s(y))$. 
See Lemma~\ref{lemma:last_sol} for the formal proof.
\begin{theorem}
{WC-BA*} returns a node corresponding to a {\bf cost}-optimal solution path for the WCSPP.
\end{theorem}
\begin{proof}
{WC-BA*} conducts two {WC-A*}-like searches in parallel.
We already proved in Lemma~\ref{lemma:tuning} that the A* searches are still correct after heuristic tuning.
Lemma~\ref{lemma:early_sol} proves that the ESU strategy in both objective orderings is correct, and the algorithm correctly tracks the best tentative solutions of both directions.
Lemma~\ref{lemma:terminal_node} confirms that terminal nodes (in any order) are part of a tentative solution and their expansion is not necessary.
Lemma~\ref{lemma:astar_terminate} proves that both A* searches of {WC-A*} terminate if they prune weakly dominated nodes.
Therefore, we just need to show the correctness of our pruning methods after heuristic tuning.

Suppose a {WC-A*} search in direction~$d$ and objective ordering ($f_p,f_s$).
A* in this direction expands nodes in non-decreasing order of $f_p$-values.
Also suppose that pruning nodes via functions $h^{d'}_p$ and $\mathit{ub}^{d'}_p$ is correct before tuning (based on Lemmas~\ref{lemma:invalid}-\ref{lemma:weakly_dominated}).
We show that the affected pruning strategies will be correct if we tune $h^{d'}_p$ and $\mathit{ub}^{d'}_p$ functions in the constrained search of direction~$d$.

Pruning by global upper bounds:
Assume $h^d_p$ is a consistent and admissible heuristic function used in the constrained A* search of direction~$d$ in ($f_p,f_s$) order.
This heuristic function is not tuned during the search.
Given the correctness of pruning methods before tuning, when the search expands the first node $x$ of state $s(x)$ in direction~$d$, A* (with the admissible and consistent heuristic function $h^d_p$) guarantees that there is no valid path from the initial state to $s(x)$ with a $\mathit{cost_p}$-value better than $g_p(x)$.
Hence, the heuristic function $h^{d'}_p$ will remain admissible if we update $h^{d'}_p(s(x))$ with $g_s(x)$.
Therefore, given the admissibility of $h^{d'}_p$, pruning nodes by validity tests in direction~$d'$ is still correct.

Pruning by states' upper bounds:
when A* expands the first node $x$ of state $s(x)$ in direction~$d$, the concrete cost $g_s(x)$ sets an upper bound on $\mathit{cost_s}$ of nodes associated with $s(x)$ essentially because nodes arriving at $s(x)$ in direction~$d$ with a larger $g_s$-value than $g_s(x)$ will be dominated by the first expanded node $x$ and would not be part of any solution path
(see Lemma~\ref{lemma:dominated}).
Therefore, pruning based on the upper bound function $\mathit{ub}^{d'}_p$ is still correct after tuning.
\end{proof}
Note that there is one notable difference between the parallel search in {WC-EBBA*}\textsubscript{par} and {WC-BA*} in their high-level structure.
In {WC-EBBA*}, the algorithm does not terminate unless both (concurrent) searches have successfully been terminated.
In {WC-BA*}, however, the algorithm can terminate once it finds one of the searches terminated.
This termination criterion in the high-level structure of {WC-BA*} is always correct.
This is because each individual search in {WC-BA*} is able to solely search for an optimal solution.
For example, we can switch off one of the searches and still obtain an optimal solution via the other search.
In other words, when one of the searches of {WC-BA*} terminates early, it guarantees the optimality of the solution in $\mathit{Sol}$ and thus the concurrent search can be stopped accordingly.
\subsection{{WC-A*} and {WC-BA*} without Tie-breaking}
\label{sec:WC_A_NoTieBreak}
Our search strategy allows us to disable tie-breaking in the $\mathit{Open}$ lists of both {WC-A*} and {WC-BA*} for faster queue operations.
To achieve this, we can simply order nodes based on their $f_1$-value in {WC-A*} (and similarly in the forward search of {WC-BA*}).
For the backward search of {WC-BA*}, we only order nodes based on their $f_2$-values.
However, we will likely end up expanding more nodes in both algorithms, essentially because nodes are no longer lexicographically ordered and dominated nodes may get expanded earlier than non-dominated nodes.
Lemma~\ref{lemma:WC_A_notie} formally states the correctness of {WC-A*} and {WC-BA*} without tie-breaking.
\begin{lemma} 
\label{lemma:WC_A_notie}
{WC-A*} and {WC-BA*} return {\bf cost}-optimal solution paths even without tie-breaking in their $\mathit{Open}$ lists.
\end{lemma}
\begin{proof}
Assume we run A* in ($f_p,f_s$) order.
Let $z$ and $x$ be two solution nodes associated with the same state (in direction~$d$) where node $z$ is dominated by $x$, i.e., we have $f_p(x)=f_p(z)$ and $f_s(x) < f_s(z)$.
Without any tie-breaking, the search in direction~$d$ may temporarily accept the dominated node $z$ as a tentative solution first, essentially via the ESU strategy.
Since both $x$ and $y$ have the same primary cost estimate, i.e., $f_p(x)=f_p(z)$, the termination criterion in our constrained A* searches guarantees that $x$ will definitely be checked before termination (the search terminates if $f_p(x)$ is \textbf{strictly} larger than $\overline{f_p}$).
In the next iterations, when $x$ is extracted from $\mathit{Open}^d$, the search performs a quick dominance check by comparing the $g_s$-value of the newly extracted node $x$ against that of the previously expanded node for $s(x)$ (node $z$) stored in $g^d_{\mathit{min}}$. $x$ is not dominated by the last expanded node, as we already have $g_s(x) < g_s(z)$.
$x$ is also a valid node because it is lexicographically smaller than the last solution node $z$.
Hence, the search passes $x$ to the ESU($x,d$) procedure as a valid non-dominated node (see Procedures~\ref{alg:early_sol}~and~\ref{alg:early_sol2}).
Node $x$ is a tentative solution for the same reason that $z$ was ($x$ is lexicographically smaller than $z$).
The ESU strategy then checks the new solution node $x$ against the last solution node stored in $\mathit{Sol}$ and breaks the tie by replacing (dominated) node $z$ with the lexicographically smaller node $x$ in $\mathit{Sol}$. 
We see that the search is always able to detect dominated nodes (sometimes after their expansion) and refine the search parameters $g^d_{\mathit{min}}$ and $\mathit{Sol}$ correspondingly.
Therefore, both algorithms are able to compute a {\bf cost}-optimal solution path even without tie-breaking.
\end{proof}
\section{Compact Backtracking with Dynamic arrays}
\label{sec:compact_backtracing}
We illustrate our compact backtracking approach by exploring the paths between states $u_s$ and $u_g$ in the example graph of Figure~\ref{fig:example_mem} using the {WC-A*} algorithm and two dynamic arrays per state, namely \texttt{parent\_state} and \texttt{parent\_path\_id}. 
To implement this strategy, we consider a memory manager for each search that is responsible for node generation and node recycling.
Each node in this analysis occupies a fixed-size memory block.
Each iteration is explained as follows:
\begin{figure}[t]
\begin{subfigure}{0.34\textwidth}
\begin{tikzpicture}[
roundnode/.style={circle, draw=black,  thick, minimum size=5mm},
roundnode2/.style={circle, draw=gray,  thick, minimum size=5mm},
scale=1, every node/.style={scale=1.0}]
\footnotesize

\node[roundnode,align=center] at (-3, 0) (s1) {$u_s$} ;
\node at (-3, 0.5) {$x_1$};
\node[roundnode,align=center] at (-1.5, 1.75) (s2) {$u_1$};
\node[roundnode,align=center] at (0.0, 0) (s3) {$u_2$};
\node[roundnode,align=center] at (1.5, 1.75) (sg) {$u_g$};

\draw[->,-latex,  thick] (s1) edge[auto=left] node[anchor=south, sloped]{$x_2$} (s2);
\draw[->,-latex,  thick] (s1) edge[auto=left] node{$x_3$} (s3);
\draw[->,-latex,  thick] (s2) edge[auto=left] node[anchor=south, sloped]{$x_5$} (s3);
\draw[->,-latex,  thick] (s3) edge[auto=left] node[anchor=south, sloped]{$x_6,x_7$} (sg);
\draw[->,-latex,  thick] (s2) edge[auto=left] node[anchor=south, sloped]{$x_4$} (sg);


\end{tikzpicture}

\end{subfigure}
\begin{subfigure}{0.20\textwidth}
\centering
\footnotesize
\renewcommand{\arraystretch}{1}

\begin{tabular}{|l| l |}
    \toprule
    Mem. & Nodes\\
    \midrule
     M1 & $x_1,x_4$\\
     M2 & $x_2,x_6$\\
     M3 & $x_3,x_7$\\
     M4 & $x_5$\\
    \bottomrule
\end{tabular}

\end{subfigure}
\begin{subfigure}{0.43\textwidth}
\centering
\footnotesize
\renewcommand{\arraystretch}{1}

    
\begin{tabular}{|l| *{4}{c|} }
    \toprule
    Path info. / state & $u_s$ & $u_1$ & $u_2$ & $u_g$\\
    \midrule
    \texttt{Processed\_node} & $x_1$ & $x_2$ & $x_3,x_5$ & $x_4,x_6,x_7$ \\
    \midrule
    \texttt{parent\_state} & $[\varnothing]$ & $[u_s]$ & $[u_s,u_1]$ & $[u_1,u_2,u_2]$\\
    \midrule
    \texttt{parent\_path\_id}& $[0]$ & $[1]$ & $[1,1]$ & $[1,1,2]$ \\
    \bottomrule
\end{tabular}
\end{subfigure}
\caption{\small An example graph for solution path construction with memory recycling and dynamic parent arrays. We show nodes generated during the expansions via $x_i$ on edges. The table in the middle shows the memory blocks (Mem.) allocated to each node, and the table on the right shows the status of the parent arrays after all nodes have been processed.}
\label{fig:example_mem}
\end{figure}
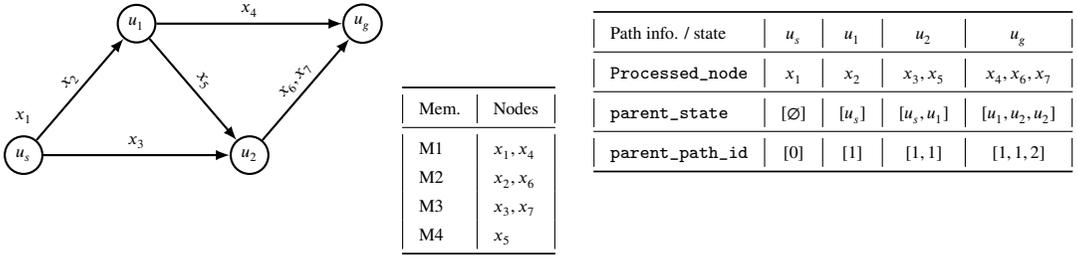
\begin{enumerate}[nosep]
    \item Assume we start expansion with node $x_1$ in the first iteration.
    $x_1$ occupies the memory block M1.
    This node is associated with $u_s$ and its expansion generates two new nodes $x_2$ and $x_3$ corresponding with states $u_1$ and $u_2$.
    $x_2$ and $x_3$ occupy memory blocks M2 and M3, respectively.
    When the first iteration is complete and $x_1$ is successfully expanded, the algorithm stores the backtracking information of $x_1$ and recycles $x_1$'s memory M1 via the memory manager.
    Since $x_1$ was the initial node, we update \texttt{parent\_state($u_s$)[1]$\leftarrow\varnothing$} and \texttt{parent\_path\_id($u_s$)[1]$\leftarrow$0}, meaning that $x_1$ does not have parent node/path.
    At the end of this iteration, we have two nodes ($x_2$ and $x_3$) in the priority queue and one memory block (M1) available.
    \item We next pick $x_2$ from the priority queue and expand it.
    $x_2$ is associated with $u_1$, so its expansion reaches states $u_g$ and $u_2$.
    As we have already processed $x_1$, 
    the memory manager uses the $x_1$'s recycled memory M1 and allocates it to one of the new nodes.
    Let $x_4$ be this new node associated with $u_g$.
    The memory manager then allocates a new memory block M4 to the other new node.
    Let $x_5$ be this new node with state $u_2$.
    To complete the expansion, the algorithm updates both nodes $x_4$ and $x_5$ with the information of the extended paths (such as costs, corresponding state, etc.) and then inserts them into the priority queue.
    Since $x_2$ is now successfully expanded, we can store its backtracking information in $u_1$'s parent arrays and recycle $x_2$'s memory M2 via the memory manager.
    $x_2$ resulted from $x_1$'s expansion, and $x_1$ was the first path of $u_s$, so we can update the parent arrays of $u_1$ (i.e., $x_2$'s state) with \texttt{parent\_state($u_1$)[1]$\leftarrow u_s$} and \texttt{parent\_path\_id($u_1$)[1]$\leftarrow$1}.
    At the end of this iteration, we have three nodes ($x_3$, $x_4$, $x_5$) in the priority queue and one memory block (M2) available.
    \item In the next iteration, we expand $x_3$ with state $u_2$ (the first node in the queue).
    $x_3$ discovers only one adjacent state (i.e., $u_g$), so $x_3$'s expansion only needs one new node.
    Now, since the memory manager sees $x_2$ as a processed node, it can allocate $x_2$'s recycled memory (M2) to the new node. 
    Let this new node be $x_6$ with state $u_g$.
    The algorithm then updates node $x_6$'s information and adds it to the priority queue.
    After expanding $x_3$ (with state $u_2$), we update \texttt{parent\_state($u_2$)[1]$\leftarrow u_s$} and \texttt{parent\_path\_id($u_2$)[1]$\leftarrow$1}, and then recycle its memory M3 via the memory manager.
    At the end of this iteration, we have three nodes ($x_4,x_5,x_6$) in the queue and one memory block (M3) available.
    \item We next pick $x_4$ from the priority queue for expansion. 
    $x_4$ is a terminal node, so its expansion does not add any new node to the queue, but frees its memory block M1.
    We then store the backtracking information for $x_4$ by updating \texttt{parent\_state($u_g$)[1]$\leftarrow u_1$} and \texttt{parent\_path\_id($u_g$)[1]$\leftarrow$1}, meaning that this first solution node (associated with $u_g$) has been the result of extending the first path of $u_1$.
    At the end of this iteration, we have two nodes in the queue ($x_5,x_6$) and two memory blocks (M3 and M1) available.
    \item So far, we have captured one path per state.
    The algorithm extracts $x_5$ (associated with $u_2$) from the queue for this iteration.
    $x_5$'s expansion leads to $u_g$ and thus a new node is needed.
    The memory manager then allocates one of the available memory blocks M3 to a new node.
    Let this new node be $x_7$ with state $u_g$.
    Once the expansion is complete and $x_7$ is inserted into the priority queue, the memory manager recycles $x_5$'s allocated memory M4.
    To store the backtracking information, since $x_5$ was the second expansion of $u_2$ (the first expansion was $x_3$), we add one more entry to the parent arrays of $u_2$ and update its second index (resp. second expanded path of $u_2$) with \texttt{parent\_state($u_2$)[2]$\leftarrow u_1$} and \texttt{parent\_path\_id($u_2$)[2]$\leftarrow$1}.
    At the end of this iteration, we have two nodes in the queue ($x_6,x_7$) and two memory blocks (M1 and M4) available.
    \item We now extract $x_6$ (associated with $u_g$) from the queue.
    $x_6$ is a terminal node, so the algorithm just needs to store its backtracking information and recycle $x_6$'s allocated memory M2.
    $x_6$ is the second node we explore with state $u_g$ (the first one was $x_4$).
    Hence, we update \texttt{parent\_state($u_g$)[2]$\leftarrow u_2$} and \texttt{parent\_path\_id($u_g$)[2]$\leftarrow$1}, meaning that the second solution node has been the result of extending the first path of $u_2$.
    We now only have one node in the queue ($x_7$) and three memory blocks (M1, M4 and M2) available.
    \item For the last iteration, we extract $x_7$ with state $u_g$.
    Again, we can see that $x_7$ is a terminal node and does not need to be expanded.
    Similar to the previous iteration, the algorithm recycles the memory allocated to the processed node (M3) and then stores its backtracking information by updating \texttt{parent\_state($u_g$)[3]$\leftarrow u_2$} and \texttt{parent\_path\_id($u_g$)[3]$\leftarrow$2}, meaning that our third solution node has been the result of extending the second path of $u_2$.
    At the end of this iteration, the priority queue is empty, and we have all memory blocks M1-4 available.
\end{enumerate}
The example above shows that we can effectively re-allocate the memory of processed nodes in the search and reduce the space requirement of exhaustive search in A*.
More precisely, the memory manager of the algorithm only requires four memory blocks to handle the seven nodes generated in the search.
We also showed in the example graph of Figure~\ref{fig:example} the memory blocks allocated to each node and the status of the parent arrays when the search terminates.
As a further optimisation, instead of \texttt{parent\_state} arrays, one can store the index of incoming edges (which are usually very small integers).
\section{Extended Experimental Analysis}
This section provides more detailed discussions on benchmark instances and also experimental results of this study on both algorithmic performance of the algorithms and the impacts of priority queues on constrained search with A*. 
We start with benchmark setup.
\subsection{Benchmark Setup and Details:}
\label{sec:benchmark_SM}
This section reviews the details of our 2000 WCSP instances in \citet{AhmadiTHK22_socs}.
Among the benchmarks used in the literature, the instances of \citet{sedeno2015enhanced} are among the largest ones.
The benchmark contains 440 WCSPP instances from large road networks in the 9th DIMACS Implementation Challenge.
Table~\ref{table:DIMACS_spec} shows the details of these DIMACS maps, with the largest map (USA) containing around 24~million nodes and 57~million edges.
The benchmark instances of \citeauthor{sedeno2015enhanced} include five ($\mathit{start}$, $\mathit{goal}$) pairs in each map, 
$\mathit{start}$ is always the first state (with \texttt{id}=1) but $\mathit{goal}$ is chosen from $\{n,n/2,n/log(n),\sqrt{n},log(n)\}$ where $n$ is the total number of states (last state \texttt{id} in the graph).
In this setting, we can see that instances are order based on state ids. 
However, state ids are essentially random and this setting might not produce challenging WCSPP instances, even if we arbitrary pick (\texttt{first id}, \texttt{last id}) as the farthest ids in the graph.
We can also observe from the recent results in \citet{AhmadiTHK21} that such instances are no longer considered hard WCSPP instances mainly because there now exists a fast algorithm that can solve all instances under 10 minutes.
To this end, we designed a larger benchmark set with 2000 easy-to-hard realistic instances to better evaluate our new algorithms under challenging cases.
We briefly explain our benchmark setup as follows.
\begin{table}[!t]
\caption{\small Details of 12 USA road networks as part of the DIMACS implementation challenge.}
\label{table:DIMACS_spec}
\begin{multicols}{2}
\centering
\small
\begin{adjustbox}{width=1\textwidth}
\renewcommand{\arraystretch}{1}
\begin{tabular}{l l *{2}{r}}
\headrow
    Map & Description & \multicolumn{1}{c}{\#States} & \multicolumn{1}{c}{\#Edges} \\
    \hline
    NY & New York City & 264,346 & 730,100\\
    BAY	& San Francisco Bay Area & 321,270 & 794,830\\
    COL & Colorado & 435,666 & 1,042,400\\
    FLA & Florida & 1,070,376 & 2,687,902\\
    NW & Northwest USA & 1,207,945 & 2,820,774\\
    NE & Northeast USA & 1,524,453 & 3,868,020\\
    \hline   
\end{tabular}
\begin{tabular}{l l *{2}{r}}
\headrow
    Map & Description & \multicolumn{1}{c}{\#States} & \multicolumn{1}{c}{\#Edges}\\
    \hline
    CAL & California and Nevada & 1,890,815 & 4,630,444\\
    LKS & Great Lakes & 2,758,119 & 6,794,808 \\
    E &	Eastern USA & 3,598,623 & 8,708,058\\
    W & Western USA & 6,262,104 & 15,119,284\\
    CTR	& Central USA & 14,081,816 & 33,866,826\\
    USA	& Full USA & 23,947,347 & 57,708,624\\
    \hline   
\end{tabular}
\end{adjustbox}
\end{multicols}
\end{table}

We start with the existing benchmark instances originally designed to evaluate solution approaches to the BOSPP \cite{Sedeno-NodaC19,AhmadiTHK21_esa}.
These instances together include 1000 random $\mathit{start}$-$\mathit{goal}$ pairs from 10 large road networks of the DIMACS challenge with (\textit{distance}, \textit{time}) as objectives.
Both attributes are represented with non-negative integers.
We also use the competition's random pair generator to produce an additional set of 200 random instances for two other large maps of the DIMACS challenge: NW and USA (100 instances each).
We then solved all the 1200 random instances in the bi-objective setting using BOBA* and then sorted the instances of each map based on their number of Pareto-optimal solution paths.
We use the number of Pareto-optimal paths as a measure of difficulty, to avoid generating too many easy instances with few optimal solutions.
In general, if we assume instances with more Pareto-optimal solution paths explore a larger search space (which is usually the case in practice), such instances would likely appear challenging for the WCSPP, mainly because of the shared search space between the two problems.
After ordering all (solved) instances of each map on their number of Pareto-optimal solutions, we evenly sample 10 instances (out of 100) for each map such that they together present a range of easy-to-hard point-to-point bi-criteria problems. 
To further challenge the algorithms on very large graphs, we pick five more instances for the USA map.
Thus, in total, we have 125 $\mathit{start}$-$\mathit{goal}$ pairs over 12 maps.
We then double the number of instances by adding a set of reversed pairs, i.e., a set of 125 $\mathit{goal}$-$\mathit{start}$ pairs.
Studying both normal and reversed pairs will help us to better measure the performance of unidirectional search algorithms in both search directions.
For the last step, we need to determine weight limits (for each pair).
Given the tightness level $\delta \in\{10\%,20\%,\dots,70\%,80\%\}$, we then have eight WCSPP instances for each pair, resulting in 2000 test cases in total.
\subsection{More Detailed Runtime and Memory Analysis}
\label{sec:time_SM}
\textbf{Batch analysis:}
To better compare the studied algorithms in terms of runtime, we present the cumulative runtime of each algorithm over all instances per map in Table~\ref{table:runtime_map} with maps ordered based on their size.
\begin{table}[t]
\caption{\small Cumulative runtime of the algorithms in \textit{minutes} per map (total time needed to solve all instances of each map considering 60 minutes timeout for the runtime of unsolved cases).}
\label{table:runtime_map}
\small
\begin{adjustbox}{width=1\textwidth}
\renewcommand{\arraystretch}{1}
\begin{tabular}{lrrrrrrrrrrrr}
\headrow
Alg. / Map           & \multicolumn{1}{c}{NY} & \multicolumn{1}{c}{BAY} & \multicolumn{1}{c}{COL} & \multicolumn{1}{c}{FLA} & \multicolumn{1}{c}{NW} & \multicolumn{1}{c}{NE} & \multicolumn{1}{c}{CAL} & \multicolumn{1}{c}{LKS} & \multicolumn{1}{c}{E} & \multicolumn{1}{c}{W} & \multicolumn{1}{c}{CTR} & \multicolumn{1}{c}{USA} \\

WC-A*      & \textbf{0.16} & \textbf{0.25} & 0.58          & 3.87          & 4.88          & 5.53          & 10.92         & 67.87          & 101.94         & 88.22          & 198.99          & 1886.43          \\

WC-BA*     & \textbf{0.16} & \textbf{0.25} & \textbf{0.53} & \textbf{2.96} & 5.87          & 5.84          & 4.99          & 75.14          & 101.10         & 108.73         & 214.06          & 1576.45          \\

WC-EBBA*   & 0.21          & 0.33          & 0.66          & 3.44          & 5.62          & 7.12          & 5.13          & 77.69          & 122.18         & 113.95         & 265.04          & 1516.78          \\

WC-EBBA*\textsubscript{2c} & 0.17          & 0.29          & 0.59          & 3.06          & \textbf{4.83} & \textbf{4.97} & \textbf{4.19} & \textbf{53.68} & \textbf{94.18} & \textbf{99.93} & \textbf{179.01} & \textbf{1194.90} \\

BiPulse    & 2.28          & 2.89          & 7.59          & 94.11         & 436.00        & 633.55        & 410.94        & 3839.58        & 4946.38        & 4757.22        & -             &  -        \\
\bottomrule


\end{tabular}
\end{adjustbox}
\end{table}
The table shows times in minutes, considering 60 minutes for the runtime of unsolved cases.
In this table, we are interested in analysing the performance of each algorithm on solving a set of WCSPP problems.
Boldface values denote the smallest cumulative runtime among algorithms (per map).
Looking at the results in Table~\ref{table:runtime_map}, we can clearly see the advantages of best-first search in A* against the depth-first search in BiPulse.
As an example, we can solve all 160 instances of the NE map in around 5-8 minutes using any of the A* algorithms, however, BiPulse needs about 11 hours to solve the same set of instances.
Among the A*-based algorithms, we can see {WC-EBBA*}\textsubscript{par} as the best performing algorithm in almost all larger maps.
However, the results show that {WC-BA*} and {WC-A*} perform better in small graphs.
Furthermore, if we compare our new algorithms against the state-of-the-art {WC-EBBA*}, we can see that they perform better on cumulative runtime than {WC-EBBA*} on two-thirds of the maps.
In the CTR map, for example, both {WC-A*} and {WC-BA*} solve all instances at least 15 minutes faster than {WC-EBBA*}.
In our largest map USA, however, we see our parallel {WC-EBBA*} algorithm saves at least 5.3 hours of computation time when compared with the other A*-based methods.

\begin{figure}[t]
\centering
\includegraphics[width=1\textwidth]{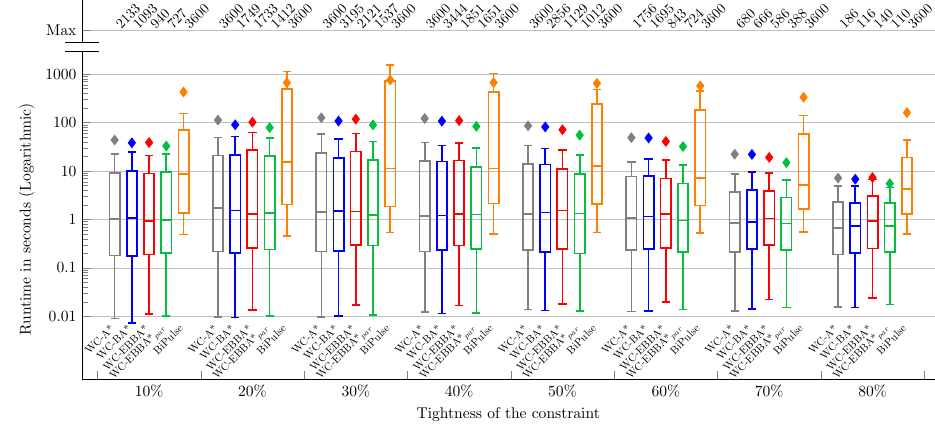}
\caption{\small Runtime (in seconds) of the algorithms in all levels of tightness. Values above the plots show the maximum runtime observed in the experiments (outliers are not shown).}
\label{fig:boxplot_runtime}
\end{figure}

\textbf{Distribution of runtimes:}
The plots in Figure~\ref{fig:boxplot_runtime}
also show the (logarithmic) runtime distribution of algorithms over all maps (250 instances in each constraint).
We do not show outliers in the box-plots, but the maximum runtime observed in each constraint is presented above the corresponding plot separately.
From the results in Figure~\ref{fig:boxplot_runtime}, we can see BiPulse as the weakest algorithm in all levels of tightness (the data does not include the larger maps CTR and USA).
Nonetheless, the plots show that there is at least one instance in each level of tightness where BiPulse is unable to solve it in the time limit (shown with the maximum runtime of 3600 seconds).
For the A*-based algorithms, however, we see far better results.
Looking at the range of runtimes across levels of tightness, we generally see larger runtimes in 20\%-50\% tightness.
It becomes obvious that the four unsolved cases of {WC-A*} are located in this range. 
Nonetheless, we can see that all algorithms perform faster in the looser weight constraints (from 50\% onward) and also in the very tight constraint 10\%.

\begin{figure}[!t]
\begin{subfigure}{0.47\textwidth}
\includegraphics[width=1\textwidth]{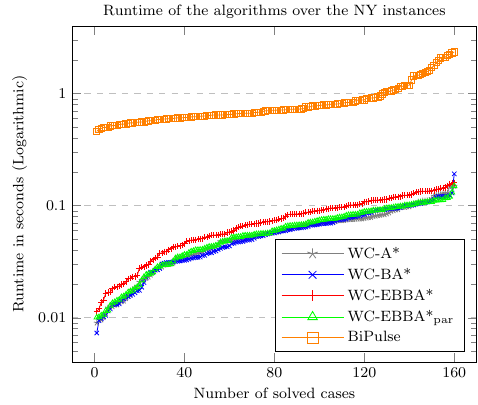} 
\end{subfigure}
\hfill
\vspace{0.5 \baselineskip}
\begin{subfigure}{0.49\textwidth}
\includegraphics[width=1\textwidth]{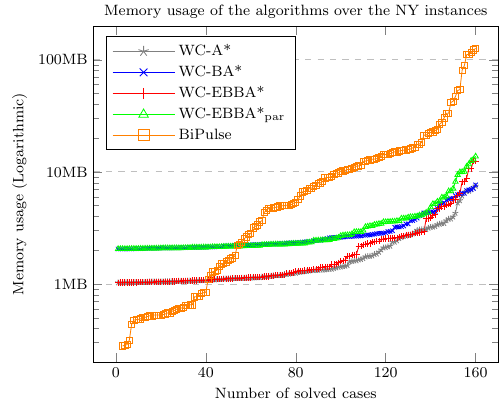}
\end{subfigure}
\vspace{0.5 \baselineskip}
\begin{subfigure}{0.47\textwidth}
\includegraphics[width=1\textwidth]{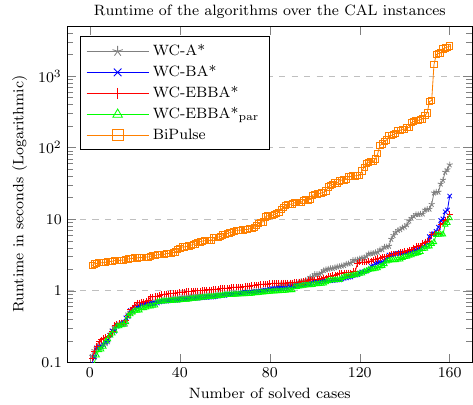}
\end{subfigure}
\hfill
\begin{subfigure}{0.49\textwidth}
\includegraphics[width=1\textwidth]{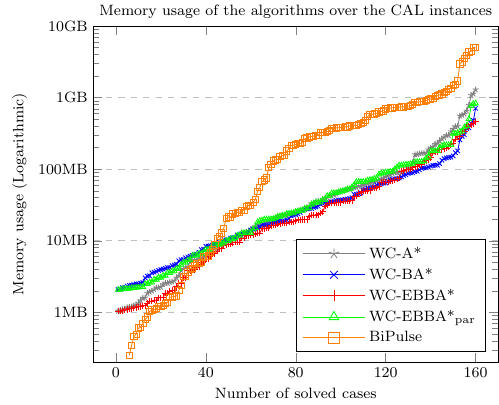}
\end{subfigure}
\begin{subfigure}{0.47\textwidth}
\includegraphics[width=1\textwidth]{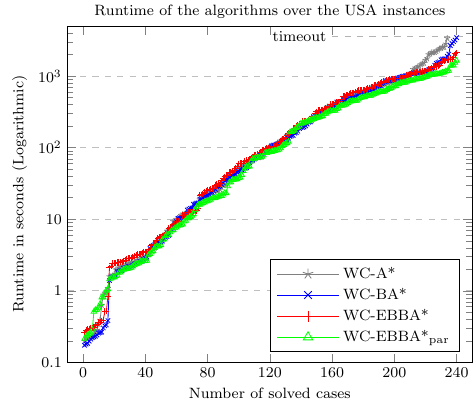}
\end{subfigure}
\hfill
\begin{subfigure}{0.49\textwidth}
\includegraphics[width=1\textwidth]{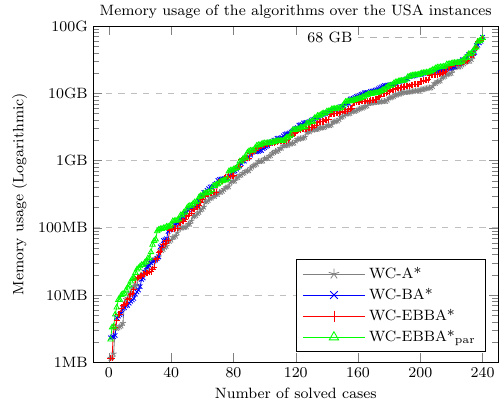}
\end{subfigure}
\caption{\small Cactus plots of algorithms' performance on the NY, CAL, and USA maps. Left: runtime (in seconds). Right: memory.
Memory is reported only for the solved cases. 1MB=10\textsuperscript{3}KB and 1GB=10\textsuperscript{3}MB.
}
\label{fig:cactus_performance}
\end{figure}

\textbf{Performance vs. solved cases:}
To better illustrate the performance of each algorithm over the instances, we show cactus plots of the runtime and memory requirements of algorithms over three maps with different sizes (NY, CAL, and USA) in Figure~\ref{fig:cactus_performance}.
The figure does not report BiPulse's performance on the USA map as previously highlighted.
The cactus plots help us to understand how fast/efficient algorithms are in limited time/space.
In the CAL map, for example, we can see that {WC-EBBA*} and {WC-EBBA*}\textsubscript{par} have been able to solve almost all the instances in under 10 seconds, whereas BiPulse solves around 80 instances (out of 160) within 10 seconds.
Figure~\ref{fig:cactus_performance} describes that BiPulse is dominated by all the A* algorithms in both runtime and memory usage.
However, the plot shows that none of the A*-based algorithms consistently dominates the other ones in both metrics, and we cannot declare a clear winner.
Having said that, we can see better runtime performance for {WC-EBBA*}\textsubscript{par} and lighter memory usage for {WC-A*} across the instances of our largest map USA.

It is also worth discussing cases where BiPulse uses less memory than our A*-based algorithms in Figure~\ref{fig:cactus_performance}, in the CAL instance, for example.
The difference here is mainly rooted in the dynamic node allocation in our implementation.
In contrast to methods where each node is individually allocated a space upon creation, our memory manager generates a block of nodes and makes them available to the search.
If the search requires more nodes, the code then generates another block of nodes (in a form of two-dimensional memory allocation with about 1~MB size).
Hence, there may be cases where the code generates a few more nodes than what the search requires, and thus we see no significant changes in the memory plots.
Each search in our parallel algorithms has its own memory manager, so {WC-BA*} and {WC-EBBA*}\textsubscript{par} start with a bigger pre-allocated memory (about 2~MB).
\subsection{Comparison with Other Constrained Search Methods}
\label{sec:experment_result_SM}
We extend our empirical study and compare our proposed algorithms with the existing solutions that have shown good performance on large networks, though proven to be less effective than {WC-EBBA*} in \citet{AhmadiTHK21}.
The selected algorithms
are CSP \cite{sedeno2015enhanced}, Pulse \cite{lozano2013exact} and RC-BDA* \cite{thomas2019exact}.
None of the algorithms use parallelism.
Thus, for the sake of fairness, we compare them against our unidirectional algorithm {WC-A*} only.
We used the C implementation of the CSP and Pulse algorithms kindly provided to us by \citet{sedeno2015enhanced}.
We noticed that there is no path-tracking method implemented for Pulse.
For the {RC-BDA*} algorithm, we were unable to obtain the original implementations. We, therefore, implemented the algorithms based on the descriptions provided in the original paper.
Since the structure of {RC-BDA*} is similar to our bidirectional algorithm {WC-EBBA*}, we used our improved implementation of {RC-BDA*} in \citet{AhmadiTHK21}.
We used the same set of 2000 WCSPP instances described earlier in the section, and ran all experiments on a single core of an Intel Xeon-Platinum-8260 @ 2.50GHz processor running at 2.5~GHz and with 128~GB of RAM, under the CentOs Linux 7 environment and with a one-hour timeout.
Table~\ref{table:Results_extended} presents the experimental results for all the algorithms.
\begin{table}[ht]
\caption{\small Number of solved cases $|S|$ (out of 240 for USA and 160 for the other maps), runtime and memory use of the algorithms. Runtime of unsolved instances is considered 3600~seconds. Memory is reported only for the solved cases and 1MB=10\textsuperscript{3}KB.}
\label{table:Results_extended}
\begin{multicols}{2}
\centering
\small
\begin{adjustbox}{width=1\textwidth}
\renewcommand{\arraystretch}{1}
\begin{tabular}{| lll | *{3}{r} | *{2}{r}|}
\hline
 & & & \multicolumn{3}{c|}{Runtime(s)} & \multicolumn{2}{c|}{Memory(MB)} \\ \cline{4-8}
\headrow
\textbf{Map} & \textbf{Algorithm} & \textbf{$|S|$} & \textbf{Min} & \textbf{Avg.} & \textbf{Max} & \textbf{Avg.} & \textbf{Max} \\
\hline
NY  & Pulse   & 140 & 0.13          & 539.16        & 3600.00        & 0   & 4    \\
    & CSP     & 160 & 0.15          & 2.07          & 22.43          & 14  & 89   \\
    & RC-BDA*  & 160 & 0.36          & 2.88          & 86.84          & 15  & 100   \\
    & WC-A*    & 160 & \textbf{0.02} & \textbf{0.09} & \textbf{0.30}           & 3   & 8    \\
\hline
BAY & Pulse   & 128 & 0.15          & 836.64        & 3600.00        & 0   & 4    \\
    & CSP     & 160 & 0.18          & 5.08          & 67.26          & 28  & 292  \\
    & RC-BDA*  & 156 & 0.45          & 152.17        & 3600.00        & 480 & 8991 \\
    & WC-A*    & 160 & \textbf{0.02} & \textbf{0.14} & \textbf{0.79}           & 4   & 23   \\
\hline
COL & Pulse   & 83  & 0.20          & 1895.88       & 3600.00        & 1   & 4    \\
    & CSP     & 160 & 0.25          & 32.85         & 655.97         & 84  & 934  \\
    & RC-BDA*  & 158 & 0.61          & 123.20        & 3600.00        & 97  & 702  \\
    & WC-A*    & 160 & \textbf{0.03} & \textbf{0.35} & \textbf{3.56}           & 10  & 75   \\
\hline
FLA & Pulse   & 37  & 0.61          & 2864.50       & 3600.00        & 1   & 5    \\
    & CSP     & 158 & 1.51          & 379.87        & 3600.00        & 419 & 3710 \\
    & RC-BDA*  & 102 & 1.77          & 1375.36       & 3600.00        & 185 & 809  \\
    & WC-A*    & 160 & \textbf{0.22} & \textbf{2.37}          & \textbf{35.82}          & 44  & 344  \\
\hline
NW  & Pulse   & 34  & 1.64          & 2970.95       & 3600.00        & 1   & 5    \\
    & CSP     & 159 & 0.79          & 469.50        & 3600.00        & 689 & 4573 \\
    & RC-BDA*  & 99  & 1.96          & 1624.83       & 3600.00        & 315 & 1794 \\
    & WC-A*    & 160 & \textbf{0.15} & \textbf{3.05} & \textbf{26.79}          & 62  & 325  \\
\hline
NE  & Pulse   & 26  & 1.35          & 3118.31       & 3600.00        & 1   & 5    \\
    & CSP     & 155 & 1.09          & 474.21        & 3600.00        & 423 & 3258 \\
    & RC-BDA*  & 145 & 2.44          & 642.43        & 3600.00        & 577 & 5386 \\
    & WC-A*    & 160 & \textbf{0.23} & \textbf{3.07} & \textbf{43.99} & 55  & 561 \\
\hline
\end{tabular}
\begin{tabular}{| lll | *{3}{r} | *{2}{r} |}
\hline
\hiderowcolors
 & & & \multicolumn{3}{c|}{Runtime(s)} & \multicolumn{2}{c|}{Memory(MB)} \\ \cline{4-8}
 \headrow
\textbf{Map} & \textbf{Algorithm} & \textbf{$|S|$} & \textbf{Min} & \textbf{Avg.} & \textbf{Max} & \textbf{Avg.} & \textbf{Max} \\
\showrowcolors
CAL & Pulse   & 22  & 1.14          & 3149.95         & 3600.00          & 1    & 6     \\
    & CSP     & 146 & 1.21          & 719.43          & 3600.00          & 569  & 4389  \\
    & RC-BDA*  & 92  & 3.11          & 1714.49         & 3600.00          & 472  & 3802  \\
    & WC-A*    & 160 & \textbf{0.16}          & \textbf{6.71}            & \textbf{145.31}           & 101   & 1301  \\
\hline
LKS & Pulse   & 24  & 1.48          & 3139.85         & 3600.00          & 1    & 5     \\
    & CSP     & 98  & 1.73          & 1859.10         & 3600.00          & 805  & 3592  \\
    & RC-BDA*  & 80  & 4.45          & 2067.73         & 3600.00          & 577  & 2856  \\
    & WC-A*    & 160 & \textbf{0.12}          & \textbf{41.75}  & \textbf{467.19}  & 513  & 4258  \\
\hline
E   & Pulse   & 24  & 5.77          & 3159.62         & 3600.00          & 1    & 5     \\
    & CSP     & 88  & 3.29          & 2041.80         & 3600.00          & 840  & 4084  \\
    & RC-BDA*  & 71  & 6.24          & 2264.51         & 3600.00          & 614  & 3347  \\
    & WC-A*    & 160 & \textbf{0.11} & \textbf{58.19}  & \textbf{534.28}  & 678  & 5184  \\
\hline
W   & Pulse   & 7   & 3.72          & 3444.79         & 3600.00          & 0    & 0     \\
    & CSP     & 89  & 4.29          & 2079.09         & 3600.00          & 870  & 3934  \\
    & RC-BDA*  & 48  & 11.10         & 2689.49         & 3600.00          & 461  & 3019  \\
    & WC-A*    & 160 & \textbf{0.44} & \textbf{48.31}  & \textbf{421.11}  & 575  & 5014  \\
\hline
CTR & Pulse   & 9   & 31.80         & 3429.75         & 3600.00          & 0    & 0     \\
    & CSP     & 75  & 14.10         & 2273.46         & 3600.00          & 645  & 3173 \\
    & RC-BDA*  & 60  & 31.13         & 2429.42         & 3600.00          & 488  & 2880  \\
    & WC-A*    & 160 & \textbf{0.33} & \textbf{118.10} & \textbf{1478.18} & 1072 & 7557  \\
\hline
USA & Pulse   & 7   & 16.22         & 3526.22         & 3600.00          & 0    & 0\\
    & CSP     & 64  & 17.89         & 2838.22         & 3600.00          & 715  & 3692  \\
    & RC-BDA*  & 49  & 46.81         & 2972.83         & 3600.00          & 498  & 2355  \\
    & WC-A*    & 229 & \textbf{0.22} & \textbf{622.32}          & 3600.00          & 4731 & 29861 \\
\hline
\end{tabular}
\end{adjustbox}
\end{multicols}
\end{table}

The table shows the number of instances solved to optimality within the timeout (denoted by $|S|$) and minimum, average and maximum runtime in \textit{seconds}, including the initialisation time.
Boldface values denote the best runtime among the algorithms.
For unsolved instances, the runtime is considered to be the timeout.
We also report for each map the average and maximum memory (in MB) used by each algorithm over the solved instances. 
Among the competitors, CSP shows the best performance and solves more cases compared to {RC-BDA*} and Pulse, but far less than {WC-A*} in larger maps.
For example, more than half of the instances of the CTR map remained unsolved after the one-hour timeout.
In contrast, our {WC-A*} algorithm fully solves all instances of 11 maps, and is around two orders of magnitude faster than CSP in the COL map (on average runtime) where both {WC-A*} and CSP solve all instances.
In terms of memory use, we can see that {WC-A*} require far less memory than CSP and {RC-BDA*}.
In particular, CSP's average memory requirement to solve the NW instances is 10 times the average space needed by {WC-A*}.
The other algorithm, Pulse, seems efficient in terms of memory use, which is mainly due to its branch-and-bound nature and also not having a path tracking method implemented.
However, it is very slow in almost all maps, showing very large runtimes even for the instances of our smallest graph NY.

\subsection{Priority Queue Impacts: More detailed analysis}
\label{sec:PQ_impact_extended}
\begin{figure}[!t]
\begin{subfigure}{0.49\textwidth}
\includegraphics[width=1\textwidth]{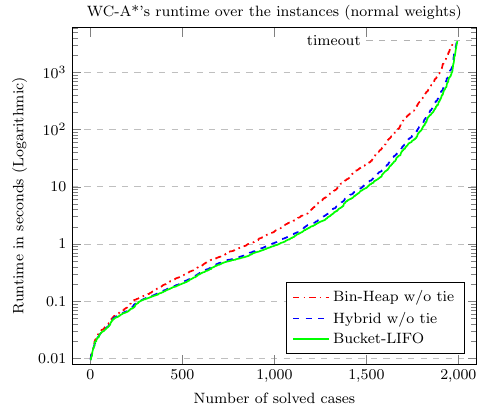} 
\end{subfigure}
\hfill
\begin{subfigure}{0.49\textwidth}
\includegraphics[width=1\textwidth]{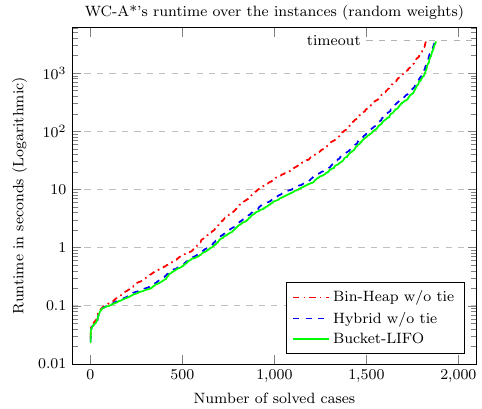}
\end{subfigure}
\caption{\small Cactus plots of the {WC-A*}'s performance over all instances with three priority queues on the original DIMACS graphs (left) and randomised graphs (right). Instances are sorted based on the runtime.}
\label{fig:pqueue_without}
\end{figure}

\textbf{Performance vs. solved cases:}
Following our discussion in Section~\ref{sec:experiment_extended}, we extend our analysis and compare in Figure~\ref{fig:pqueue_without} the performance of {WC-A*} using the faster variant of each queue (binary heap and hybrid queues without tie-breaking, and the LIFO strategy for the bucket queue with linked lists) to better understand the impacts of each type of priority queue on our constrained A* search.
The figure shows three cactus plots for both (original and randomised) types of graphs.
We see that {WC-A*} with bucket queues outperforms the best-performer of other types in both graph types.
Nonetheless, {WC-A*} with the hybrid queue is still significantly faster than {WC-A*} with the conventional binary heap queue.
In addition, we see larger gaps in the randomised graph, which indicate that both queues (binary heap and hybrid) are vulnerable to difficult problems with a significant number of node expansions.

\textbf{Hybrid queues vs. bucket queues:}
The results in Tables~\ref{table:runtime_tie_org_map} and \ref{table:runtime_tie_rand_map} illustrated that hybrid queues without tie-breaking are not as effective as bucket queues with linked lists.
One can see hybrid queues without tie-breaking and $\Delta f=1$ as a form of bucket queue, but there is one potential reason for this considerable runtime difference in Tables~\ref{table:runtime_tie_org_map}~and~\ref{table:runtime_tie_rand_map}.
Both queues initialise and traverse the same number of buckets.
In the bucket queue with linked lists and $\Delta f=1$ (as a one-level queue), we always have constant time (push and pop) operations in (high-level) buckets.
In the hybrid queue with $\Delta f=1$ and without tie-breaking, we always extract nodes from the low-level binary heap but add nodes to high-level buckets or the binary heap, depending on the nodes' $f$-value.
Since all nodes in the bucket are of the same primary cost, and since we do not aim to break the tie between nodes, every insertion or deletion (push or pop operation) in the low-level binary heap becomes a constant time operation.
However, there is always a form of overhead when ordering nodes with hybrid queues, which is rooted in the mechanism of transferring nodes from a high-level bucket to the low-level binary heap (this can be avoided in one-level bucket queues).
Therefore, although both bucket and hybrid queues offer fast operations in cases with integer costs and $\Delta f =1$, hybrid queues are less likely to outperform bucket queues due to the aforementioned overhead caused by its two-level structure. 

\textbf{Special cases:}
Our results show that {WC-A*} with bucket-based queues and $\Delta f = 1$ outperforms {WC-A*} with binary heap in both realistic and randomised graph.
However, for the WCSPP on large graphs, there might be cases where bucket-based queues do not offer any performance improvement over binary heap queues if the edge costs are extremely small/large values.
On the one hand, if the edge costs are very large values, scanning the entire range of buckets may become very expensive in terms of time.
On the other hand, if edge costs are very small values, selecting $\Delta f = 1$ may turn the hybrid queue into a binary heap.
This is because all nodes will fall into the first bucket if $\lfloor f_{\mathit{max}} \rfloor = \lfloor f_{\mathit{min}} \rfloor$.
The most straightforward approach in dealing with such special cases is scaling (all) edge costs. In this approach, we need to make sure that the scaling does not produce very small/large values.
However, if we have both extreme cases in the graph where cost scaling is not effective, an alternative approach could be changing the bucket width based on $f_{\mathit{max}} - f_{\mathit{min}}$.
If edge costs are integer values, our bucket queue would then turn into a two-level bucket structure with $\Delta f > 1$, as explained in Section~\ref{priority_queus}.
In other words, we can increase the bucket width $\Delta f$ to reduce the computational costs incurred by the lengthy bucket list in case of very large $\lfloor f_{\mathit{max}} - f_{\mathit{min}} \rfloor$, or adding more buckets by defining $\Delta f < 1$ in case of very small $\lfloor f_{\mathit{max}} - f_{\mathit{min}} \rfloor$.
For this approach, we may need to study the trade-off between the bucket width and bucket size to determine the most effective value of our essential parameter, $\Delta f$.

\subsection{Tie-breaking Impacts: More Detailed Analysis}
We observed in the previous section that bucket queues and hybrid queues are far faster than traditional binary heaps for the constrained search using A*.
Furthermore, the experimental results on the original and randomised graphs presented in Section~\ref{sec:experiment_extended} showed that tie-breaking does not necessarily lead to faster search in exhaustive label-setting approaches, such as all of our A*-based algorithms.
In this section, we investigate in detail potential reasons for these two observations, namely by analysing the operations in the constrained search of {WC-A*} and also in the priority queue.
In terms of search operations, we already discussed that disabling tie-breaking may lead to expanding dominated nodes, whereas performing lexicographical ordering (enabling tie-breaking) guarantees that none of our A* algorithms will expand (weakly) dominated nodes since such nodes are always pruned via dominance tests.
Therefore, if we use {WC-A*} to solve an instance of the WCSPP with any type of priority queue able to handle tie-breaking, we should see the same number of node expansions.
In other words, {WC-A*} with either of binary heap and hybrid queues will expand the same number of nodes if we force the internal binary heaps to break the ties.
In addition, compared to the number of nodes expanded in the presence of tie-breaking, we know that it is not possible for the search to solve each instance with fewer node expansions when tie-breaking is disabled.
Hence, for the analysis of this section, we measure the total queue operations and total expanded nodes to investigate why priority queues work much better without tie-breaking.
We define the parameters as follows.
\begin{itemize}[nosep]
    \item Queue operations:
    For binary heap queues, we report the total number of (node) swaps performed in the heap over the search.
    For hybrid queues, we report the total number of buckets checked/drained, plus the number of nodes transferred from high-level buckets to the low-level heap, plus the total number of (node) swaps performed in the low-level binary heap.
    For bucket queues, we report the total number of buckets checked/drained during the search. We consider linked list operations to require constant time in this analysis. 
    \item Expanded nodes:
    We report the total number of nodes {WC-A*} expands during the search.
    This number includes dominated nodes expanded by the algorithm (via the ExP($x,d$) procedure) in the absence of tie-breaking.
\end{itemize}
For all the plots we present in this section, instances (in the horizontal axis) are sorted based on the number of non-dominated node expansions, i.e., nodes expanded in {WC-A*} using binary heap with tie-breaking enabled.
To be consistent with the previous analysis, we set $\Delta f = 1$ for both bucket and hybrid queues.
In addition, for the sake of fairness, we only consider instances mutually solved by all types of priority queues.

\begin{figure}[t]
\begin{subfigure}{0.49\textwidth}
\includegraphics[width=1\textwidth]{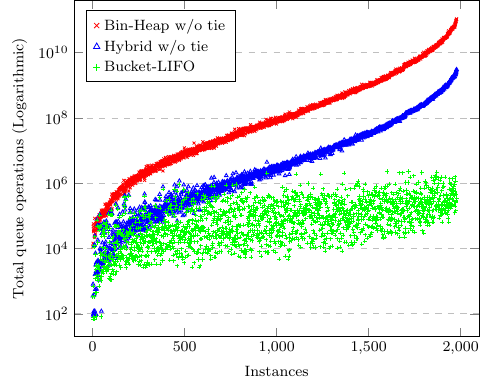} 
\end{subfigure}
\hfill
\begin{subfigure}{0.49\textwidth}
\includegraphics[width=1\textwidth]{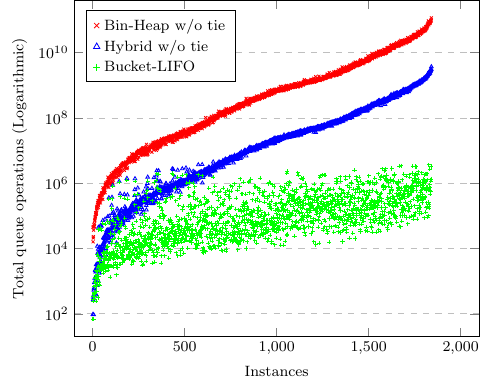}
\end{subfigure}
\caption{\small Total number of queue operations in each type of priority queue on the original DIMACS graphs (left) and randomised graphs (right). Instance are sorted based on the number of node expansions. The plot only shows instanced mutually solved via all queue types.}
\label{fig:pqueue_total_operation}
\end{figure}

\textbf{Queue operations vs. queue type:}
We compare the total number of queue operations in each type of priority queue.
We consider their best variant, i.e., no tie-breaking in binary heap and hybrid queues, and the LIFO variant in bucket queues.
Figure~\ref{fig:pqueue_total_operation} shows the results on the original DIMACS graphs (left) and randomised graphs (right).
In both plots, we can see that the binary heap queue performs far more operations than the other two priority queues.
In particular, the number of operations in the binary heap queue is on average 10 times the number of operations in the hybrid queue.
We can see on both graphs, the binary heap queue performs around 100 billion operations during the constrained search of the problem instance with the most expanded nodes.
While the situation is less severe in the hybrid queue, we can always achieve fewer operations with bucket queues.
The figure also shows that the bucket queue might perform similarly to hybrid queues (in terms of total operations) in cases with small number of node expansions, but it significantly outperforms other queue types (by several orders of magnitude) in problem instances with a large number of node expansions.
Interestingly, from the detailed results, we observed only four cases (over all instances on both graph types) where the binary heap queue had the smallest number of operations.

\begin{figure}[!t]
\begin{subfigure}{0.49\textwidth}
\includegraphics[width=1\textwidth]{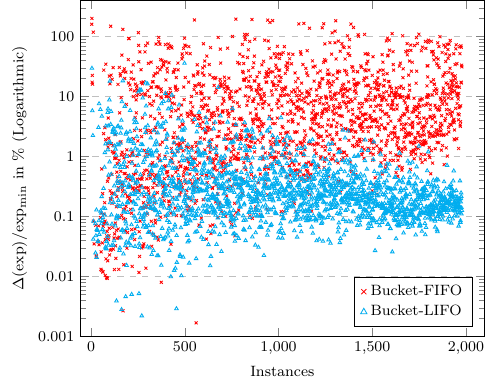} 
\end{subfigure}
\hfill
\vspace{1.0 \baselineskip}
\begin{subfigure}{0.49\textwidth}
\includegraphics[width=1\textwidth]{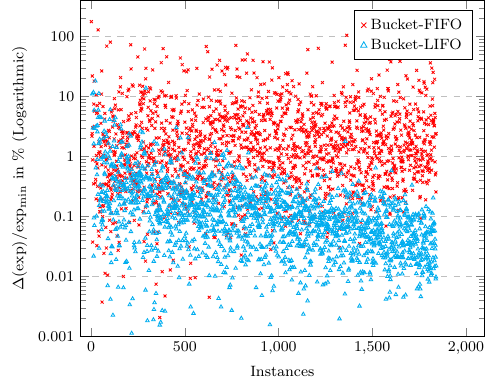}
\end{subfigure}
\begin{subfigure}{0.49\textwidth}
\includegraphics[width=1\textwidth]{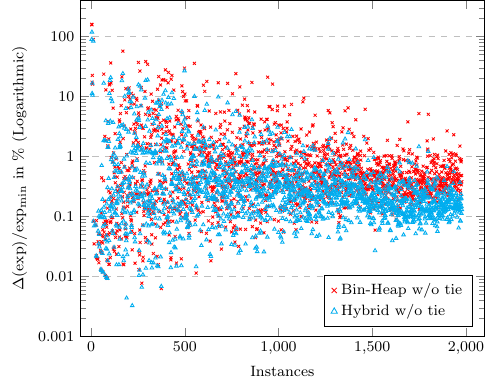} 
\end{subfigure}
\hfill
\begin{subfigure}{0.49\textwidth}
\includegraphics[width=1\textwidth]{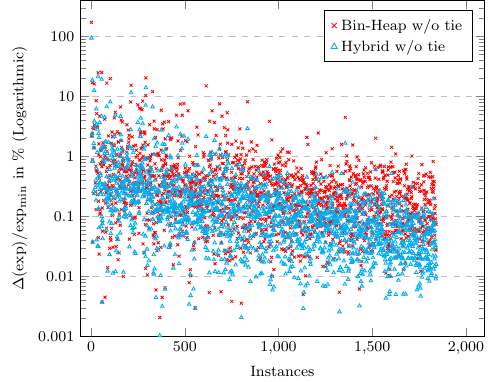}
\end{subfigure}
\caption{\small Percentage of extra expansions in each type of priority queue on the original DIMACS graphs (left) and randomised graphs (right). Instances are sorted based on the number of node expansions in {WC-A*}.}
\label{fig:pqueue_total_expansion}
\end{figure}

\textbf{Extra node expansions without tie-breaking:}
We calculate for every solved instance the percentage of extra expansions if there is no tie-breaking in place.
For our studied priority queues, the plots in Figure~\ref{fig:pqueue_total_expansion} show the percentage of extra expansions across all mutually solved instances on both graphs.
In this figure, $\Delta$(exp) and exp\textsubscript{min} denote extra expansions and minimum required expansions (obtained via tie-breaking), respectively. 
We first discuss the results for bucket queues.
As mentioned earlier, none of our bucket queues are able to handle tie-breaking,
so there is always a risk of expanding (weakly) dominated nodes with bucket queues regardless of the node extraction strategy.
Figure~\ref{fig:pqueue_total_expansion} shows the percentage of extra expansions if we use bucket queues in {WC-A*} with either of LIFO and FIFO strategies.
Surprisingly, bucket queues with the FIFO strategy can potentially expand 100\% extra nodes in {WC-A*}.
We see more critical situations on the realistic graphs (left plot) where the objectives are correlated, and thus the search generates more dominated nodes.
However, we observe that the bucket queue with the LIFO strategy generates far fewer extra nodes, less than 1\% on many instances.
We further elaborate this behaviour using an example.

\textbf{Example:}
Imagine the algorithm has inserted three nodes $x_1,x_2,x_3$ into the bucket queue in order.
These nodes are all in the same bucket and all associated with the same state $\mathit{u}$, i.e., we have $f_p(x_1) = f_p(x_2) = f_p(x_3)$.
In addition, we assume $x_3$ dominates $x_2$, and $x_2$ dominates $x_1$, i.e., we have $f_s(x_1) > f_s(x_2) > f_s(x_3)$.
We expand nodes in order and achieve $y_1,y_2,y_3$ for the adjacent state $v$ via the link ($u,v$).
We were unable to prune dominated nodes $x_1,x_2$ because the search expanded the non-dominated node $x_3$ later.
Nodes $y_1,y_2$ and $y_3$ are inserted into one of the next buckets, essentially because they still have $f_p(y_1) = f_p(y_2) = f_p(y_3)$.
Note that we first expanded $x_1$, so $y_1$ is inserted earlier than $y_2$ and $y_3$.
The same applies to $y_2$ and the search inserts it to the bucket before $y_3$. In addition, we can see that $y_3$ dominates $y_2$, and $y_2$ dominates $y_1$ (they have been extended via the same link).
During the search, we eventually reach the bucket of nodes $y_1,y_2,y_3$.
If we extract nodes based on the FIFO strategy, we have to expand dominated nodes $y_1$ and $y_2$ before expanding the non-dominated node $y_3$.
In the LIFO strategy, however, we start with the non-dominated node $y_3$ (as the most recent insertion) and will be able to prune earlier insertions $y_2$ and $y_1$ (via the dominance test).
Therefore, we can see that the LIFO strategy is more effective in preventing cascading expansions of dominated nodes.

The results presented in Figure~\ref{fig:pqueue_total_expansion} explain the performance difference (for bucket queues) in Tables~\ref{table:runtime_tie_org_map} and \ref{table:runtime_tie_rand_map}, confirming our previous observation on cascading expansions of dominated nodes in the FIFO strategy.
Nonetheless, we see smaller percentages for both strategies on graphs with random costs, mainly around 0.1\% (resp. 1\%) for the LIFO (resp. FIFO) strategy.
The figure also shows the percentage of extra expansions for the binary heap and hybrid queues without tie-breaking.
Similar to the previous analysis, we see larger values on the realistic graphs, and also smaller ratios in instances with more expansions (instances are sorted based on their minimum number of expansions).
Although both queues are very close in the percentage of extra expansions (around 0.1\% on both graph types), the plots show larger values for the binary heap queue across instances of both graph types.
There is also one other observation if we (indirectly) compare the values for bucket and hybrid queues.
The plots show that the bucket queue with the LIFO strategy works quite similarly with the hybrid queue in terms of extra expansions (both around 0.1\%).
This observation highlights that, in problem instances with integer costs, ordering nodes via binary heap is even more effective than linked lists with the FIFO strategy in preventing dominated node expansions.
Note that the node ordering in the low-level binary heap of the hybrid queue is neither LIFO nor FIFO.

\begin{figure}[!t]
\begin{subfigure}{0.49\textwidth}
\includegraphics[width=1\textwidth]{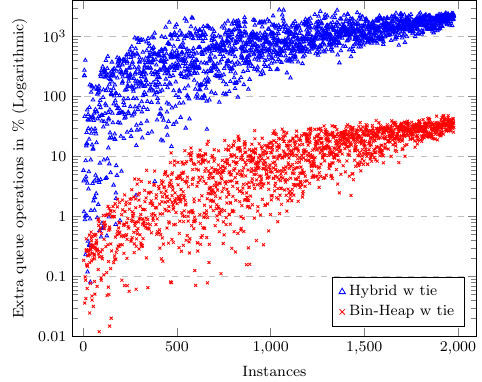} 
\end{subfigure}
\hfill
\begin{subfigure}{0.49\textwidth}
\includegraphics[width=1\textwidth]{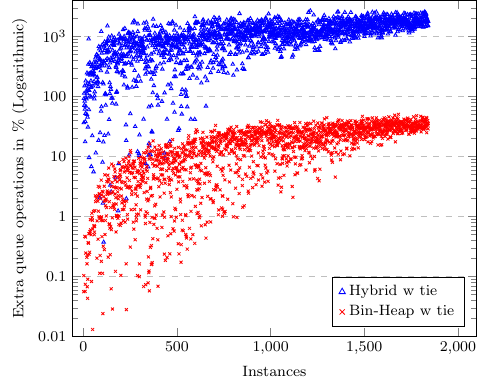}
\end{subfigure}
\caption{\small Percentage of extra queue operations in the binary-heap and hybrid priority queues on the original DIMACS graphs (left) and randomised graphs (right). Instances are sorted based on the number of node expansions.}
\label{fig:pqueue_operation}
\end{figure}

\textbf{Extra queue operations with tie-breaking:}
So far, we have seen that disabling tie-breaking can potentially lead to expanding as much as 100\% extra nodes in some priority queues.
We now investigate why binary heap and hybrid queues still perform poorly with tie-breaking.
For this purpose, we calculate for every solved instance the percentage of extra queue operations if we enable tie-breaking.
More precisely, we aim to figure out how many queue operations we actually save (or possibly waste) by not doing tie-breaking.
Hence, we consider the baseline to be the queue type without tie-breaking.
Figure~\ref{fig:pqueue_operation} shows for binary heap and hybrid queues the percentage of extra queue operations with tie-breaking on both realistic and randomised graphs.
Note that there might be cases where the queue performs fewer operations in the presence of a tie-breaker (because of fewer expansions), but we observed that such cases do not normally happen in difficult instances and are not shown in Figure~\ref{fig:pqueue_operation} (due to the logarithmic scale).
According to the results on both graph types, the binary heap queue performs up to 50\% more operations during the search if the tie-breaking is enabled.
This value for the hybrid queue is as big as 2800\%.
There is one potential reason for this huge difference in the number of extra operations.
In the hybrid queue with $\Delta f =1 $, the low-level binary heap is responsible for breaking the ties between all nodes showing the smallest $f_p$-value after every extraction and insertion (either nodes directly added to the low-level heap or nodes transferred from a high-level bucket).
In other words, the hybrid queue with $\Delta f =1 $ collects all tied nodes in the same bucket, so the low-level binary heap will inevitably have to perform far more operations compared with the other form (with no tie-breaking).
In the binary heap queue, however, tied nodes are distributed in the entire queue and thus tie-breaking would likely happen on a fraction of tied nodes.
Hence, tie-breaking in the binary heap queue involves fewer operations than the hybrid queue with $\Delta f =1$.
In addition, if we compare the results in Figures~\ref{fig:pqueue_total_expansion} and \ref{fig:pqueue_operation}, we notice that enabling tie-breaking leads to more queue operations than extra node expansions (Figure~\ref{fig:pqueue_operation} shows significantly larger percentages). 
Therefore, we can now relate the performance difference in Tables~\ref{table:runtime_tie_org_map} and \ref{table:runtime_tie_rand_map} to the results in Figures~\ref{fig:pqueue_total_expansion} and \ref{fig:pqueue_operation}.
\begin{figure}[!t]
\begin{subfigure}{0.49\textwidth}
\includegraphics[width=1\textwidth]{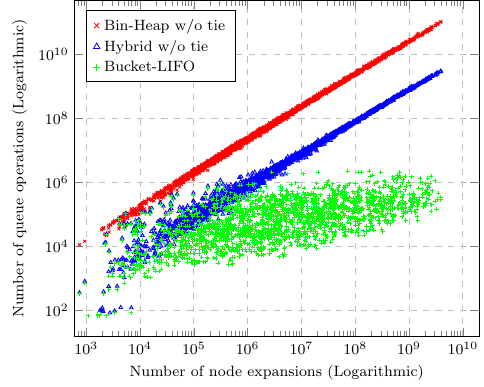} 
\end{subfigure}
\hfill
\begin{subfigure}{0.49\textwidth}
\includegraphics[width=1\textwidth]{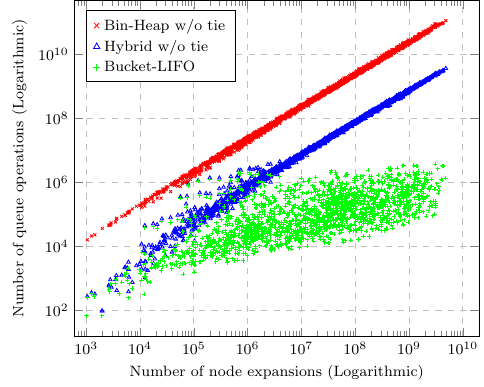}
\end{subfigure}
\begin{subfigure}{0.49\textwidth}
\includegraphics[width=1\textwidth]{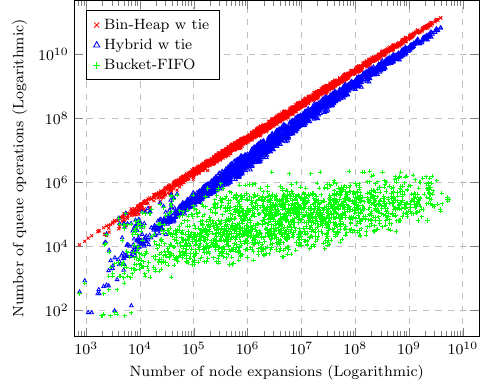} 
\end{subfigure}
\hfill
\begin{subfigure}{0.49\textwidth}
\includegraphics[width=1\textwidth]{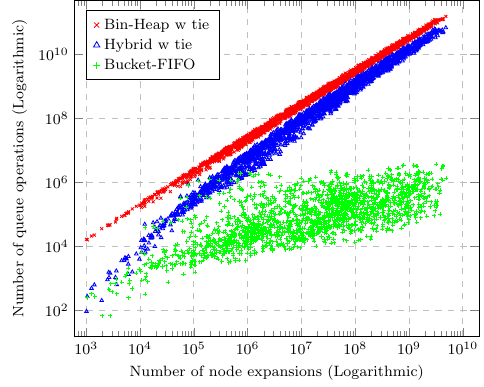}
\end{subfigure}
\caption{\small Total queue operations versus total node expansions for three types of priority queues studied using {WC-A*} on the original DIMACS graphs (left) and randomised graphs (right).}
\label{fig:pqueue_expansion_operation}
\end{figure}

\textbf{Queue operations vs. node expansions:}
We conclude our analysis by presenting in Figure~\ref{fig:pqueue_expansion_operation} scatter plots on the number of expanded nodes and total queue operations using three queue types. 
These plots help us to better understand the difference between the performance of {WC-A*} with each queue type in Figure~\ref{fig:pqueue_without}.
According to the results on both (realistic and randomised) graph types, we can see that the number of queue operations linearly increases with the number of node expansions in the binary heap and hybrid queues. 
For the bucket queue, however, the number of queue operations does not linearly grow with the number of expanded nodes.
This is because the bucket size is always fixed (known in advance) and the search will never exceed this (always bounded) size.
Comparing the ratio of queue operations and node expansions, we can observe that the binary heap queue performs far more queue operations than node expansions in general.
In particular, this queue type performs about 27 (resp. 23) operations per node expansion on average with tie-breaking (resp. without tie-breaking).
This ratio is around one for the hybrid queue without tie-breaking, and around eight when we enable tie-breaking in the low-level heap.
This observation elaborates the significance of high-level buckets in reducing the cost of node ordering in the exhaustive search of A*.
Nevertheless, the plots show cases when {WC-A*} with bucket queue performs more queue operations than node expansions.
In such cases, we can claim that the bucket list has had less than one node per bucket on average.
In other words, we have had more buckets than the total number of nodes expanded.

\end{document}